\documentclass[10pt]{article}
\usepackage[accepted]{tmlr}

\usepackage{amsmath,amsfonts,bm}

\DeclareMathOperator{\prox}{prox}
\DeclareMathOperator{\dist}{dist}
\DeclareMathOperator{\dom}{dom}

\def\eqref#1{equation~\ref{#1}}

\def\1{\bm{1}}

\def\eps{{\epsilon}}

\DeclareMathAlphabet{\mathsfit}{\encodingdefault}{\sfdefault}{m}{sl}
\SetMathAlphabet{\mathsfit}{bold}{\encodingdefault}{\sfdefault}{bx}{n}

\newcommand{\E}{\mathbb{E}}
\newcommand{\Ls}{\mathcal{L}}
\newcommand{\R}{\mathbb{R}}

\usepackage{hyperref}
\usepackage{url}
\usepackage{amsthm}
\usepackage{amssymb}
\usepackage{algorithm}
\let\OldAND\AND
\let\AND\relax
\usepackage{algpseudocode}
\let\AND\OldAND

\usepackage{booktabs}
\usepackage{graphicx}
\usepackage{enumitem}

\newtheorem{theorem}{Theorem}
\newtheorem{proposition}{Proposition}
\newtheorem{lemma}{Lemma}

\theoremstyle{definition}
\newtheorem{assumption}{Assumption}

\theoremstyle{remark}
\newtheorem{remark}{Remark}

\title{Stochastic Difference-of-Convex Optimization with Momentum}
\author{\name El Mahdi Chayti \email el-mahdi.chayti@epfl.ch\\
      \name Martin Jaggi \email martin.jaggi@epfl.ch \\
      \addr Machine Learning and Optimization Laboratory (MLO), EPFL, Switzerland}

\def\month{10}
\def\year{2026}
\def\openreview{\url{https://openreview.net/forum?id=7aIxLa5L9E}}

\begin{document}
\maketitle

\begin{abstract}
We study the online stochastic difference-of-convex (DC) program $\min_{x \in \R^n} F(x) = G(x) - H(x) + r_1(x) - r_2(x)$, which covers sparse learning with nonconvex penalties, robust regression with smooth robust losses, and non-negative positive--unlabeled learning. We propose \emph{momentum-based DCA} (MDCA), which combines a damped DCA step with a momentum estimator for $\nabla H$ and uses a constant per-iteration batch size. With exact solutions of the convex subproblems, MDCA with Polyak momentum uses $\mathcal{O}( \eps^{-4})$ stochastic gradients of $H$ to reach $\eps$-stationarity, and momentum-based variance reduction (MVR) uses $\mathcal{O}(\eps^{-3})$ under averaged smoothness, with a single $\mathcal O(\eps^{-1})$ warm-up batch in place of the periodic large batches of DCA-PAGE, whose $\eps$-dependence it matches. We also introduce a single-loop variant (S-MDCA) that replaces the subproblem solve by one stochastic prox-gradient step; for every fixed smoothing parameter $\gamma$ for which the separately smoothed surrogate $F_\gamma$ is bounded below, it reaches $\eps$-stationarity of $F_\gamma$ in $\mathcal{O}(\eps^{-4})$ iterations, which does not by itself imply stationarity of $F$. A variance--drift trade-off lemma describes how the momentum parameter trades estimator variance against iterate-drift bias in steady-state upper bounds. Guarantees are given for two criteria: the critical-distance measure, and a DCA fixed-point residual that requires no smoothness of $r_2$ and vanishes exactly at DC-critical points; both give the same $\eps$-dependence. Experiments on two LIBSVM classification datasets illustrate the methods' behaviour.
\end{abstract}

\section{Introduction}
\label{sec:intro}

We consider
\begin{equation}
\label{eq:dc-problem}
\min_{x \in \R^n}\; F(x) := G(x) - H(x) + r_1(x) - r_2(x),
\end{equation}
where $G + r_1$ is proper, closed and convex, $H$ and $r_2$ are finite convex functions on
$\R^n$ (Assumption~\ref{assu:wellposed}); $H$ is
differentiable and accessed through unbiased stochastic gradients with bounded variance;
$r_1, r_2$ are deterministic. We focus on the \emph{online} regime, where $H = \E_\xi[h_\xi]$
is in principle a population average and we cannot evaluate full gradients of $H$.

The DC structure in~\eqref{eq:dc-problem} is the natural language of several model
families that come up routinely in machine learning. Before our results, we recall four
concrete instances that motivate why this problem class deserves dedicated attention.

\paragraph{Example 1 (Sparse learning with non-convex penalties).}
Many of the regularizers used to control over-shrinkage in high-dimensional regression
have a clean DC decomposition $p(t) = \lambda |t| - h_p(t)$ with $h_p$ convex and often
smooth. Three canonical examples:
\begin{itemize}[leftmargin=*]
\item \emph{Capped-}$\ell_1$ \citep{Zhang2010}: $\mathrm{cap}(t; \tau) = \min\{|t|, \tau\}$ admits the
  decomposition $\mathrm{cap}(t) = |t| - \max\{|t| - \tau, 0\}$, with both pieces convex.
\item \emph{MCP} \citep{ZhangCH2010}: with parameter $\gamma_M > 0$,
  $\mathrm{MCP}(t; \lambda, \gamma_M) = \lambda|t| - h_{\mathrm{MCP}}(t)$ where
  $h_{\mathrm{MCP}}$ is convex and $1/\gamma_M$-smooth.
\item \emph{SCAD} \citep{FanLi2001}: similarly admits a decomposition $\mathrm{SCAD}(t) = \lambda|t| - h_{\mathrm{SCAD}}(t)$
  with $h_{\mathrm{SCAD}}$ convex and $1/(a-1)$-smooth, for the SCAD parameter $a > 2$.
\end{itemize}
Penalized empirical risk minimization $\min_x \frac{1}{N}\sum_{i} \ell(a_i^\top x, y_i) + \sum_j p(x_j)$
with such a penalty falls within~\eqref{eq:dc-problem}. If $\ell$ is also nonconvex but
$L$-smooth, the loss $\ell$ itself can be DC-decomposed via the quadratic-shift trick
$\ell = (L\|\cdot\|^2/2) - (L\|\cdot\|^2/2 - \ell)$; see \citet[Eq.~(S)]{Nguyen2025}.

\paragraph{Example 2 (Robust regression with smooth robust losses).}

Losses that grow slowly in the residual are essential for robustness to outliers. The
truncated quadratic $\ell_\tau(r) = \min\{r^2, \tau^2\} = r^2 - \max\{r^2 - \tau^2, 0\}$ is DC,
but its concave part is not differentiable at $|r| = \tau$ and its sample gradients grow
linearly with the residual, so it lies \emph{outside} Assumption~\ref{assu:smooth-H}; we use it
only as an explicitly out-of-assumption stress test (Appendix~\ref{sec:experiments-regression}).
Smooth robust losses fit the framework through a curvature split. For the Cauchy loss
$\ell_c(r) = \log(1 + r^2/c^2)$ one has $\ell_c'' \in [-1/(4c^2),\, 2/c^2]$; let $h_c$ be the
even $C^2$ function with $h_c(0) = 0$ and
\[
h_c'(r) = \operatorname{sign}(r)\Bigl(\frac{1}{c} - \frac{2|r|}{c^2 + r^2}\Bigr)\ \text{ for } |r| > c,
\qquad h_c'(r) = 0\ \text{ for } |r| \le c,
\]
so that $h_c'' = \max\{-\ell_c'', 0\}$. Then $h_c$ is convex with $|h_c'| \le 1/c$ and
$0 \le h_c'' \le 1/(4c^2)$, and $g_c := \ell_c + h_c$ is convex, giving
$\ell_c = g_c - h_c$. For $\min_x \frac1N\sum_i \ell_c(y_i - a_i^\top x)$ this yields
$H(x) = \frac1N\sum_i h_c(y_i - a_i^\top x)$, which is $L$-smooth with
$L \le \max_i\|a_i\|^2/(4c^2)$ and has sample gradients bounded by $\|a_i\|/c$, so
Assumption~\ref{assu:smooth-H} holds whenever $\E\|a\|^2 < \infty$. The same construction
applies to any even $C^2$ loss with bounded derivative whose negative curvature has finite
total mass $\int_0^\infty \max\{-\ell'', 0\} < \infty$ (e.g.\ Geman--McClure). A nonconvex
sparsity penalty adds the convex correction term of its DC decomposition to $r_2$
(nonsmooth for capped-$\ell_1$, smooth for SCAD and MCP).

\paragraph{Example 3 (Positive-unlabelled (PU) learning).}
\citet{Kiryo2017} introduce the non-negative risk estimator
\[
R_{\mathrm{PU}}(g) = \pi_p R_p^+(g) + \max\bigl\{0,\; R_u(g) - \pi_p R_p^-(g)\bigr\},
\]
where $\pi_p$ is the class prior. For a model linear in its parameters and a loss convex
in the score, $R_p^\pm$ and $R_u$ are convex in the parameters \citep{Kiryo2017}. For convex
$A, B$ one has $\max\{0, A - B\} = \max\{A, B\} - B$, hence
\[
R_{\mathrm{PU}} = \Bigl[\pi_p R_p^+ + \max\bigl\{R_u,\ \pi_p R_p^-\bigr\}\Bigr] - \pi_p R_p^- ,
\]
which is~\eqref{eq:dc-problem} with convex side $G + r_1 = \pi_p R_p^+ + \max\{R_u, \pi_p R_p^-\}$
(nonsmooth through the pointwise maximum, which is handled inside the convex subproblem) and
$H = \pi_p R_p^-$, an expectation over positive samples; for a smooth loss such as the
logistic loss and bounded features, $H$ satisfies Assumption~\ref{assu:smooth-H}. The
non-negative-risk correction in PU learning is, structurally, a DC device.

\paragraph{Example 4 (Fair classification with adversarial penalties).}
\citet{Zhang2018} introduce adversarial debiasing, where a model is trained to predict the
target while an adversary tries to predict a sensitive attribute from the model's output:
\[
\min_\theta \E[\ell_{\mathrm{task}}(f_\theta(X), Y)] - \lambda\, \E[\ell_{\mathrm{adv}}(f_\theta(X), A)].
\]
In adversarial debiasing the adversary is trained jointly with the model, which gives a
min--max problem that is \emph{not} of the form~\eqref{eq:dc-problem}. The DC form arises for
a \emph{fixed} adversary (for instance a fixed probe on the model score, or one refreshed only
between outer stages): if the model is linear, $\ell_{\mathrm{task}}$ is convex in the score,
and $\ell_{\mathrm{adv}}$ is convex and smooth in the score, then $\theta \mapsto
\E[\ell_{\mathrm{task}}]$ and $\theta \mapsto \lambda\E[\ell_{\mathrm{adv}}]$ are convex and the
objective is~\eqref{eq:dc-problem} with $G = \E[\ell_{\mathrm{task}}]$,
$H = \lambda\E[\ell_{\mathrm{adv}}]$. Our guarantees apply to this fixed-adversary problem, not
to the joint adversarial game.


These problems share two features: (i) the natural decomposition $F = G - H + r_1 - r_2$ is
given by the model and is much cheaper to maintain than any global convex relaxation;
(ii) modern instances are large enough or streaming enough that \emph{stochastic} access
to $\nabla H$ is the only practical option. The deterministic difference-of-convex
algorithm (DCA) of \citet{Tao1986,Tao1997,Tao2005,LeThiPhamDinh2018},
the closely related concave-convex procedure
\citep{YuilleRangarajan2003,Sriperumbudur2009}, and structural extensions through
directional stationarity \citep{AhnPangXin2017} handle (i) but not (ii).

\paragraph{Stochastic DC: state of the art.}
\citet{Nitanda2017} introduced stochastic DCA with increasing batches; \citet{Xu2019}
extended this to nonsmooth $r_1$. Building on the SVRG \citep{JohnsonZhang2013},
SAGA \citep{Defazio2014}, SAG \citep{SchmidtLeRouxBach2017} variance-reduction line,
\citet{LeThi2022} introduced finite-sum variance-reduced DCA with $\mathcal O(N + N^{2/3}\eps^{-2})$
complexity. \citet{Phan2024} introduced majorization-minimization variants achieving
$\mathcal O(N + N^{1/2}\eps^{-2})$ for the special decomposition (S) of \citet{Nguyen2025}.
Most recently, \citet{Nguyen2025} introduced DCA-PAGE (built on the PAGE estimator of
\citet{LiBaoZhangRichtarik2021PAGE} and the SPIDER/SpiderBoost line of
\citet{Fang2018,Wang2019,NguyenSARAH2017}), attaining the finite-sum
rate $\mathcal O(N + N^{1/2}\eps^{-2})$, which matches the optimal rate for smooth nonconvex
finite-sum optimization; it extends to the online setting as
$\mathcal O(b + b^{1/2}\eps^{-2})$ \emph{using a large batch of size
$b = \Theta(\sigma^2/\eps^{2})$}, drawn with probability $p = b^{-1/2}$ in PAGE's refresh
branch. Substituting $b$ gives $\mathcal O(\sigma^2\eps^{-2} + \sigma\eps^{-3})$.
\citet{Hu2024} introduced SMAG, a single-loop algorithm for differences of max-structured
weakly convex functions that approximates the gradients of the Moreau envelopes of
\emph{both} components with one stochastic step each, under a uniformly bounded
stochastic-subgradient assumption on both components. In a true online regime, where data arrive in a
stream and per-step memory is bounded, the large refresh batches of DCA-PAGE are exactly
what we want to avoid.

\textbf{Momentum in nonconvex optimization.} Our estimators draw upon the classical foundation of Polyak heavy-ball momentum \citep{Polyak1964} alongside the rich algorithmic lineage of momentum-based variance reduction (MVR/STORM) \citep{Cutkosky2020,Levy2021,TranDinh2022,Driggs2022} and earlier variance-reduced methods for non-convex optimization \citep{Reddi2016,LanLiZhou2019,AllenZhu2018}. Momentum is a well-established mechanism for eliminating the reliance on large batch sizes, having been successfully adapted for normalized gradient descent \citep{CutkoskyMehta2020}, frameworks leveraging auxiliary information \citep{chayti2024optimizationaccessauxiliaryinformation}, and stochastic cubic Newton methods \citep{chayti2024improvingstochasticcubicnewton}. We extend both Polyak and MVR momentum to stochastic DC programming.

\paragraph{Contributions.}
\begin{itemize}[leftmargin=*]
\item We propose \textbf{momentum-based DCA} (MDCA, Algorithm~\ref{alg:mdca}), a
  single-update-per-iteration DCA variant that uses a \emph{constant} per-iteration
  batch (as small as $b = 1$); only the initial batch $b_0$ may be larger to seed the
  estimator. A stepsize $\eta_t$ damps the inner DCA step, and momentum coefficient
  $\alpha_t$ controls the estimator. Two estimator variants are studied: Polyak heavy-ball
  \citep{Polyak1964} and momentum-based variance reduction (MVR) \citep{Cutkosky2020}.
\item We prove convergence in two criteria: the \emph{critical-distance} measure
  $\E[\mathcal D(\hat x)^2] \le \eps^2$ \citep[Definition~6]{Nguyen2025}, and a
  choice-independent \emph{DCA fixed-point residual} $\E[R_{\min}(\hat x)] \le \eps^2$ that
  requires no smoothness of $r_2$ and whose zero set is exactly the set of DC-critical
  points, unlike the gap function of \citep[Definition~12]{Nguyen2025}
  (Lemmas~\ref{lem:R-criticality}--\ref{lem:R-vs-d}, Remark~\ref{rem:gap-issue}).
  With exact subproblem solves, Polyak momentum uses
  $\mathcal O(L\Delta_0\sigma^2 \eps^{-4})$ stochastic gradients of $H$ and MVR uses
  $\mathcal O(L\Delta_0\sigma\,\eps^{-3} + \sigma^3/(L\Delta_0\eps))$ under averaged
  smoothness, i.e.\ $\mathcal O(L\Delta_0\sigma\eps^{-3})$ for $\eps \le L\Delta_0/\sigma$,
  in the small-$\eps$ regime stated in the theorems. This matches the $\eps$-dependence of
  online DCA-PAGE with a constant main-loop batch
  (Theorems~\ref{thm:polyak-rate}--\ref{thm:mvr-rate-gap}); we prove no lower bound and make
  no optimality claim for MDCA itself.
\item We record how momentum enters through a \textbf{variance--drift trade-off lemma}
  (Lemma~\ref{lem:separation}). With iterate drift bounded by $\bar D$, the steady-state
  variance of the no-momentum estimator is at most $\sigma^2/b$ (with equality when the
  oracle variance equals $\sigma^2$ at every point), while Polyak and MVR admit upper bounds that the parameter
  $\alpha \in (0, 1]$ trades between variance and drift bias. The optimized bounds scale as
  $\bar D^{1/3}$ and $\bar D^{1/2}$, a heuristic counterpart of the $\eps^{-4}$ versus
  $\eps^{-3}$ upper bounds; the rate proofs do not use the lemma. For an explicit
  Gaussian instance, Proposition~\ref{prop:noise-floor} shows a non-vanishing stationarity
  floor for the undamped no-momentum method at fixed $b$.
\item We give inexact-solve guarantees (Appendix~\ref{app:inexact}): function-value accuracy
  of the subproblems suffices for the residual criterion, while the critical-distance
  criterion needs an approximate optimality (KKT) certificate.
\item We give a single-loop variant S-MDCA (Algorithm~\ref{alg:smdca}) that replaces the
  inner subproblem solve by one stochastic prox-gradient step. Both Polyak and MVR
  variants reach $\eps$-stationarity of the smoothed objective $F_\gamma$, for every fixed
  $\gamma > 0$ for which $F_\gamma$ is bounded below (Assumption~\ref{assu:Fgamma-bbel}),
  in $\mathcal O(\eps^{-4})$ outer iterations
  (Remark~\ref{rem:gamma-tight} delimits what this does and does not say about $F$); we
  explain why, in our analysis, the MVR $\eps^{-3}$ rate of the double-loop algorithm does
  not carry over to the single-loop algorithm without variance reduction on $\nabla G$
  (Remark~\ref{rem:inner-vr}).
\item Preliminary experiments on the LIBSVM \texttt{a9a} and \texttt{w8a} datasets, on
  robust+sparse classification with capped-$\ell_1$ regularization, illustrate the methods;
  a synthetic regression experiment that lies outside our assumptions is reported separately
  as a stress test.
\end{itemize}

\paragraph{Notation.}
$\|\cdot\|$ denotes the Euclidean norm. For a closed set $C$, $\dist(x, C) = \inf_{y \in C}\|x - y\|$.
$\partial f(x)$ denotes the convex subdifferential. We write $\E_t[\cdot] := \E[\cdot \mid \mathcal F_t]$
for conditional expectation given the natural filtration through iteration $t$.

\section{Setup, assumptions, and convergence criteria}
\label{sec:setup}

\subsection{Problem and assumptions}

We adopt the formulation~\eqref{eq:dc-problem} of \citet{Nguyen2025}: $G$ and $r_1$ form
the convex side, $H$ and $r_2$ the concave side. Throughout, the following standing
assumption makes the problem and the algorithms well defined.

\setcounter{assumption}{-1}
\begin{assumption}[Well-posedness]
\label{assu:wellposed}
(i) $f := G + r_1 : \R^n \to \R \cup \{+\infty\}$ is proper, closed and convex;
(ii) $H : \R^n \to \R$ is finite, convex and differentiable;
(iii) $r_2 : \R^n \to \R$ is finite and convex;
(iv) $x^0 \in \dom f$;
(v) the stochastic oracle $\nabla h_\xi(x)$ is defined for every $x \in \R^n$.
\end{assumption}

Under Assumption~\ref{assu:wellposed}, $r_2$ is continuous and $\partial r_2(x)$ is nonempty
and compact for every $x$, so every selection $w^t \in \partial r_2(x^t)$ required by the
algorithms exists; $F$ is finite on $\dom f$ and equal to $+\infty$ outside it (no
$\infty - \infty$ arises), and $F(x^0) < \infty$. The iterates of Algorithm~\ref{alg:mdca}
remain in $\dom f$: each $\tilde x^{t+1}$ minimizes a proper function with domain $\dom f$,
and $x^{t+1}$ is a convex combination of two points of the convex set $\dom f$. All concave-side
terms in our examples (convex corrections of capped-$\ell_1$, SCAD, MCP; smooth data terms;
quadratics) are finite. Extended-valued $r_2$ would require a common effective domain, a
convention for the DC difference, and an invariant set on which $\partial r_2$ is nonempty; we
do not pursue this. We do not need $r_1$ and $r_2$ to be smooth;
we may exploit smoothness of $r_2$ when it holds (Assumption~\ref{assu:smooth-r2}), and
we drop it for the fixed-point-residual results.

For any DC pair $(G, H)$ and any $\varrho > 0$, the reformulation
$F = (G + r_1 + \tfrac{\varrho}{2}\|\cdot\|^2) - (H + r_2 + \tfrac{\varrho}{2}\|\cdot\|^2)$
leaves $F$ unchanged but makes both DC components $\varrho$-strongly convex. The next
assumption is therefore without loss of generality. If the concave-side quadratic is
assigned to $H$, the oracle is shifted by the deterministic term $\varrho x$: its variance
bound $\sigma^2$ in Assumption~\ref{assu:smooth-H} is unchanged, while the smoothness constant
becomes $L + \varrho$ (in Assumptions~\ref{assu:smooth-H} and~\ref{assu:avg-smooth} alike).
It may instead be assigned to $r_2$, which keeps $r_2$ finite. For the double-loop method the
choice is immaterial, but the single-loop surrogate $F_\gamma$ depends on it
(Remark~\ref{rem:Fgamma-decomp}).

\begin{assumption}[Strong convexity]
\label{assu:sc}
$G + r_1$ is $\rho_1$-strongly convex and $H + r_2$ is $\rho_2$-strongly convex
($\rho_1, \rho_2 > 0$), with $\rho_1 \ge \rho_2$. We write $\rho := \rho_1 + \rho_2$.
\end{assumption}

The asymmetry $\rho_1 \ge \rho_2$ is without loss of generality: a $\rho_2$-strongly convex
function is $\rho'$-strongly convex for every $\rho' \le \rho_2$, so we may replace $\rho_2$
by $\min(\rho_1, \rho_2)$. It is used only to simplify the descent template
(Section~\ref{sec:descent-template}).

\begin{assumption}[Smoothness and noise on $H$]
\label{assu:smooth-H}
$H$ is $L$-smooth: $\|\nabla H(x) - \nabla H(y)\| \le L \|x - y\|$ for all $x, y$. The
stochastic-gradient oracle $\nabla h_\xi$ satisfies, for all $x$,
\[
\E_\xi[\nabla h_\xi(x)] = \nabla H(x), \qquad \E_\xi\|\nabla h_\xi(x) - \nabla H(x)\|^2 \le \sigma^2.
\]
We write $\hat\nabla H_{\mathcal I}(x) := \tfrac{1}{|\mathcal I|}\sum_{i \in \mathcal I}\nabla h_i(x)$ for
the mini-batch estimator. By independence,
$\E\|\hat\nabla H_{\mathcal I}(x) - \nabla H(x)\|^2 \le \sigma^2/|\mathcal I|$.
\end{assumption}

\begin{assumption}[Smoothness of $r_2$, optional]
\label{assu:smooth-r2}
$r_2$ is differentiable and $L_{r_2}$-smooth.
\end{assumption}

\begin{assumption}[Averaged smoothness of $H$, optional]
\label{assu:avg-smooth}
For all $x, y$, $\E_\xi\|\nabla h_\xi(x) - \nabla h_\xi(y)\|^2 \le L^2\|x - y\|^2$.
\end{assumption}

Assumption~\ref{assu:smooth-r2} is needed only for the critical-distance results; the
fixed-point-residual results (Section~\ref{sec:rates-residual}) dispense with it, as does
the descent template (Lemma~\ref{lem:descent}). Assumption~\ref{assu:avg-smooth} is used
only for the MVR rate. It is implied by per-sample $L$-smoothness
$\|\nabla h_\xi(x) - \nabla h_\xi(y)\| \le L\|x-y\|$ and, by Jensen's inequality, implies
the $L$-smoothness part of Assumption~\ref{assu:smooth-H} (with the same constant $L$).

\begin{assumption}[Boundedness from below]
\label{assu:bbel}
$F^* := \inf_x F(x) > -\infty$. We write $\Delta_0 := F(x^0) - F^*$.
\end{assumption}

The next two assumptions are used only in Section~\ref{sec:single-loop}.

\begin{assumption}[Stochastic oracle for $G$, single-loop only]
\label{assu:G-oracle}
$G$ is $L_G$-smooth and $\E_{\xi^G}[\nabla G(x;\xi^G)] = \nabla G(x)$,
$\E_{\xi^G}\|\nabla G(x;\xi^G) - \nabla G(x)\|^2 \le M^2$ for all $x$. The samples
$\{\xi^G_t\}$ are i.i.d.\ and, conditionally on the past, $\xi^G_t$ is independent of
$\mathcal I^{t+1}$. We use the filtration
$\mathcal F_t := \sigma(\mathcal I^0,\dots,\mathcal I^t,\xi^G_0,\dots,\xi^G_{t-1})$.
\end{assumption}

\begin{assumption}[Boundedness of the smoothed objective, single-loop only]
\label{assu:Fgamma-bbel}
$F_\gamma^* := \inf_x F_\gamma(x) > -\infty$, where $F_\gamma$ is defined
in~\eqref{eq:Fgamma}. We write $\Delta_\gamma := F_\gamma(x^0) - F_\gamma^*$.
\end{assumption}

\begin{lemma}[Sufficient condition for Assumption~\ref{assu:Fgamma-bbel}]
\label{lem:Fgamma-lb}
Let Assumption~\ref{assu:bbel} hold and set $\Phi_\gamma := (G+r_1)_\gamma - H - r_2$. Then
for every $x$ and every $\gamma > 0$,
\[
F_\gamma(x) \;\ge\; \Phi_\gamma(x) \;\ge\;
F^* - \frac{\gamma}{2}\Bigl(\|\nabla H(x)\| + \dist\bigl(0,\partial r_2(x)\bigr)\Bigr)^2 .
\]
In particular, if $\|\nabla H(x)\| \le M_H$ and $\dist(0,\partial r_2(x)) \le M_{r_2}$ for
\emph{all} $x \in \R^n$, then $F_\gamma^*$ and $\Phi_\gamma^*$ are both at least
$F^* - \tfrac{\gamma}{2}(M_H + M_{r_2})^2 > -\infty$. If the two bounds hold only on a set
$S$, the display gives $\inf_S F_\gamma \ge F^* - \tfrac{\gamma}{2}(M_H + M_{r_2})^2$, which
is not a bound on $F_\gamma^*$.
\end{lemma}

\begin{proof}
Write $u^* := u^*(x) = \prox_{\gamma(G+r_1)}(x)$ and $s := \|u^* - x\|$. Since
$(r_2)_\gamma \le r_2$ we have $F_\gamma \ge \Phi_\gamma$. By definition of the Moreau
envelope, $(G+r_1)_\gamma(x) = (G+r_1)(u^*) + \tfrac{1}{2\gamma}s^2$, so
\[
\Phi_\gamma(x) = (G+r_1)(u^*) + \tfrac{1}{2\gamma}s^2 - H(x) - r_2(x) .
\]
Since $F(u^*) \ge F^*$ we have $(G+r_1)(u^*) \ge F^* + H(u^*) + r_2(u^*)$, whence
\[
\Phi_\gamma(x) \;\ge\; F^* + \bigl[H(u^*) - H(x)\bigr] + \bigl[r_2(u^*) - r_2(x)\bigr] + \tfrac{1}{2\gamma}s^2 .
\]
By convexity of $H$, $H(u^*) - H(x) \ge \langle\nabla H(x), u^* - x\rangle \ge -\|\nabla H(x)\|\,s$;
by convexity of $r_2$, $r_2(u^*) - r_2(x) \ge \langle w, u^* - x\rangle \ge -\|w\|\,s$ for any
$w \in \partial r_2(x)$; take $w$ of minimal norm, which exists because $\partial r_2(x)$ is
nonempty and compact (Assumption~\ref{assu:wellposed}). Minimizing
$-(\|\nabla H(x)\| + \|w\|)s + s^2/(2\gamma)$ over $s \ge 0$ gives the claim.
\end{proof}

\noindent
Two remarks on the use of Lemma~\ref{lem:Fgamma-lb}. First, the proof of
Theorem~\ref{thm:smdca} uses $F_\gamma^*$ only through $F_\gamma(x^T) \ge F_\gamma^*$ and
$\Delta_\gamma = F_\gamma(x^0) - F_\gamma^*$; if $S$ is a deterministic set that contains all
iterates almost surely, Theorem~\ref{thm:smdca} therefore holds with $F_\gamma^*$ replaced by
$\inf_S F_\gamma$, and the local version above suffices. Such a set is not known a priori in
general. Second, the global condition cannot hold together with Assumption~\ref{assu:sc}:
since $\dist(0,\partial(H+r_2)(x)) \le \|\nabla H(x)\| + \dist(0,\partial r_2(x))$ and the
subgradients of the $\rho_2$-strongly convex function $H + r_2$ are unbounded, the two bounds
cannot both hold on $\R^n$. It is intended for the single-loop results, which do not use
Assumption~\ref{assu:sc}.

\noindent
The condition is sufficient but not necessary, and some hypothesis is genuinely needed: for
$G+r_1 = \tfrac{\rho_1}{2}\|\cdot\|^2$, $H = \tfrac{L}{2}\|\cdot\|^2$, $r_2 = 0$ with
$L < \rho_1$, the gradient $\nabla H$ is unbounded and $F_\gamma$ is bounded below exactly
when $\gamma \le 1/L - 1/\rho_1$ (cf.\ Remark~\ref{rem:gamma-tight}). Thus
Assumption~\ref{assu:Fgamma-bbel} genuinely restricts $\gamma$, and it depends on the DC
decomposition and not only on $F$ (Remark~\ref{rem:Fgamma-decomp}).

\subsection{Convergence criteria}
\label{sec:criteria}

The pointwise \emph{critical-distance residual} is \citep[Definition~6]{Nguyen2025}
\begin{equation}
\label{eq:d}
\mathcal D(x) := \dist\bigl(\nabla H(x) + \nabla r_2(x),\; \partial(G + r_1)(x)\bigr),
\end{equation}
defined when $r_2$ is differentiable; $\mathcal D(x) = 0$ characterizes critical points of
$F$. The associated criterion is $\E[\mathcal D(\hat x)^2] \le \eps^2$ for the random
output $\hat x$.

The critical-distance residual requires $r_2$ differentiable, and it is a poor measure when
$G+r_1$ is nonsmooth: for $\min_x |x|$ one has $\mathcal D(x) = 1$ at every $x \ne 0$,
including points arbitrarily close to the optimum \citep[\S V.A]{Nguyen2025}. We therefore
also work with a fixed-point residual for the exact DCA map, which needs no smoothness of
$r_2$. For $x \in \dom f$ and $w \in \partial r_2(x)$, define the exact DCA map and the
\emph{selected residual}
\begin{equation}
\label{eq:barx}
T(x, w) := \arg\min_{y \in \R^n}\Bigl\{(G+r_1)(y) - \langle \nabla H(x) + w,\, y\rangle\Bigr\},
\qquad
R(x; w) := \|T(x,w) - x\|^2 .
\end{equation}
The minimizer exists and is unique by $\rho_1$-strong convexity of $G+r_1$
(Assumption~\ref{assu:sc}). We repeatedly use the following contraction. If $\psi$ is
$\rho_1$-strongly convex and $x_a := \arg\min\{\psi - \langle a,\cdot\rangle\}$,
$x_b := \arg\min\{\psi - \langle b,\cdot\rangle\}$, then $a \in \partial\psi(x_a)$,
$b \in \partial\psi(x_b)$, and strong monotonicity of $\partial\psi$ with Cauchy--Schwarz
gives $\rho_1\|x_a-x_b\|^2 \le \langle a-b, x_a-x_b\rangle \le \|a-b\|\,\|x_a-x_b\|$, i.e.
\begin{equation}
\label{eq:argmin-contraction}
\|x_a - x_b\| \le \|a - b\|/\rho_1 .
\end{equation}

When $r_2$ is nonsmooth, $R(x; w)$ depends on the selected subgradient and is not a
function of $x$ alone (for $r_2 = \lambda|\cdot|$ at $x = 0$, different
$w \in [-\lambda, \lambda]$ give different values). We therefore also use the
\emph{choice-independent residual}
\begin{equation}
\label{eq:Rmin}
R_{\min}(x) := \min_{w \in \partial r_2(x)} R(x; w) .
\end{equation}
The minimum is attained: $\partial r_2(x)$ is nonempty and compact
(Assumption~\ref{assu:wellposed}), and $w \mapsto T(x,w)$ is $(1/\rho_1)$-Lipschitz
by~\eqref{eq:argmin-contraction}, so $w \mapsto R(x;w)$ is continuous. When $r_2$ is
differentiable, $R_{\min}(x) = R(x; \nabla r_2(x))$.

The point $T(x^t, w^t)$ is the \emph{exact} DCA step from $x^t$: the inner subproblem of
Algorithm~\ref{alg:mdca} with $g^t$ replaced by the true gradient $\nabla H(x^t)$. Along the
trajectory of Algorithm~\ref{alg:mdca} we take $w^t$ to be the subgradient the algorithm
itself selected at $x^t$ and write $\bar x^{t+1} := T(x^t, w^t)$, so that $\bar x^{t+1}$ and
$\tilde x^{t+1}$ differ only in the use of $\nabla H(x^t)$ versus $g^t$; this is what makes
Lemma~\ref{lem:R-bound} available. Since $R_{\min}(x^t) \le R(x^t; w^t)$, every bound on the
selected residual transfers to $R_{\min}$. Thus $R$ is the squared residual of the DCA
fixed-point operator,
in exact analogy with the prox-gradient mapping norm used for composite problems -- and
with the stationarity measure we report in Section~\ref{sec:experiments}. Like the gap
function of \citet[Definition~12]{Nguyen2025}, $R$ involves the true $\nabla H(x^t)$ and is
an analysis quantity, not a computable stopping criterion.

\begin{lemma}[$R_{\min}$ characterizes criticality]
\label{lem:R-criticality}
Let Assumptions~\ref{assu:wellposed} and~\ref{assu:sc} hold, and let $x \in \dom f$.
\begin{enumerate}[leftmargin=*,topsep=2pt,itemsep=2pt]
\item[(i)] If $R(x; w) = 0$ for some $w \in \partial r_2(x)$, then $x$ is a DC-critical point
of $F$, i.e.\ $\partial(G+r_1)(x) \cap \partial(H+r_2)(x) \ne \emptyset$
\citep[Definition~17]{Nguyen2025}.
\item[(ii)] Conversely, if $x$ is DC-critical then $R(x; w) = 0$ for some
$w \in \partial r_2(x)$.
\end{enumerate}
Consequently $R_{\min}(x) = 0$ if and only if $x$ is DC-critical.
\end{lemma}

\begin{proof}
(i) Optimality in \eqref{eq:barx} gives
$\nabla H(x) + w \in \partial(G+r_1)(T(x,w))$; if $T(x,w) = x$ this reads
$\nabla H(x)+w \in \partial(G+r_1)(x)$. Since $H$ is convex and differentiable and
$w \in \partial r_2(x)$, also
$\nabla H(x)+w \in \nabla H(x) + \partial r_2(x) \subseteq \partial(H+r_2)(x)$,
so the two subdifferentials intersect.
(ii) If $s \in \partial(G+r_1)(x) \cap \partial(H+r_2)(x)$, the sum rule (valid as $H$ and
$r_2$ are finite convex and $H$ is differentiable on $\R^n$) gives $s = \nabla H(x) + w$ for
some $w \in \partial r_2(x)$. Then $x$ satisfies the optimality condition of
\eqref{eq:barx}, so $T(x,w) = x$.
The final claim follows because the minimum in~\eqref{eq:Rmin} is attained.
\end{proof}

\noindent
DC-criticality is weaker than the directional stationarity of \citet{AhnPangXin2017}; it is
the notion that DCA-type methods target, and it is the one used throughout the stochastic-DC
literature we compare against.

\begin{lemma}[Comparison with the critical-distance residual]
\label{lem:R-vs-d}
Let Assumptions~\ref{assu:wellposed} and~\ref{assu:sc} hold. Then for every $x \in \dom f$
and every $w \in \partial r_2(x)$,
\[
\rho_1\sqrt{R(x; w)} \;\le\; \dist\bigl(\nabla H(x)+w,\ \partial(G+r_1)(x)\bigr).
\]
If in addition Assumptions~\ref{assu:smooth-H} and~\ref{assu:smooth-r2} hold, then
$\partial r_2(x) = \{\nabla r_2(x)\}$ and
\[
\rho_1\sqrt{R_{\min}(x)} \;\le\; \mathcal D(x),
\qquad
\mathcal D\bigl(T(x, \nabla r_2(x))\bigr) \;\le\; (L+L_{r_2})\sqrt{R_{\min}(x)} .
\]
\end{lemma}

\begin{proof}
\emph{Left inequality.} Let $s \in \partial(G+r_1)(x)$ attain
$\delta := \dist(\nabla H(x)+w, \partial(G+r_1)(x))$; the infimum is attained because
$\partial(G+r_1)(x)$ is closed and convex (if it is empty, $\delta = +\infty$ and there is
nothing to prove). Since $s \in \partial(G+r_1)(x)$, the point
$x$ minimizes $(G+r_1) - \langle s,\cdot\rangle$, and \eqref{eq:argmin-contraction} with
$a = s$, $b = \nabla H(x)+w$ gives $\|x - T(x,w)\| \le \delta/\rho_1$.
\emph{Right inequality.} Write $\bar x := T(x, \nabla r_2(x))$. By optimality in
\eqref{eq:barx}, $\nabla H(x) + \nabla r_2(x) \in \partial(G+r_1)(\bar x)$, so
\[
\mathcal D(\bar x)
\le \|\nabla H(\bar x) + \nabla r_2(\bar x) - \nabla H(x) - \nabla r_2(x)\|
\le (L+L_{r_2})\|\bar x-x\| . \qedhere
\]
\end{proof}

The two inequalities are one-step implications evaluated at \emph{different} points, not an
equivalence: a small $\mathcal D(x)$ forces a small $R_{\min}(x)$, whereas a small
$R_{\min}(x)$ forces a small critical distance at the DCA image $T(x,\nabla r_2(x))$, not at
$x$ itself. No bound of $\mathcal D(x)$ by $R_{\min}(x)$ at the same point is possible in
general: in the example below, $\mathcal D(x) = 1$ for every $x \ne 0$ while
$R_{\min}(x) \to 0$ as $x \to 0$.

On the example that motivates the gap function ($\min_x|x|$, i.e.\
$G+r_1 = |\cdot| + \tfrac{\varrho}{2}\|\cdot\|^2$ and
$H+r_2 = \tfrac{\varrho}{2}\|\cdot\|^2$ after the reformulation above, so that
$\rho_1 = \rho_2 = \varrho$), we have
$\bar x^{t+1} = \arg\min_x\{|x| + \tfrac{\varrho}{2}(x - x^t)^2\} = \prox_{|\cdot|/\varrho}(x^t)$
and $\sqrt{R_{\min}(x^t)} = \min(|x^t|, 1/\varrho) \to 0$ as $x^t \to 0$, whereas
$\mathcal D(x^t) = \dist(\varrho x^t, \operatorname{sign}(x^t) + \varrho x^t) = 1$ for every
$x^t \ne 0$. Thus $R_{\min}$ handles the nonsmooth case the gap function was introduced for,
\emph{and} characterizes criticality; the example also shows that $R_{\min}$ and
$\mathcal D$ are not equivalent at a single point.

\begin{remark}[On the gap function of \citet{Nguyen2025}]
\label{rem:gap-issue}
An earlier version of this work used the gap function
\citep[Definition~12]{Nguyen2025}, which linearizes the concave part at $x^{t-1}$:
$\mathrm{gap}(x^t) = \max_x (G+r_1)(x^t) - (G+r_1)(x) - \langle \nabla H(x^{t-1}) + w^{t-1},\, x^t - x\rangle$.
However,  it is not a stationarity measure for $F$ when $H$ is nontrivial:
$\mathrm{gap}(x^t) = 0$ certifies that $x^t$ solves the subproblem linearized at $x^{t-1}$,
whereas criticality of $F$ requires the inclusion at $x^t$; the two coincide only at a fixed
point. Concretely, with $G = x^2$, $H = x^2/2$, $r_1 = r_2 = 0$ and exact gradients, DCA
gives $x^{t+1} = x^t/2$, which minimizes the subproblem linearized at $x^t$ by
construction, so $\mathrm{gap} \equiv 0$ along the entire trajectory while
$\nabla F(x^t) = x^t \ne 0$. The failure is invisible in the example used to motivate the
measure \citep[\S V.A]{Nguyen2025}, where $H = r_2 = 0$ and the off-diagonal linearization
is vacuous. A structural symptom is that the gap admits a bound by the estimator variance
$V^t$ \emph{alone}, with no iterate-drift term -- the signature of a quantity certifying
subproblem accuracy rather than criticality. The results of \citet{Nguyen2025} resting on
this definition (their Lemma~13, Theorems~14--15, Remark~16) remain valid bounds on the gap,
but they do not by themselves establish the stationarity interpretation attached to it.
The residual \eqref{eq:barx} repairs this while retaining the property that no smoothness of
$r_2$ is required.
\end{remark}

\section{Algorithm}
\label{sec:algorithm}

\subsection{Momentum-based DCA}

\begin{algorithm}[t]
\caption{Momentum-based DCA (MDCA)}
\label{alg:mdca}
\begin{algorithmic}[1]
\Require $x^0 \in \dom(G + r_1)$ (Assumption~\ref{assu:wellposed}); momentum schedule $\{\alpha_t\} \subset (0, 1]$;
stepsize schedule $\{\eta_t\} \subset (0, 1]$; initial batch $b_0$; per-iteration
batch $b$; momentum type $\in \{\textsc{Polyak}, \textsc{MVR}\}$; iterations $T$.
\State Sample $\mathcal I^0$ of size $b_0$. Set $g^0 = \tfrac{1}{b_0}\sum_{i \in \mathcal I^0}\nabla h_i(x^0)$.
\State Choose $w^0 \in \partial r_2(x^0)$.
\For{$t = 0, 1, \dots, T - 1$}
  \State \emph{Inner subproblem (strongly convex by Assumption~\ref{assu:sc}):}
  \[
  \tilde x^{t+1} \in \arg\min_{x \in \R^n}\Big\{ G(x) + r_1(x) - \langle g^t + w^t,\, x\rangle\Big\}.
  \]
  \State \emph{Damped step:}\quad $x^{t+1} = x^t + \eta_t(\tilde x^{t+1} - x^t)$.
  \State Sample $\mathcal I^{t+1}$ of size $b$; update $g^{t+1}$:
  \begin{align*}
    \textsc{Polyak}\!:\;\;& g^{t+1} = (1 - \alpha_t)\, g^t + \alpha_t \cdot \hat\nabla H_{\mathcal I^{t+1}}(x^{t+1}). \\
    \textsc{MVR}\!:\;\;   & g^{t+1} = (1 - \alpha_t)\Big(g^t + \hat\nabla H_{\mathcal I^{t+1}}(x^{t+1}) - \hat\nabla H_{\mathcal I^{t+1}}(x^t)\Big) + \alpha_t\, \hat\nabla H_{\mathcal I^{t+1}}(x^{t+1}).
  \end{align*}
  \State $w^{t+1} \in \partial r_2(x^{t+1})$.
\EndFor
\State Draw $\tau$ uniformly from $\{0, \dots, T-1\}$.
\State \Return $\hat x := \tilde x^{\tau+1}$ (critical-distance criterion), or the pair
$(\hat x, \hat w) := (x^\tau, w^\tau)$ (residual criterion).
\end{algorithmic}
\end{algorithm}

The algorithm has three components per iteration:
\begin{enumerate}[leftmargin=*]
\item \emph{Inner subproblem:} solve a strongly convex problem in which $\nabla H(x^t)$
  has been replaced by an estimator $g^t$. By $\rho_1$-strong convexity of $G + r_1$
  (Assumption~\ref{assu:sc}), the minimizer $\tilde x^{t+1}$ is unique and well-defined.
\item \emph{Damped step:} take a fraction $\eta_t \in (0, 1]$ of the way from $x^t$ to
  $\tilde x^{t+1}$. With $\eta_t = 1$, this is a standard DCA step.
\item \emph{Momentum update:} update $g^{t+1}$ using a fresh batch of size $b$.
\end{enumerate}

For the critical-distance criterion the output is $\tilde x^{t+1}$ rather than $x^t$: the
reason is that $g^t + w^t \in \partial(G + r_1)(\tilde x^{t+1})$ by the optimality of
$\tilde x^{t+1}$, so the critical-distance bound is most naturally stated at
$\tilde x^{t+1}$. The residual is instead evaluated at $x^t$, with the algorithm's own
subgradient $w^t$. Returning the pair $(\hat x, \hat w)$ makes the selected residual
$R(\hat x; \hat w)$ well defined, and the choice-independent
$R_{\min}(\hat x) \le R(\hat x; \hat w)$ is controlled as well.

\subsection{Discussion of the algorithm}

\paragraph{Why a stepsize $\eta_t$.} Without damping ($\eta_t \equiv 1$), the iterate
jumps to the full inner solution $\tilde x^{t+1}$. The variance recursion
(Section~\ref{sec:variance}) has a bias term proportional to $L^2 \|x^{t+1} - x^t\|^2$,
which equals $L^2 \|\tilde x^{t+1} - x^t\|^2$ in this case. We have no a~priori control on
$\|\tilde x^{t+1} - x^t\|^2$ at constant per-iteration sample budget $b$: the inner
subproblem's solution can move by an amount comparable to $\|g^t - \nabla H(x^t)\|/\rho_1$,
which is itself $\Theta(\sigma/\sqrt b)$ for vanilla SGD-style estimators. Damping by
$\eta_t < 1$ shrinks $\|x^{t+1} - x^t\|^2$ by $\eta_t^2$ and, in our analysis, allows the
choice of $\eta_t$ in terms of $\alpha$ that makes the bias term in the variance recursion
controlled by the descent term.

\paragraph{Constant batch.} The per-iteration batch size $b$ is constant: it does not grow
with $T$ or with $\eps$. Only the initial batch $b_0$ seeds $g^0$, entering the analysis
through $V^0 = \E\|g^0 - \nabla H(x^0)\|^2 \le \sigma^2/b_0$. For Polyak momentum $V^0$
enters only additively with $L\Delta_0$, so $b_0 = 1$ suffices
(Theorem~\ref{thm:polyak-rate}). For MVR a single warm-up batch of size
$b_0 = \Theta(\eps^{-1})$ is required (Theorem~\ref{thm:mvr-rate},
Eq.~\eqref{eq:mvr-b0}); this is standard in the analysis of momentum-based variance
reduction, it is paid once rather than periodically, and for $\eps \le L\Delta_0/\sigma$ it is
dominated by the $\Theta(\eps^{-3})$ cost of the main loop, so the total complexity is
unaffected.

\paragraph{Inner-subproblem solves.} For ease of analysis the algorithm assumes exact
solves of the inner convex problem
$\Psi^t(y) := G(y) + r_1(y) - \langle g^t + w^t, y\rangle$. Appendix~\ref{app:inexact}
treats two kinds of inexactness, in which line~5 is applied to the returned point
$\hat x^{t+1}$.
\emph{(a) Function-value accuracy,}
$\Psi^t(\hat x^{t+1}) - \min_y \Psi^t(y) \le \nu_t$. By $\rho_1$-strong convexity,
$\|\hat x^{t+1} - \tilde x^{t+1}\| \le \sqrt{2\nu_t/\rho_1}$. This suffices for the residual
criterion: the descent template holds with $\rho$ replaced by $\rho/2$ and an additive
$2\eta_t\nu_t$ (Lemma~\ref{lem:inexact-descent}), and Theorem~\ref{thm:polyak-rate-gap} holds
up to constants with an additional term $72\bar\nu/\rho_1$, $\bar\nu := \tfrac1T\sum_t\E\nu_t$
(Theorem~\ref{thm:inexact-residual}), so $\bar\nu \lesssim \rho_1\eps^2$ preserves the rate. It
does \emph{not} suffice for the critical-distance criterion: when $G + r_1$ is nonsmooth, a
point can have arbitrarily small function gap while $g^t + w^t$ stays at distance bounded
away from zero from its subdifferential. (For $\Psi(y) = |y| + \tfrac{\varrho}{2}y^2$ with
minimizer $0$, the point $y = \delta > 0$ has gap $\delta + \varrho\delta^2/2 \to 0$, but
$\dist(0, \partial\Psi(\delta)) = 1 + \varrho\delta$.)
\emph{(b) Approximate optimality certificate:} the solver returns
$(\hat x^{t+1}, s^{t+1})$ with $s^{t+1} \in \partial(G+r_1)(\hat x^{t+1})$ and
$\|s^{t+1} - (g^t + w^t)\| \le \delta_t$. This implies (a) with $\nu_t = \delta_t^2/(2\rho_1)$
and gives $\mathcal D(\hat x^{t+1})^2 \le 3(L+L_{r_2})^2\|\hat x^{t+1}-x^t\|^2 + 3\|e^t\|^2 +
3\delta_t^2$, so Theorem~\ref{thm:polyak-rate} holds up to constants with an additional
$\mathcal O\bigl((1 + (L+L_{r_2})^2/\rho_1^2)\,\bar\delta^2\bigr)$
(Theorem~\ref{thm:inexact-crit}). Such a certificate is computable when $G$ is
$L_G$-smooth and $\prox_{r_1}$ is available: a proximal-gradient step
$y^+ = \prox_{s r_1}\bigl(y - s(\nabla G(y) - g^t - w^t)\bigr)$ yields
$s^+ := \nabla G(y^+) - \nabla G(y) + (y - y^+)/s + g^t + w^t \in \partial(G+r_1)(y^+)$ with
$\|s^+ - (g^t + w^t)\| \le (L_G + 1/s)\|y - y^+\|$.

\emph{Solver cost.} Strong convexity alone does not provide a linearly convergent inner
solver (with only a subgradient oracle, for instance, it does not). If $G$ is $L_G$-smooth and
$\mu_G$-strongly convex with $\mu_G > 0$ (e.g.\ by assigning the quadratic of the
reformulation above to $G$) and $\prox_{r_1}$ is available, proximal gradient with step
$1/L_G$ contracts the squared distance to $\tilde x^{t+1}$ by $1 - \mu_G/L_G$ per step, so
certificate accuracy $\delta_t$ is reached in $\mathcal O((L_G/\mu_G)\log(1/\delta_t))$
deterministic inner iterations; we claim logarithmic inner cost only under such structure.
When $G$ is itself an expectation accessed through stochastic gradients, as in
Section~\ref{sec:experiments}, only route (a) is available, in expectation: standard
guarantees for (proximal) SGD on strongly convex problems give $\E\nu_t = \mathcal O(\sigma_G^2/(\rho_1 k))$
after $k$ inner steps under smoothness of $G$, where $\sigma_G^2$ bounds the variance of the
stochastic gradients of $G$, i.e.\ $\mathcal O(\sigma_G^2/(\rho_1^2\eps^2))$ stochastic
gradients per outer step for $\E\nu_t \lesssim \rho_1\eps^2$. Removing this cost is the
motivation for the single-loop variant of Section~\ref{sec:single-loop}.

\paragraph{Connection to existing methods.}
\begin{itemize}[leftmargin=*]
\item \emph{$\alpha_t \equiv 1$, $\eta_t \equiv 1$:} stochastic DCA with a constant batch,
  the constant-batch analogue of the SDCA of \citet{Nitanda2017} (which uses increasing
  batches). The estimator variance is at most $\sigma^2/b$ at every step; at fixed $b$ the
  method can have a non-vanishing stationarity floor (Proposition~\ref{prop:noise-floor}),
  and the standard remedy is $b = \Theta(\sigma^2/\eps^2)$, for a total cost
  $\mathcal O(\sigma^2/\eps^4)$.
\item \emph{DCA-PAGE} \citep{Nguyen2025} is \emph{not} a special case of
  Algorithm~\ref{alg:mdca}. PAGE replaces the estimator by a fresh large-batch estimate with
  probability $p$ and otherwise applies an unweighted same-sample difference correction
  $g^{t+1} = g^t + \hat\nabla H_{\mathcal I'}(x^{t+1}) - \hat\nabla H_{\mathcal I'}(x^t)$;
  MVR applies a $(1-\alpha_t)$-weighted correction plus an $\alpha_t$-weighted fresh estimate
  at every step. The two estimators share the use of same-sample gradient differences, and
  formally, MVR with $\alpha_t = 0$ coincides with PAGE's no-refresh branch; PAGE's refresh branch has no
  counterpart in Algorithm~\ref{alg:mdca}.
\item \emph{SMAG} \citep{Hu2024} is related to but \emph{not} equivalent to the single-loop
  variant of Section~\ref{sec:single-loop}, including at $\alpha = 1$: SMAG maintains
  stochastic approximations of the proximal points of both DC components, whereas S-MDCA
  computes $\prox_{\gamma r_2}$ exactly, tracks only the proximal point of $G + r_1$, and
  estimates $\nabla H$ separately by momentum. S-MDCA builds on SMAG's single-step tracking
  of the first proximal point; because $H$ is smooth here, it need not be smoothed.
\end{itemize}
The distinguishing features of MDCA are the damped step and the constant main-loop batch
with a single warm-up batch; with these, the MVR variant matches the $\eps$-dependence of
online DCA-PAGE without periodic large batches.

\begin{remark}[Relation to stochastic proximal-point methods]
\label{rem:spp}
The difficulty addressed by the damping $\eta_t$ is not specific to DCA. Any method whose
update is an implicit $\arg\min$ rather than an explicit linear step -- stochastic
proximal-point, model-based, and prox-linear methods among them -- can have iterate drift
of order $\sigma/\sqrt b$ governed by the estimator noise, since the subproblem solution
can move by up to $\|g^t - \nabla H(x^t)\|/\rho_1$. Momentum-based variance reduction
requires small drift in order to cancel noise, so the two requirements conflict. Damping the
implicit step is a general mechanism for resolving this, and we expect the analysis of
Section~\ref{sec:analysis} to transfer to those settings.
\end{remark}

\section{Analysis: variance, descent, and the role of momentum}
\label{sec:analysis}

We now develop the convergence analysis. Section~\ref{sec:variance} gives the variance
recursions for the two estimators. Section~\ref{sec:separation} records the resulting
variance--drift trade-off in steady-state upper bounds. Section~\ref{sec:descent-template}
establishes the descent template, and Sections~\ref{sec:rates}--\ref{sec:rates-residual}
combine these into rates.

We use throughout the notation
\[
e^t := g^t - \nabla H(x^t),\quad
V^t := \E\|e^t\|^2,\quad
\tilde D^t := \E\|\tilde x^{t+1} - x^t\|^2,\quad
D^t := \E\|x^{t+1} - x^t\|^2 = \eta_t^2\, \tilde D^t.
\]

\subsection{Variance recursions}
\label{sec:variance}

The first lemma bounds $V^{t+1}$ in terms of $V^t$, $D^t$ and $\sigma^2/b$ for the Polyak
estimator.

\begin{lemma}[Polyak variance recursion]
\label{lem:polyak-var}
Under Assumption~\ref{assu:smooth-H}, the Polyak update of $g^{t+1}$ in
Algorithm~\ref{alg:mdca} satisfies, for all $\alpha_t \in (0, 1]$,
\begin{equation}
\label{eq:polyak-var-rec}
V^{t+1} \;\le\; (1 - \alpha_t)\, V^t \;+\; \frac{L^2}{\alpha_t}\, D^t \;+\; \alpha_t^2\, \frac{\sigma^2}{b}.
\end{equation}
\end{lemma}

\begin{proof}[Proof sketch]
Let $\zeta^{t+1} = \hat\nabla H_{\mathcal I^{t+1}}(x^{t+1}) - \nabla H(x^{t+1})$, with
$\E_t[\zeta^{t+1}] = 0$ and $\E_t\|\zeta^{t+1}\|^2 \le \sigma^2/b$. The Polyak update gives
$e^{t+1} = (1-\alpha_t)[e^t + \nabla H(x^t) - \nabla H(x^{t+1})] + \alpha_t\zeta^{t+1}$.
By bias--variance with $\E_t[\zeta^{t+1}] = 0$, then Young's inequality with parameter
$\theta = \alpha_t/(1-\alpha_t)$ on the deterministic part, then $L$-smoothness of $H$, we
obtain~\eqref{eq:polyak-var-rec}. Full algebra in Appendix~\ref{app:lem-polyak-proof}.
\end{proof}

\begin{lemma}[MVR variance recursion]
\label{lem:mvr-var}
Under Assumptions~\ref{assu:smooth-H} and \ref{assu:avg-smooth}, the MVR update of $g^{t+1}$
in Algorithm~\ref{alg:mdca} satisfies, for all $\alpha_t \in (0, 1]$,
\begin{equation}
\label{eq:mvr-var-rec}
V^{t+1} \;\le\; (1 - \alpha_t)\, V^t \;+\; \frac{2 L^2}{b}\, D^t \;+\; \frac{2 \alpha_t^2 \sigma^2}{b}.
\end{equation}
\end{lemma}

\begin{proof}[Proof sketch]
Let $\xi^{t+1}(x) = \hat\nabla H_{\mathcal I^{t+1}}(x) - \nabla H(x)$. By Assumption~\ref{assu:avg-smooth}
and the iid mini-batch identity, $\E_t\|\xi^{t+1}(x) - \xi^{t+1}(y)\|^2 \le L^2 \|x-y\|^2/b$.
The MVR update is $e^{t+1} = (1-\alpha_t) e^t + \Xi_t$ with $\Xi_t = (1-\alpha_t)[\xi^{t+1}(x^{t+1}) - \xi^{t+1}(x^t)] + \alpha_t \xi^{t+1}(x^{t+1})$,
zero-mean given $\mathcal F_t$. Bias--variance plus $\|a + c\|^2 \le 2\|a\|^2 + 2\|c\|^2$ gives
$\E_t\|\Xi_t\|^2 \le 2L^2\|x^{t+1}-x^t\|^2/b + 2\alpha_t^2\sigma^2/b$, and
$(1-\alpha_t)^2 \le (1-\alpha_t)$ on the deterministic part yields~\eqref{eq:mvr-var-rec}. Full
algebra in Appendix~\ref{app:lem-mvr-proof}.
\end{proof}

\subsection{Variance--drift trade-off}
\label{sec:separation}

The two variance recursions, side by side, show how the momentum parameter trades
estimator variance against bias from iterate drift. The following lemma records the
resulting steady-state \emph{upper bounds}. It is not used in the proofs of the rate
theorems.

\begin{lemma}[Variance--drift trade-off]
\label{lem:separation}
Fix $b \ge 1$, run Algorithm~\ref{alg:mdca} with constant $\alpha$ and $\eta$, and suppose the
iterate drift is bounded: $D^t \le \bar D$ for all $t$. Let
$V_\infty := \limsup_{t \to \infty} V^t$. Then:
\begin{itemize}[leftmargin=*]
\item \emph{(No momentum, $\alpha = 1$):}\quad $V^t \le \sigma^2/b$ for every $t \ge 1$,
whatever $\bar D$; in particular $V_\infty \le \sigma^2/b$.
\item \emph{(Polyak, any $\alpha \in (0, 1]$):}\quad
$V_\infty \le \Phi_{\mathrm P}(\alpha) := \dfrac{\alpha\, \sigma^2}{b} + \dfrac{L^2 \bar D}{\alpha^2}$.
\item \emph{(MVR, any $\alpha \in (0, 1]$, under Assumption~\ref{assu:avg-smooth}):}\quad
$V_\infty \le \Phi_{\mathrm M}(\alpha) := \dfrac{2\alpha\, \sigma^2}{b} + \dfrac{2 L^2 \bar D}{\alpha\, b}$.
\end{itemize}
Let $\sigma > 0$ and put $\alpha_{\mathrm P} := (2L^2 b\bar D/\sigma^2)^{1/3}$ and
$\alpha_{\mathrm M} := L\sqrt{\bar D}/\sigma$. Over $\alpha \in (0,1]$,
\[
\inf_{\alpha}\Phi_{\mathrm P} =
\begin{cases}
\tfrac{3}{2^{2/3}}\,L^{2/3}\sigma^{4/3}\, b^{-2/3}\, \bar D^{1/3} & \text{if } \alpha_{\mathrm P} \le 1,\\[2pt]
\sigma^2/b + L^2\bar D \ \ (\text{at } \alpha = 1) & \text{if } \alpha_{\mathrm P} > 1,
\end{cases}
\qquad
\inf_{\alpha}\Phi_{\mathrm M} =
\begin{cases}
4\, L\sigma\, b^{-1}\, \bar D^{1/2} & \text{if } \alpha_{\mathrm M} \le 1,\\[2pt]
2(\sigma^2 + L^2\bar D)/b \ \ (\text{at } \alpha = 1) & \text{if } \alpha_{\mathrm M} > 1.
\end{cases}
\]
The infima are attained at $\min\{1, \alpha_{\mathrm P}\}$ and $\min\{1,\alpha_{\mathrm M}\}$
when these are positive; if $L\bar D = 0$ the infimum $0$ is approached as $\alpha \downarrow 0$
but not attained. (If $\sigma = 0$, both bounds are minimized at $\alpha = 1$.)
\end{lemma}

\begin{proof}
\textbf{Comparison principle.} If $v_{t+1} \le (1-\alpha)v_t + c$ with $\alpha \in (0,1]$ and
$c \ge 0$, then by induction $v_t \le (1-\alpha)^t v_0 + c/\alpha$, so
$\limsup_t v_t \le c/\alpha$.

\textbf{No momentum.} With $\alpha = 1$, the Polyak update gives
$g^{t+1} = \hat\nabla H_{\mathcal I^{t+1}}(x^{t+1})$, so $e^{t+1} = \zeta^{t+1}$ and
$V^{t+1} = \E\,\E_t\|\zeta^{t+1}\|^2 \le \sigma^2/b$ by Assumption~\ref{assu:smooth-H}.

\textbf{Polyak and MVR bounds.} Apply the comparison principle to~\eqref{eq:polyak-var-rec}
with $c = L^2\bar D/\alpha + \alpha^2\sigma^2/b$, and to~\eqref{eq:mvr-var-rec} with
$c = 2L^2\bar D/b + 2\alpha^2\sigma^2/b$, using $D^t \le \bar D$.

\textbf{Minimization.} $\Phi_{\mathrm P}$ is strictly convex on $(0,\infty)$ when $L\bar D > 0$,
with $\Phi_{\mathrm P}'(\alpha) = \sigma^2/b - 2L^2\bar D/\alpha^3$ vanishing only at
$\alpha_{\mathrm P}$. If $\alpha_{\mathrm P} \le 1$, then using
$L^2\bar D/\alpha_{\mathrm P}^2 = \alpha_{\mathrm P}\sigma^2/(2b)$,
$\Phi_{\mathrm P}(\alpha_{\mathrm P}) = \tfrac32\alpha_{\mathrm P}\sigma^2/b
= \tfrac{3}{2^{2/3}}L^{2/3}\sigma^{4/3}b^{-2/3}\bar D^{1/3}$; if $\alpha_{\mathrm P} > 1$,
$\Phi_{\mathrm P}$ is decreasing on $(0,1]$ and the minimum is $\Phi_{\mathrm P}(1)$.
Likewise $\Phi_{\mathrm M}'(\alpha) = 2\sigma^2/b - 2L^2\bar D/(\alpha^2 b)$ vanishes only at
$\alpha_{\mathrm M}$, where $\Phi_{\mathrm M}(\alpha_{\mathrm M}) = 4L\sigma\sqrt{\bar D}/b$;
otherwise the minimum over $(0,1]$ is $\Phi_{\mathrm M}(1)$. If $L\bar D = 0$, both functions
are increasing and tend to $0$ as $\alpha \downarrow 0$.
\end{proof}

\begin{remark}[When the no-momentum bound is attained]
\label{rem:nomom-equality}
Assumption~\ref{assu:smooth-H} only bounds the oracle variance from above, so for a general
oracle nothing prevents $V^t < \sigma^2/b$. If the noise is state independent with
$\E_\xi\|\nabla h_\xi(x) - \nabla H(x)\|^2 = \sigma^2$ for all $x$ (e.g.\ additive noise, as in
Proposition~\ref{prop:noise-floor}), then the no-momentum estimator has $V^t = \sigma^2/b$
exactly for every $t \ge 1$, independently of the drift. For that instance,
Proposition~\ref{prop:noise-floor} shows that the undamped no-momentum method has a
stationarity floor bounded away from zero at fixed $b$; this, not
Lemma~\ref{lem:separation}, is our existence-of-a-floor statement.
\end{remark}

\begin{remark}[How to read Lemma~\ref{lem:separation}]
\label{rem:reading-separation}
The lemma describes a trade-off between upper bounds, not a separation. At fixed $b$ the
no-momentum bound has no tunable parameter. For Polyak and MVR, the bounds can be made small
by driving $\alpha$ and $\bar D$ to zero \emph{jointly} ($\bar D$ is controlled by the
stepsize $\eta$); for fixed $\bar D > 0$, decreasing $\alpha$ below $\alpha_{\mathrm P}$ or
$\alpha_{\mathrm M}$ \emph{increases} the bound. The optimized bounds scale as
$\bar D^{1/3}$ for Polyak and $\bar D^{1/2}$ for MVR; this difference is a heuristic
counterpart of the different $\eps$-dependence of Theorems~\ref{thm:polyak-rate}
and~\ref{thm:mvr-rate}, whose proofs do not use the lemma. Because only upper bounds are
compared, the lemma shows neither that MVR is faster than Polyak on a given instance, nor
that momentum is necessary, nor that MDCA is optimal.
\end{remark}

\begin{remark}[What we are \emph{not} claiming]
\label{rem:not-claiming}
Lemma~\ref{lem:separation} concerns \emph{the variance of the specific estimator}
produced by each update rule, with iterate drift held fixed at $\bar D$. It does not say
that no-momentum cannot reach $\eps$-stationarity: an SGD-style method with a
\emph{decreasing} stepsize sequence (so that the drift $\bar D$ itself goes to zero) does
drive $\E\|\nabla F\|$ to zero \citep{Ghadimi2016}. Likewise, increasing $b$ proportionally
to $1/\eps^2$ also reaches $\eps$-stationarity for vanilla SGD, at total cost
$\mathcal O(\sigma^2/\eps^4)$ stochastic gradients -- the same $\eps$-dependence we obtain for
Polyak with constant $b$. The lemma is not an ingredient of the proof of
Theorem~\ref{thm:mvr-rate}, and it implies no lower bound for any method.
\end{remark}

\subsection{Descent template}
\label{sec:descent-template}

The descent template adapts \citet[Lemma~7]{Nguyen2025} to inexact gradients of $H$ \emph{and}
to the damped step $x^{t+1} = x^t + \eta_t(\tilde x^{t+1} - x^t)$.

\begin{lemma}[Descent inequality, MDCA]
\label{lem:descent}
Suppose Assumption~\ref{assu:sc} holds and $H$ is convex and differentiable. For all
$\eta_t \in (0, 1]$, any $\beta_t \in (0, \rho)$, and any choice
$w^t \in \partial r_2(x^t)$ in Algorithm~\ref{alg:mdca},
\begin{equation}
\label{eq:descent-template}
\E\bigl[F(x^{t+1}) - F(x^t)\bigr]
\;\le\;
\frac{\eta_t}{2\beta_t}\, V^t
\;-\;
\frac{\eta_t(\rho - \beta_t)}{2}\, \tilde D^t.
\end{equation}
Neither Assumption~\ref{assu:smooth-H} nor Assumption~\ref{assu:smooth-r2} is needed.
\end{lemma}

\begin{proof}[Proof sketch]
By $\rho_2$-strong convexity of $H + r_2$ at $x^t$ with the subgradient
$\nabla H(x^t) + w^t \in \partial(H+r_2)(x^t)$, evaluated at $\tilde x^{t+1}$, together with
the optimality $g^t + w^t \in \partial(G + r_1)(\tilde x^{t+1})$ and $\rho_1$-strong
convexity of $G + r_1$,
\[
F(\tilde x^{t+1}) - F(x^t) \le \langle g^t - \nabla H(x^t), \tilde x^{t+1} - x^t\rangle - \tfrac{\rho}{2}\|\tilde x^{t+1} - x^t\|^2,
\]
the $w^t$ terms cancelling exactly between the two strong-convexity inequalities, since the
same subgradient appears in both. To extend to
$x^{t+1} = x^t + \eta_t(\tilde x^{t+1} - x^t)$, we use the same strong-convexity inequalities
at $x^{t+1}$, scaling each occurrence of $\|\tilde x^{t+1} - x^t\|^2$ by the appropriate power
of $\eta_t$ from the convex combination, plus the standard ``three-point'' inequality
for strongly convex functions. The curvature coefficient becomes $\eta_t[\rho_1(2-\eta_t) + \eta_t\rho_2]/2$
which is bounded below by $\eta_t\rho/2$ thanks to $\rho_1 \ge \rho_2$
(Assumption~\ref{assu:sc}) since
$\rho_1(2-\eta_t) + \eta_t\rho_2 - \rho = (1-\eta_t)(\rho_1 - \rho_2) \ge 0$ for $\eta_t \le 1$.
Young's inequality with parameter $\beta_t$ then gives~\eqref{eq:descent-template}.
Full algebra in Appendix~\ref{app:lem-descent-proof}.
\end{proof}

\subsection{Convergence rates: critical-distance criterion}
\label{sec:rates}

We now combine the descent template (Lemma~\ref{lem:descent}) with the variance recursions
(Lemmas~\ref{lem:polyak-var} and \ref{lem:mvr-var}) to obtain rates in the
critical-distance criterion.

\begin{lemma}[Distance--variance bound]
\label{lem:dist-bound}
Under Assumptions~\ref{assu:smooth-H} and \ref{assu:smooth-r2}, with $w^t = \nabla r_2(x^t)$,
\begin{equation}
\label{eq:dist-vd}
\E\bigl[\mathcal D(\tilde x^{t+1})^2\bigr] \;\le\; 2\,(L + L_{r_2})^2\, \tilde D^t \;+\; 2\, V^t.
\end{equation}
\end{lemma}

\begin{proof}
By the optimality condition established in the proof of Lemma~\ref{lem:descent},
$g^t + w^t \in \partial(G + r_1)(\tilde x^{t+1})$, so
$\dist(\nabla H(\tilde x^{t+1}) + \nabla r_2(\tilde x^{t+1}),\, \partial(G + r_1)(\tilde x^{t+1}))
\le \|\nabla H(\tilde x^{t+1}) + \nabla r_2(\tilde x^{t+1}) - g^t - w^t\|$. Triangle inequality,
$w^t = \nabla r_2(x^t)$, and Assumptions~\ref{assu:smooth-H}--\ref{assu:smooth-r2} give
\begin{align*}
\|\nabla H(\tilde x^{t+1}) + \nabla r_2(\tilde x^{t+1}) - g^t - w^t\|
&\le \|\nabla H(\tilde x^{t+1}) - \nabla H(x^t)\| + \|\nabla H(x^t) - g^t\| \\
&\quad + \|\nabla r_2(\tilde x^{t+1}) - \nabla r_2(x^t)\| \\
&\le (L + L_{r_2})\,\|\tilde x^{t+1} - x^t\| + \|e^t\|.
\end{align*}
Squaring and using $(a + c)^2 \le 2a^2 + 2c^2$, then taking expectations, gives~\eqref{eq:dist-vd}.
\end{proof}

\begin{theorem}[Polyak rate, critical distance]
\label{thm:polyak-rate}
Let Assumptions~\ref{assu:sc}, \ref{assu:smooth-H}, \ref{assu:smooth-r2}, \ref{assu:bbel}
hold. Run Algorithm~\ref{alg:mdca} with the Polyak update for $T$ iterations, with constants
\begin{equation}
\label{eq:polyak-params}
\beta_t \equiv \rho/2, \qquad
\eta_t \equiv \eta := \min\Bigl\{1,\ \frac{\alpha\,\rho}{4 L}\Bigr\}, \qquad
\alpha_t \equiv \alpha,
\end{equation}
where $\alpha \in (0, 1]$ is a tuning parameter chosen below. Then
\begin{equation}
\label{eq:polyak-davg-bound}
\frac{1}{T} \sum_{t=0}^{T-1} \E\!\big[\mathcal D(\tilde x^{t+1})^2\big]
\;\le\;
\underbrace{\frac{C_1\,(V^0 + L\Delta_0)}{\alpha\, T}
+ \frac{8(L+L_{r_2})^2\Delta_0}{\rho\,T}}_{\text{transient}}
\;+\;
\underbrace{\frac{C_2\, \alpha\, \sigma^2}{b}}_{\text{noise floor}}
\end{equation}
where
\[
C_1 := \frac{128}{3}\,\frac{(L + L_{r_2})^2}{\rho^2} + \frac{8}{3},
\qquad
C_2 := \frac{32}{3}\,\frac{(L + L_{r_2})^2}{\rho^2} + \frac{8}{3}
\]
are constants depending only on $L$, $L_{r_2}$ and $\rho$. The additional
$\mathcal O(\Delta_0/(\rho T))$ term arises from the cap on $\eta$; it is of order $1/T$ and
does not affect the $\eps$-dependence.
Write $A := V^0 + L\Delta_0$ and choose $\alpha := \min\{1, \alpha^*\}$ with
$\alpha^* := \sqrt{C_1 bA/(C_2 T\sigma^2)}$ (and $\alpha := 1$ if $\sigma = 0$). Then, whether
$\alpha^* \le 1$ or $\alpha^* > 1$,
\begin{equation}
\label{eq:polyak-two-regime}
\frac{1}{T}\sum_{t=0}^{T-1}\E\!\big[\mathcal D(\tilde x^{t+1})^2\big]
\;\le\;
2\sqrt{\frac{C_1C_2\,A\,\sigma^2}{b\,T}} \;+\; \frac{2C_1A}{T} \;+\;
\frac{8(L + L_{r_2})^2\,\Delta_0}{\rho\,T}.
\end{equation}
Hence $\E[\mathcal D(\hat x)^2] \le \eps^2$ as soon as
\[
T \;\ge\; \max\Bigl\{\frac{36\,C_1C_2\,A\,\sigma^2}{b\,\eps^4},\ \frac{6C_1A}{\eps^2},\
\frac{24(L+L_{r_2})^2\Delta_0}{\rho\,\eps^2}\Bigr\},
\]
and in the small-$\eps$ regime where the first term of the maximum dominates (which
requires in particular $\eps^2 \le 6C_2\sigma^2/b$),
$T = \mathcal O\bigl(C_1C_2 A\sigma^2/(b\eps^4)\bigr)$.
Since $V^0$ enters only through $A$, no $\eps$-dependent seeding is needed: with $b_0 = 1$
we have $V^0 \le \sigma^2$ and, in this regime, a total stochastic-gradient cost
$T b + b_0 = \mathcal O\bigl(C_1C_2\sigma^2(\sigma^2 + L\Delta_0)/\eps^4\bigr)$ with $b \ge 1$
\emph{constant}. If $L\Delta_0 > 0$, the constant $b_0 = \lceil \sigma^2/(L\Delta_0)\rceil$
gives $V^0 \le L\Delta_0$ and the cleaner $\mathcal O(C_1C_2L\Delta_0\sigma^2/\eps^4)$.
\end{theorem}

\begin{proof}[Proof sketch]
We sketch the seven main steps; full bookkeeping is in Appendix~\ref{app:thm-polyak-proof}.

\textbf{Step 1 (telescope descent).} From Lemma~\ref{lem:descent} with $\beta_t = \rho/2$ and
the relation $D^t = \eta^2\tilde D^t$, summing from $t = 0$ to $T - 1$ and rearranging,
$\sum \tilde D^t \le 4\Delta_0/(\eta\rho) + (4/\rho^2)\sum V^t$, and
$1/\eta \le 1 + 4L/(\alpha\rho)$ by the cap.

\textbf{Step 2 (telescope variance, Polyak).} From Lemma~\ref{lem:polyak-var} with $\alpha_t = \alpha$,
summing and rearranging, $\alpha\sum V^t \le V^0 + (L^2\eta^2/\alpha)\sum\tilde D^t + T\alpha^2\sigma^2/b$.

\textbf{Step 3 (eliminate $\sum\tilde D^t$).} Substituting Step~1 into Step~2 and using
$\eta \le \alpha\rho/(4L)$, the coefficient of $\sum V^t$ on the left becomes $\alpha - \alpha/4 = 3\alpha/4$,
giving $\sum V^t \le (4/3)(V^0 + L\Delta_0)/\alpha + (4/3) T\alpha\sigma^2/b$.

\textbf{Steps 4--5 (convert to $\mathcal D^2$).} Substituting back into Step~1 gives a bound on
$\sum\tilde D^t$, and Lemma~\ref{lem:dist-bound} converts to $\sum\E[\mathcal D(\tilde x^{t+1})^2]
\le 2(L+L_{r_2})^2\sum\tilde D^t + 2\sum V^t$.

\textbf{Step 6 (choose $\alpha$).} Up to the $\mathcal O(\Delta_0/(\rho T))$ term the bound
is $A_1/(\alpha T) + A_2\alpha\sigma^2/b$ with $A_1 = C_1(V^0 + L\Delta_0)$ and
$A_2 = C_2$. If $\alpha^* = \sqrt{A_1 b/(A_2\sigma^2 T)} \le 1$ it is minimized at
$2\sqrt{A_1 A_2}\,\sigma\, b^{-1/2}T^{-1/2}$; otherwise $\alpha = 1$ gives
$A_1/T + A_2\sigma^2/b < 2A_1/T$, which is~\eqref{eq:polyak-two-regime}.

\textbf{Step 7 (sample complexity).} Setting the bound to $\eps^2$ and taking $b_0 = 1$ gives
the stated $T$ and total cost.
\end{proof}

\begin{theorem}[MVR rate, critical distance]
\label{thm:mvr-rate}
Under additionally Assumption~\ref{assu:avg-smooth}, run Algorithm~\ref{alg:mdca} with MVR
and constants
\begin{equation}
\label{eq:mvr-params}
\beta_t \equiv \rho/2, \qquad
\eta_t \equiv \eta := \min\Bigl\{1,\ \frac{\rho \sqrt{\alpha\, b}}{4 L}\Bigr\}, \qquad
\alpha_t \equiv \alpha,
\end{equation}
where $\alpha \in (0, 1]$. Then
\[
\frac{1}{T}\sum_{t=0}^{T-1}\E\!\big[\mathcal D(\tilde x^{t+1})^2\big]
\;\le\;
\frac{C'_1\, V^0}{\alpha\, T} \;+\; \frac{C'_2\, L\Delta_0}{\sqrt{\alpha b}\, T}
\;+\; \frac{C'_4\,\Delta_0}{\rho\,T} \;+\; \frac{C'_3\, \alpha\, \sigma^2}{b}
\]
for constants $C'_1, C'_2, C'_3, C'_4$ depending only on $\rho$, $L$ and $L_{r_2}$.
Choosing $\alpha = \bigl(\sqrt{b}\, L \Delta_0/(2\sigma^2 T)\bigr)^{2/3}$ (assume
$\sigma, L\Delta_0 > 0$; see Remark~\ref{rem:degenerate}), which lies in $(0,1]$ in the regime
below, the rate is
\[
\frac{1}{T}\sum_{t=0}^{T-1}\E\!\big[\mathcal D(\tilde x^{t+1})^2\big] \;=\; \mathcal O\!\Bigl((L\Delta_0)^{2/3}\sigma^{2/3}\, b^{-2/3}\, T^{-2/3} + \frac{V^0}{\alpha T} + \frac{\Delta_0}{\rho T}\Bigr).
\]
With $b$ constant, to attain $\E[\mathcal D(\hat x)^2] \le \eps^2$ it suffices, in the
small-$\eps$ regime $\eps \lesssim \min\{\sigma/\sqrt b,\ L\Delta_0/\sigma,\ \rho L\sigma/b\}$
(where $\lesssim$ hides constants depending only on $\rho, L, L_{r_2}$), to take
$T = \mathcal O\bigl(L\Delta_0\sigma/(b\eps^3)\bigr)$, at which
$\alpha = \Theta(b\eps^2/\sigma^2) \le 1$ and $\alpha T = \Theta(L\Delta_0/(\sigma\eps))$,
together with an initial batch
\begin{equation}
\label{eq:mvr-b0}
b_0 \;\ge\; \Bigl\lceil \frac{\sigma^3}{L\Delta_0\,\eps}\Bigr\rceil,
\end{equation}
which makes the initialization term $C'_1 V^0/(\alpha T)$ at most $\eps^2$. This is a
\emph{single} warm-up batch of size $\Theta(\eps^{-1})$, of the kind standard in the
analysis of momentum-based variance reduction; the per-iteration batch $b$ remains
constant. The total stochastic-gradient cost, warm-up included, is
$2 T b + b_0 = \mathcal O\bigl(L\Delta_0\sigma/\eps^3 + \sigma^3/(L\Delta_0\eps)\bigr)
= \mathcal O(L\Delta_0\sigma\eps^{-3})$ for $\eps \le L\Delta_0/\sigma$ (the factor $2$
accounts for the two evaluations per step in MVR).
\end{theorem}

\begin{remark}[Degenerate cases]
\label{rem:degenerate}
In~\eqref{eq:polyak-params} and~\eqref{eq:mvr-params}, read $\alpha\rho/(4L)$ and
$\rho\sqrt{\alpha b}/(4L)$ as $+\infty$ when $L = 0$, so that $\eta = 1$; every step of the
proofs remains valid, since each term carrying $L$ vanishes. All statements of
Theorems~\ref{thm:polyak-rate}--\ref{thm:mvr-rate-gap} remain valid with $\Delta_0$ replaced
by any $\bar\Delta \ge F(x^0) - F^*$, since the proofs use $\Delta_0$ only through
$\E[F(x^T)] - F(x^0) \ge -\Delta_0$; in the formulas that divide by $L\Delta_0$ (the Polyak
choice of $b_0$, the MVR choice of $\alpha$ and~\eqref{eq:mvr-b0}), $L\Delta_0$ should be
replaced by a positive upper bound $\Lambda \ge L\Delta_0$ when $L\Delta_0 = 0$. When
$\sigma = 0$ the oracle is exact, one takes $\alpha = 1$, and the bounds reduce to
$\mathcal O(1/T)$.
\end{remark}

\begin{proof}[Proof sketch]
The structure parallels Theorem~\ref{thm:polyak-rate} with the variance recursion replaced
by Lemma~\ref{lem:mvr-var}. The bias term in $\sum V^t$ is now $(2L^2/b)\sum D^t$, which
with $D^t = \eta^2\tilde D^t$ and $\eta \le \rho\sqrt{\alpha b}/(4L)$ becomes
at most $(\rho^2\alpha/8)\sum\tilde D^t$. After the same telescoping and elimination as in
Theorem~\ref{thm:polyak-rate}, the descent gives
$\sum V^t \le 2 V^0/\alpha + 4 L\Delta_0/\sqrt{\alpha b} + 4 T\alpha\sigma^2/b$.
Optimizing $\alpha$ in the resulting bound gives the stated rate; full bookkeeping,
including the initialization condition~\eqref{eq:mvr-b0}, is in
Appendix~\ref{app:thm-mvr-proof}.
\end{proof}

\subsection{Convergence rates: DCA fixed-point residual}
\label{sec:rates-residual}

The residual criterion drops Assumption~\ref{assu:smooth-r2} entirely. Unlike the
gap-function analysis it replaces, it is on the \emph{same} squared scale as the
critical-distance criterion, so the $\eps$-dependence of the two sets of rates coincides.

\begin{lemma}[Residual--variance bound]
\label{lem:R-bound}
Let Assumption~\ref{assu:sc} hold and let $H$ be convex and differentiable. Then for every
$t$,
\begin{equation}
\label{eq:R-vd}
\E\bigl[R_{\min}(x^t)\bigr] \;\le\; R^t := \E\bigl[R(x^t; w^t)\bigr] = \E\|\bar x^{t+1} - x^t\|^2 \;\le\; 2\,\tilde D^t \;+\; \frac{2}{\rho_1^2}\,V^t ,
\end{equation}
where $w^t$ is the subgradient selected by Algorithm~\ref{alg:mdca}.
Neither $L$-smoothness of $H$ nor smoothness of $r_2$ is used.
\end{lemma}

\begin{proof}
The points $\tilde x^{t+1}$ and $\bar x^{t+1}$ minimize the same $\rho_1$-strongly convex
function $G+r_1$ against the linear terms $g^t + w^t$ and $\nabla H(x^t) + w^t$, whose
difference is $e^t$. By \eqref{eq:argmin-contraction},
$\|\tilde x^{t+1} - \bar x^{t+1}\| \le \|e^t\|/\rho_1$, so
\[
\|\bar x^{t+1}-x^t\|^2 \le 2\|\bar x^{t+1}-\tilde x^{t+1}\|^2 + 2\|\tilde x^{t+1}-x^t\|^2
\le 2\|e^t\|^2/\rho_1^2 + 2\|\tilde x^{t+1}-x^t\|^2 ;
\]
take expectations.
\end{proof}

\noindent
This is the exact analogue of Lemma~\ref{lem:dist-bound} with $2(L+L_{r_2})^2$ replaced by
$2$ and $2$ replaced by $2/\rho_1^2$. Consequently Theorems~\ref{thm:polyak-rate}
and~\ref{thm:mvr-rate} transfer verbatim, with the output pair $(\hat x, \hat w) = (x^\tau, w^\tau)$,
$\tau$ uniform on $\{0,\dots,T-1\}$. Because $R_{\min}(\hat x) \le R(\hat x; \hat w)$, the
bounds hold both for the selected residual and for the choice-independent $R_{\min}$.

\begin{theorem}[Polyak rate, DCA residual]
\label{thm:polyak-rate-gap}
Let Assumptions~\ref{assu:sc}, \ref{assu:smooth-H}, \ref{assu:bbel} hold;
Assumption~\ref{assu:smooth-r2} is \emph{not} required and $r_2$ need not be
differentiable. Run Algorithm~\ref{alg:mdca} with the Polyak update and the constants
\eqref{eq:polyak-params}, and let $(\hat x, \hat w)$ be its output. Then
\[
\E\bigl[R_{\min}(\hat x)\bigr] \;\le\; \E\bigl[R(\hat x;\hat w)\bigr] \;\le\; \frac{K_1(V^0 + L\Delta_0)}{\alpha T}
+ \frac{8\Delta_0}{\rho T} + \frac{K_2\,\alpha\sigma^2}{b},
\qquad K_1 := \tfrac{32}{\rho^2} + \tfrac43\kappa_R, \quad K_2 := \tfrac43\kappa_R,
\]
with $\kappa_R := 8/\rho^2 + 2/\rho_1^2$. With $A := V^0 + L\Delta_0$ and
$\alpha = \min\{1,\sqrt{K_1 bA/(K_2\sigma^2 T)}\}$, the right-hand side is at most
$2\sqrt{K_1K_2A\sigma^2/(bT)} + 2K_1A/T + 8\Delta_0/(\rho T)$ in both regimes, as
in~\eqref{eq:polyak-two-regime}. With $b_0 = 1$, it is at most $\eps^2$ once
$T \ge \max\{36K_1K_2A\sigma^2/(b\eps^4),\ 6K_1A/\eps^2,\ 24\Delta_0/(\rho\eps^2)\}$, i.e.\
$T = \mathcal O\bigl(\sigma^2(\sigma^2+L\Delta_0)/(b\eps^4)\bigr)$ in the small-$\eps$
regime, for a total cost $Tb + b_0 = \mathcal O\bigl(\sigma^2(\sigma^2+L\Delta_0)/\eps^4\bigr)$.
\end{theorem}

\begin{proof}
Steps~1--3 of Appendix~\ref{app:thm-polyak-proof} use only Lemma~\ref{lem:polyak-var} and
Lemma~\ref{lem:descent}, neither of which requires Assumption~\ref{assu:smooth-r2}. By
Lemma~\ref{lem:R-bound} and \eqref{eq:proof-thm1-D},
\[
\sum_{t=0}^{T-1} R^t \;\le\; 2\Bigl[\tfrac{4\Delta_0}{\eta\rho} + \tfrac{4}{\rho^2}\sum V^t\Bigr]
+ \tfrac{2}{\rho_1^2}\sum V^t
= \tfrac{8\Delta_0}{\eta\rho} + \kappa_R \sum V^t .
\]
Insert \eqref{eq:proof-thm1-sumV} and use $1/\eta \le 1 + 4L/(\alpha\rho)$, which gives
$8\Delta_0/(\eta\rho) \le 8\Delta_0/\rho + 32L\Delta_0/(\alpha\rho^2)$; divide by $T$ and
collect, bounding $L\Delta_0 \le V^0 + L\Delta_0$; the first inequality of the statement is
$R_{\min}(x^t) \le R(x^t; w^t)$. Choosing $\alpha$ as in Step~6 of
Appendix~\ref{app:thm-polyak-proof} gives the stated complexity.
\end{proof}

\begin{theorem}[MVR rate, DCA residual]
\label{thm:mvr-rate-gap}
Under additionally Assumption~\ref{assu:avg-smooth}, with the constants
\eqref{eq:mvr-params},
\[
\E\bigl[R_{\min}(\hat x)\bigr] \;\le\; \E\bigl[R(\hat x;\hat w)\bigr] \;=\;
\mathcal O\!\Bigl(\frac{V^0}{\alpha T} + \frac{L\Delta_0}{\sqrt{\alpha b}\,T}
+ \frac{\Delta_0}{\rho T} + \frac{\alpha\sigma^2}{b}\Bigr).
\]
With $\alpha = \bigl(\sqrt b\,L\Delta_0/(2\sigma^2T)\bigr)^{2/3}$,
$T = \mathcal O\bigl(L\Delta_0\sigma/(b\eps^3)\bigr)$ and $b_0$ as in \eqref{eq:mvr-b0},
in the small-$\eps$ regime of Theorem~\ref{thm:mvr-rate},
$\E[R_{\min}(\hat x)] \le \eps^2$ at total cost $\mathcal O(L\Delta_0\sigma\eps^{-3})$ -- the same
$\eps$-dependence as Theorem~\ref{thm:mvr-rate}, but without
Assumption~\ref{assu:smooth-r2}.
\end{theorem}

\noindent
The proof is Step~3 of Appendix~\ref{app:thm-mvr-proof} followed by
Lemma~\ref{lem:R-bound} in place of Lemma~\ref{lem:dist-bound}, exactly as above.

\section{Single-loop variant}
\label{sec:single-loop}

When the inner subproblem in MDCA is too costly to solve to high accuracy at every
iteration, one can replace it by one stochastic prox-gradient step on $G + r_1$. We
also smooth $r_2$ via its Moreau envelope; this avoids requiring a subgradient of $r_2$
in the descent direction and gives a uniform smoothness bound on the surrogate
objective. Writing $f_\gamma(x) := \min_y\{f(y) + \tfrac{1}{2\gamma}\|y - x\|^2\}$ for the
Moreau envelope of a convex function $f$, define
\begin{equation}
\label{eq:Fgamma}
F_\gamma(x) := (G + r_1)_\gamma(x) - H(x) - (r_2)_\gamma(x).
\end{equation}
By standard Moreau-envelope theory \citep{Beck2017,ParikhBoyd2014}, both
$(G + r_1)_\gamma$ and $(r_2)_\gamma$ are $1/\gamma$-smooth and convex; combined with
$L$-smoothness of $H$, $F_\gamma$ is $L_\gamma$-smooth with
$L_\gamma = L + 2/\gamma$. Its gradient is
\begin{equation}
\label{eq:grad-Fgamma}
\nabla F_\gamma(x) = \frac{z^*(x) - u^*(x)}{\gamma} - \nabla H(x),
\quad
u^*(x) := \prox_{\gamma(G+r_1)}(x),
\quad
z^*(x) := \prox_{\gamma r_2}(x).
\end{equation}
A stationary point of $F_\gamma$ corresponds, via the prox of $G + r_1$, to an approximate
critical point of $F$ with an error that is $\mathcal O(\gamma)$ when $r_2$ is smooth and
$\|p_1\|, \|\nabla H\|$ are bounded along the iterates
(Lemma~\ref{lem:gamma-equivalence}, proved in Appendix~\ref{app:smoothing-eq}); see
Remark~\ref{rem:gamma-tight} for why no unconditional statement is available.

Note that $F_\gamma$ smooths the two DC \emph{components} separately; it is not the Moreau
envelope of $F$, which would require $F$ to be weakly convex and a prox for the whole
nonconvex objective. Consequently $F_\gamma$-stationarity does not reduce to
$F$-criticality at fixed $\gamma$; see Remark~\ref{rem:gamma-tight}.

\begin{algorithm}[t]
\caption{Single-loop momentum-based DCA (S-MDCA)}
\label{alg:smdca}
\begin{algorithmic}[1]
\Require $x^0$, auxiliary $u^0 = x^0$; $\alpha_t, \eta^0_t, \eta^1_t$, $\gamma > 0$,
batches $b_0, b$; iterations $T$.
\State Sample $\mathcal I^0$ of size $b_0$. Set $g^0 = \tfrac{1}{b_0}\sum_{i \in \mathcal I^0}\nabla h_i(x^0)$.
\For{$t = 0, 1, \dots, T - 1$}
  \State $z^t = \prox_{\gamma r_2}(x^t)$ \hfill (smoothing of $r_2$; assumes prox is cheap)
  \State Inner stochastic prox-gradient step on $G + r_1 + \tfrac{1}{2\gamma}\|\cdot - x^t\|^2$:
  \[
    u^{t+1} = \prox_{\eta^1_t r_1}\!\Big(u^t - \eta^1_t\big(\nabla G(u^t; \xi^G_t) + (u^t - x^t)/\gamma\big)\Big).
  \]
  \State Search direction:\quad $\mathcal G^t = (z^t - u^{t+1})/\gamma - g^t$ \hfill (approximation to $\nabla F_\gamma(x^t)$)
  \State Outer step:\quad $x^{t+1} = x^t - \eta^0_t \mathcal G^t$.
  \State Sample $\mathcal I^{t+1}$ of size $b$, update $g^{t+1}$ as in Algorithm~\ref{alg:mdca} (Polyak or MVR).
\EndFor
\State \Return $\hat x$ chosen uniformly from $\{x^0, \dots, x^{T-1}\}$.
\end{algorithmic}
\end{algorithm}

\begin{remark}[If $\prox_{\gamma r_2}$ is unavailable]
\label{rem:smdca-subgrad}
When the prox of $r_2$ is not in closed form, one may replace
$z^t = \prox_{\gamma r_2}(x^t)$ by $z^t := x^t - \gamma w^t$ for any
$w^t \in \partial r_2(x^t)$, giving $\mathcal G^t = (x^t - u^{t+1})/\gamma - w^t - g^t$. This
is not an inexact version of Algorithm~\ref{alg:smdca} but an exact method for the
\emph{partially} smoothed objective $\Phi_\gamma := (G+r_1)_\gamma - H - r_2$, in which
$r_2$ is left unsmoothed. Theorem~\ref{thm:smdca-subgrad} in
Appendix~\ref{app:smdca-subgrad} shows that the guarantee of Theorem~\ref{thm:smdca} then
holds verbatim for $\Phi_\gamma$, with the same constants, for every $\gamma > 0$ for which
$\Phi_\gamma$ is bounded below, and with no bound on $\partial r_2$; the certificate is the
algorithm's selected $v^t = \nabla\psi_\gamma(x^t) - w^t$. It does not target $F_\gamma$: the two objectives have different
stationary points, with a discrepancy that does not vanish as $\gamma \to 0$
(Remark~\ref{rem:subgrad-not-Fgamma}). We therefore retain the closed-form-prox assumption
in Theorem~\ref{thm:smdca}.
\end{remark}

\begin{theorem}[Single-loop convergence]
\label{thm:smdca}
Let Assumptions~\ref{assu:wellposed}, \ref{assu:smooth-H}, \ref{assu:G-oracle} hold, and let
$\gamma > 0$ be fixed such that Assumption~\ref{assu:Fgamma-bbel} holds
(Assumptions~\ref{assu:sc} and~\ref{assu:bbel} are not needed; convexity of the components
suffices).
\textbf{We assume throughout this section that $\prox_{\eta^1 r_1}$ and $\prox_{\gamma r_2}$
are available in closed form.} Since $r_1$ and $r_2$ must be convex, this refers to the
convex pieces of the DC decompositions, not to the nonconvex penalties themselves: for
instance $r_1 = \lambda\|\cdot\|_1$ or group-$\ell_1$ or the indicator of a convex set with
closed-form projection, and $r_2 = \lambda\sum_j\max\{|x_j| - \tau, 0\}$ (the correction term of
capped-$\ell_1$) or the Huber-type corrections $h_{\mathrm{MCP}}$, $h_{\mathrm{SCAD}}$ of MCP and
SCAD, all of which have closed-form scalar proxes. Inexact-prox extensions are deferred to
Appendix~\ref{app:smdca-inexact-prox}. Run Algorithm~\ref{alg:smdca} with constant Polyak
(or MVR) momentum $\alpha$, smoothing parameter $\gamma > 0$, inner step $\eta^1$, outer step
$\eta^0$, and per-iteration batch $b$, with the schedule
\begin{equation}
\label{eq:smdca-params}
\eta^1 \le \frac{\gamma}{1 + \gamma L_G}\;\Bigl(= \frac{1}{L_\phi}\Bigr),
\qquad
\eta^0 = \min\!\Big\{\frac{1}{4 L_\gamma},\ \frac{\eta^1}{24},\ \frac{\kappa_V}{L}\Big\},
\end{equation}
where $\kappa_V = \alpha/\sqrt{96}$ for Polyak and $\kappa_V = \sqrt{\alpha b}/\sqrt{192}$
for MVR, and $L_\phi := L_G + 1/\gamma$. Set
$\bar\Ls := \Delta_\gamma + \tfrac{\gamma}{4}\|p_1^0\|^2$ with
$p_1^0 := (x^0 - u^*(x^0))/\gamma$. Then with $\hat x$ chosen uniformly from
$\{x^0, \dots, x^{T-1}\}$:
\begin{itemize}[leftmargin=*,topsep=2pt,itemsep=2pt]
\item \emph{Polyak.}
\(
\E\|\nabla F_\gamma(\hat x)\|^2 \le \dfrac{8\,\bar\Ls}{\eta^0 T} + \dfrac{24\, V^0}{\alpha T}
+ \dfrac{86\,\eta^1 M^2}{\gamma} + \dfrac{24\,\alpha \sigma^2}{b}.
\)

\item \emph{MVR} (additionally requiring Assumption~\ref{assu:avg-smooth}).
\(
\E\|\nabla F_\gamma(\hat x)\|^2 \le \dfrac{8\,\bar\Ls}{\eta^0 T} + \dfrac{24\, V^0}{\alpha T}
+ \dfrac{86\,\eta^1 M^2}{\gamma} + \dfrac{48\,\alpha \sigma^2}{b}.
\)
\end{itemize}

\noindent
\emph{Optimization of $\eta^1, \alpha, T$ to attain $\E\|\nabla F_\gamma(\hat x)\|^2 \le \eps^2$.}
The $G$-side noise forces $\eta^1 = \mathcal O(\gamma\eps^2/M^2)$, which through the second
constraint in~\eqref{eq:smdca-params} gives $\eta^0 = \Theta(\gamma\eps^2/M^2)$ for any
sufficiently small $\eps$. The descent term then requires
\begin{equation}
\label{eq:smdca-G-bound}
T \;\ge\; T_G \;:=\; \mathcal O\!\Big(\frac{\bar\Ls M^2}{\gamma\,\eps^4}\Big),
\end{equation}
while the $H$-side variance requires, after optimizing $\alpha = \mathcal O(b\eps^2/\sigma^2)$,
$T \ge T_H = \mathcal O(L\bar\Ls\sigma^2/(b\eps^4))$ (the factor $L$ enters through the
constraint $\eta^0 \le \alpha/(L\sqrt{96})$ of \eqref{eq:eta0-constraints}). The
initialization term needs only
$V^0 \lesssim L\bar\Ls$, i.e.\ a \emph{constant} $b_0 = \Theta(\sigma^2/(L\bar\Ls))$.
Total iterations $T = \max(T_G, T_H)$; per-iteration cost is $2b$ on $H$ and $1$ on $G$, so
the total stochastic-gradient cost is
$\mathcal O\bigl(\bar\Ls(L\sigma^2 + bM^2/\gamma)/\eps^4\bigr)$ for both variants, the same
$\eps^{-4}$ scaling as the double-loop variant in Theorem~\ref{thm:polyak-rate}. These choices
assume $L, M, \sigma > 0$, fixed $\gamma$, and $\eps$ small enough that $\eta^1/24$ and
$\kappa_V/L$ fall below $1/(4L_\gamma)$. In the edge cases: if $M = 0$ the $G$-side term
vanishes and $\eta^1 = 1/L_\phi$ may be kept constant; if $\sigma = 0$ one takes $\alpha = 1$;
if $L = 0$ the third constraint in~\eqref{eq:smdca-params} is void. The MVR
$\eps^{-3}$ advantage of the double-loop variant (Theorem~\ref{thm:mvr-rate}) does
\emph{not} carry over here; see Remark~\ref{rem:inner-vr}.
\end{theorem}

\begin{remark}[Why MVR gives no asymptotic gain in our single-loop bound]
\label{rem:inner-vr}
Substituting Lemma~\ref{lem:mvr-var} for Lemma~\ref{lem:polyak-var} in the Lyapunov
argument of Appendix~\ref{app:thm-smdca-proof} relaxes the third constraint
of~\eqref{eq:smdca-params} from $\eta^0 \lesssim \alpha/L$ to
$\eta^0 \lesssim \sqrt{\alpha b}/L$ and doubles the $\sigma^2$ constant; nothing else
changes. Because the $G$-side noise forces $\eta^1 = \Theta(\gamma\eps^2/M^2)$ and hence
$\eta^0 \le \eta^1/24 = \Theta(\eps^2)$, while $\sqrt{\alpha b}/L = \Theta(\eps)$ at
$\alpha = \Theta(b\eps^2/\sigma^2)$, the binding constraint for small $\eps$ is the
$\eta^1$ one in \emph{both} cases. MVR therefore relaxes a constraint that is not binding,
and our bound on the iteration complexity remains $\mathcal O(\bar\Ls M^2/(\gamma\eps^4))$. The
bottleneck in the analysis is the variance $M^2$ of the inner stochastic prox-gradient step;
we expect that recovering $\eps^{-3}$ requires variance reduction on $\nabla G$ as well, which changes the
algorithm's memory footprint (extra storage of one previous gradient per iteration).
\end{remark}

A full self-contained proof, including the careful $\eta^0, \eta^1, \alpha$ optimization,
is given in Appendix~\ref{app:thm-smdca-proof}. Briefly: the proof tracks three quantities
-- the gap $F_\gamma(x^t) - F_\gamma^*$, the proximal-tracking error
$\mathcal E^t := \E\|u^t - u^*(x^t)\|^2$ measured at the \emph{current} base point, and the
gradient variance $V^t := \E\|g^t - \nabla H(x^t)\|^2$ -- and combines them into a single
Lyapunov function $\mathcal V^t$ whose one-step decrease bounds $\E\|\nabla F_\gamma(x^t)\|^2$ up
to noise terms. Appendix~\ref{app:smdca-inexact-prox} discusses how the analysis extends to
inexact prox-evaluations of $r_1$ and $r_2$.

\begin{remark}[$F_\gamma$-stationarity at fixed $\gamma$ does not imply $F$-criticality]
\label{rem:gamma-tight}
Theorem~\ref{thm:smdca} applies to every fixed $\gamma > 0$ for which
Assumption~\ref{assu:Fgamma-bbel} holds and imposes no further condition on $\gamma$. It
should not, however, be read as a guarantee about $F$: the $\gamma$-dependent
terms in Lemma~\ref{lem:gamma-equivalence}(iii) are unavoidable and are attained with
equality. Take $r_1 = r_2 = 0$, $G(x) = \tfrac a2x^2$, $H(x) = \tfrac b2x^2$ with
$a > b > 0$, so Assumptions~\ref{assu:sc}, \ref{assu:smooth-H}, \ref{assu:smooth-r2},
\ref{assu:bbel} hold with $\rho_1 = a$, $\rho_2 = b$, $L = b$, $L_{r_2} = 0$ and
$F(x) = \tfrac{a-b}{2}x^2$. Since $(G)_\gamma(x) = \tfrac{a}{2(1+a\gamma)}x^2$,
\[
F_\gamma(x) = \Bigl(\tfrac{a}{1+a\gamma} - b\Bigr)\tfrac{x^2}{2},
\qquad
\nabla F_\gamma(x) = \Bigl(\tfrac{a}{1+a\gamma} - b\Bigr)x .
\]
For $\gamma > 1/b - 1/a$ the coefficient is negative, $F_\gamma$ is unbounded below, and
Theorem~\ref{thm:smdca} does not apply.
At $\gamma = 1/b - 1/a > 0$ the coefficient vanishes identically: $F_\gamma \equiv 0$, so
$\nabla F_\gamma \equiv 0$ at \emph{every} point and $\Delta_\gamma = 0$. Yet
$\hat x = u^*(x) = x/(1+a\gamma)$ gives $\nabla F(\hat x) = (a-b)x/(1+a\gamma)$, which is
unbounded. With $a = 2$, $b = 1$, $\gamma = 1/2$ one has $p_1 = x$ and
$\gamma L\|p_1\| = |x|/2 = |\nabla F(\hat x)|$, so the bound of
Lemma~\ref{lem:gamma-equivalence}(iii) holds with equality. The cause is structural:
$F_\gamma$ smooths the two DC components \emph{separately}, displacing the critical set;
$F_\gamma$ is not the Moreau envelope of $F$. Hence $F_\gamma$-stationarity is a genuinely
distinct criterion, and converting it to $\mathcal O(\eps)$-criticality of $F$ via
Lemma~\ref{lem:gamma-equivalence}(iii) would require $\gamma = \mathcal O(\eps)$ -- raising
the iteration bound to $\mathcal O(\eps^{-5})$ -- as well as Assumption~\ref{assu:smooth-r2}.
We make no such claim; all single-loop guarantees are stated for $F_\gamma$.
\end{remark}

\begin{remark}[$F_\gamma$ depends on the DC decomposition]
\label{rem:Fgamma-decomp}
Unlike $F$, the surrogate $F_\gamma$ changes when the same function is added to both DC
components, because $H$ is not smoothed while $G + r_1$ is. In particular, the
strong-convexity reformulation of Section~\ref{sec:setup} can destroy
Assumption~\ref{assu:Fgamma-bbel}. Suppose $\tfrac\varrho2\|\cdot\|^2$ is added to $G + r_1$
and to $H$. Taking $u = x/(1+\gamma\varrho)$ in the definition of the envelope gives
$(G + r_1 + \tfrac\varrho2\|\cdot\|^2)_\gamma(x) \le (G+r_1)\bigl(x/(1+\gamma\varrho)\bigr) +
\tfrac{\varrho}{2(1+\gamma\varrho)}\|x\|^2$, while $H$ gains the full $\tfrac\varrho2\|x\|^2$.
Hence if the unshifted $G + r_1$ is Lipschitz, $(G + r_1) - H \le a + c\|x\|$ and
$r_2 \ge -a - c\|x\|$ for some constants $a, c$ (as for the softplus losses, $\ell_1$ term and
capped-$\ell_1$ correction of Section~\ref{sec:experiments}), then
$F_\gamma(x) \le -\tfrac{\gamma\varrho^2}{2(1+\gamma\varrho)}\|x\|^2 + \mathcal O(\|x\|)$,
so $F_\gamma$ is unbounded below for \emph{every} $\gamma > 0$. Assigning the concave-side
quadratic to $r_2$ instead (where it is smoothed) removes this particular failure. On the
truncated-quadratic regression instance of Appendix~\ref{sec:experiments-regression},
$F_\gamma$ is unbounded below for every $\gamma > 0$ regardless of this choice.
\end{remark}

\section{Experiments}
\label{sec:experiments}

We evaluate MDCA on a robust+sparse classification objective of the form
\begin{equation}
\label{eq:exp-obj}
F(\theta) = \tfrac{1}{n}\!\sum_i \zeta(1 - z_i) + \lambda_1 \|\theta\|_1 + \tfrac{\varrho}{2}\|\theta\|^2
- \tfrac{1}{n}\!\sum_i \zeta(-2 - z_i) - \lambda_2\!\sum_j \max\{|\theta_j| - \tau, 0\}
- \tfrac{\varrho}{2}\|\theta\|^2,
\end{equation}
where $\zeta(u) = \log(1 + e^u)$ is the softplus and $z_i = y_i\, a_i^\top \theta$. The
first three terms form $G + r_1$ (smooth data + $\ell_1$ + ridge); the last three form
$H + r_2$ (smooth data + capped-$\ell_1$ truncation as $r_2$ + matching ridge). When
$\lambda_1 = \lambda_2$ the combined regularization is the capped-$\ell_1$ penalty
$\lambda_1\sum_j\min(|\theta_j|, \tau)$, robust to outliers in the sparsity pattern.
For a linear model, $\zeta(1 - z_i)$ and $\zeta(-2 - z_i)$ are convex and smooth in $\theta$, so
this split satisfies Assumptions~\ref{assu:wellposed}--\ref{assu:smooth-H} with the ridge
supplying the strong convexity of Assumption~\ref{assu:sc}; since
$0 \le \zeta(1-z) - \zeta(-2-z) \le 3$, $F$ is bounded below. For the single-loop method the
placement of the concave-side ridge matters: by Remark~\ref{rem:Fgamma-decomp}, $F_\gamma$ is
unbounded below for every $\gamma > 0$ if the matching ridge is part of $H$, and
Theorem~\ref{thm:smdca} applies only if it is assigned to $r_2$.

\paragraph{Datasets.} \texttt{a9a} ($n = 32{,}561$, $d = 123$) and \texttt{w8a}
($n = 49{,}749$, $d = 300$) from LIBSVM~\citep{Chang2011} with the standard splits.

\paragraph{Methods.}
\textbf{MDCA-Polyak} and \textbf{MDCA-MVR} (Algorithm~\ref{alg:mdca});
\textbf{S-MDCA-Polyak} (Algorithm~\ref{alg:smdca}); \textbf{SDCA}, the constant-batch
baseline corresponding to $\alpha = 1$, $\eta_t = 1$ in Algorithm~\ref{alg:mdca};
\textbf{SMAG} \citep{Hu2024}, the single-loop baseline;
\textbf{DCA-PAGE} \citep{Nguyen2025}, with refresh batch of size
$b_{\mathrm{full}} = 64$ and refresh probability $p = 0.03$. All methods use a constant per-iteration
batch $b = 4$ (initial batch $b_0 = 64$) and run $T = 3000$ iterations.

\paragraph{Metrics and hyperparameters.} (i) Best-so-far training objective
$\min_{s \le t} F(\theta_s)$;
(ii) stationarity $\|\mathcal S(\theta_{s^*})\|^2$ at the $F$-best iterate
$s^* = \arg\min_{s \le t} F(\theta_s)$, where $\|\mathcal S(\cdot)\|^2$ is the squared
proximal-gradient mapping norm (distinct from the search direction $\mathcal G^t$ of
Algorithm~\ref{alg:smdca});
(iii) sparsity at $\theta_{s^*}$.
We report metrics (ii) and (iii) at the $F$-best iterate $s^*$, so that all three
quantities are read at the same point along the trajectory. This differs from the randomly
sampled output of the theorems, so the plots illustrate the methods rather than test the
bounds.

The horizontal axis is the total number of stochastic gradient
evaluations on $G$ and $H$. Hyperparameters tuned per method on a small grid:
SDCA uses $\eta_t = 1$ (no damping, vanilla DCA step); MDCA-Polyak uses $\alpha = 0.1$,
$\eta_t = 0.5$; MDCA-MVR uses $\alpha = 0.05$, $\eta_t = 0.5$; single-loop variants use
$\gamma = 0.5$, $\eta^0 = \eta^1 = 0.05$; DCA-PAGE matches its original (no damping)
formulation. We add 18\% label noise, so that the stochastic-gradient variance $\sigma^2$ is
non-negligible at this batch size. Note that MDCA uses the damped step $\eta_t = 0.5$ while
DCA-PAGE is undamped, so the MDCA--DCA-PAGE comparison changes \emph{both} the estimator and
the damping.

Since the inner subproblems are solved approximately with stochastic gradients of $G$, the
exact-solve theorems do not apply verbatim; the relevant guarantee is the function-value
route of Theorem~\ref{thm:inexact-residual}, which requires the inner accuracy to scale with
$\eps^2$.

\begin{figure}[t]
\centering
\includegraphics[width=\linewidth]{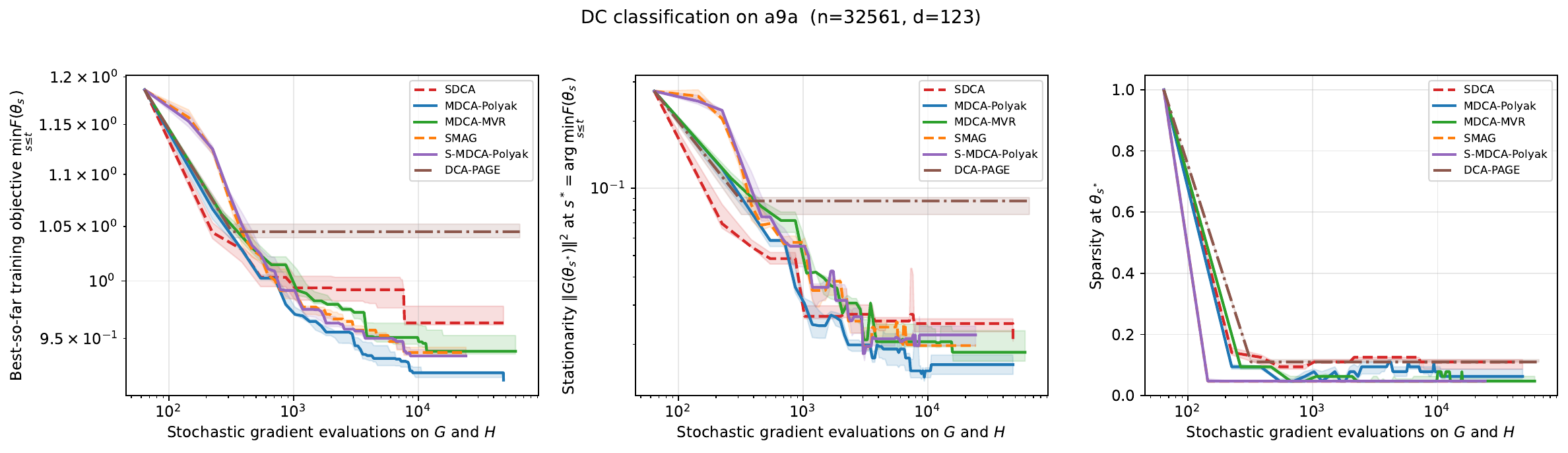}
\caption{Results on \texttt{a9a}. Median over 3 seeds; IQR shaded (with three seeds this
band is only indicative).
\textit{Left:} training objective. \textit{Center:} stationarity (squared
prox-gradient mapping norm). \textit{Right:} sparsity. Horizontal axis: stochastic
gradient evaluations on $G$ and $H$ combined.}
\label{fig:a9a}
\end{figure}

\begin{figure}[t]
\centering
\includegraphics[width=\linewidth]{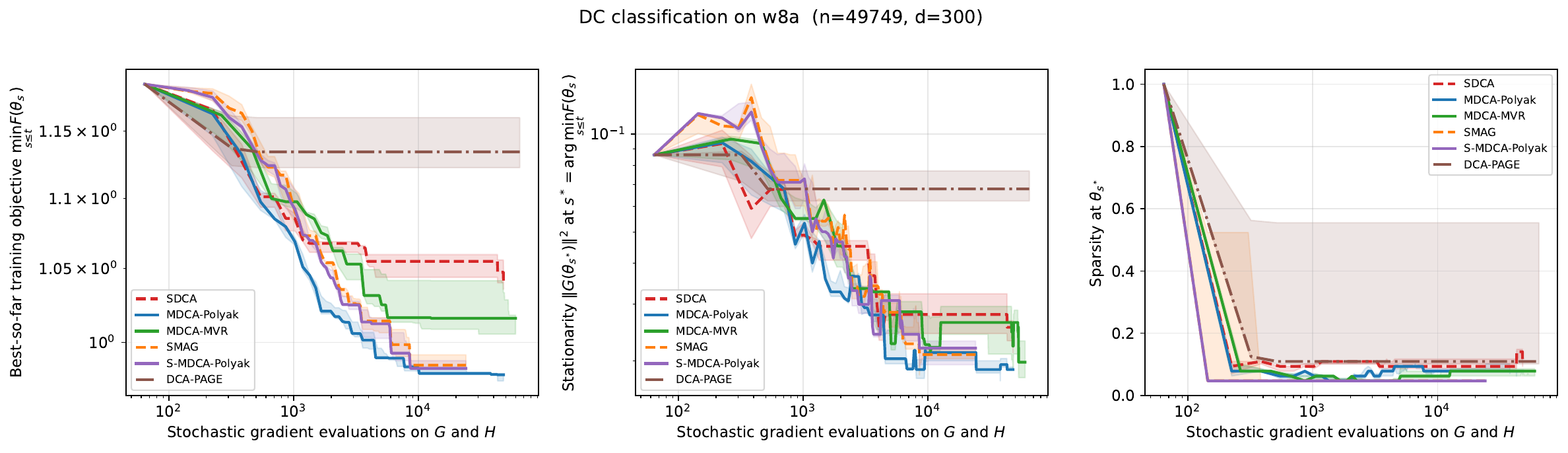}
\caption{Results on \texttt{w8a}. Same conventions as Figure~\ref{fig:a9a}.}
\label{fig:w8a}
\end{figure}

\paragraph{Results.} These experiments are preliminary: with three seeds they support
descriptive rather than statistically strong comparative conclusions. In our runs, on both
datasets MDCA-Polyak attains the lowest median training objective (BestF median $\approx 0.93$
on a9a, $0.98$ on w8a, vs.\ SDCA's $0.97$/$1.04$ and DCA-PAGE's $1.05$/$1.13$) and the lowest
median stationarity at the $F$-best iterate (0.04 vs.\ DCA-PAGE's 0.09 on a9a; 0.019 vs.\
DCA-PAGE's 0.068 on w8a).
MDCA-MVR also improves over SDCA on both measures in these runs, but more slowly than Polyak
at this horizon. The theoretical $\eps^{-3}$ versus $\eps^{-4}$ distinction concerns
worst-case upper bounds as $\eps \to 0$; it is not observed at $T = 3000$ and we have not
tested whether it appears at larger $T$. Single-loop momentum (S-MDCA-Polyak) similarly
improves over the single-loop baseline SMAG. DCA-PAGE's training objective
plateaus higher. Because the comparison changes both the estimator and the damping, we do not
attribute this plateau to the refresh mechanism; isolating the cause requires a damped
DCA-PAGE run ($x^{t+1} = x^t + \eta(\tilde x^{t+1} - x^t)$ with the same $\eta = 0.5$) and a
direct measurement of the estimator error $\|g^t - \nabla H(x^t)\|^2$ along each trajectory.
See Figures~\ref{fig:a9a}--\ref{fig:w8a}.

\section{Discussion and Conclusion}

\paragraph{What we showed.} With exact solves of the strongly convex inner subproblem and a
constant per-iteration batch, damped MDCA with Polyak momentum reaches
$\E[\mathcal D(\hat x)^2] \le \eps^2$ with $\mathcal O(L\Delta_0\sigma^2\eps^{-4})$ stochastic
gradients in the small-$\eps$ regime, and MVR under averaged smoothness reaches it with
$\mathcal O(L\Delta_0\sigma\eps^{-3} + \sigma^3/(L\Delta_0\eps))$, the second term being a
single warm-up batch. This matches the $\eps$-dependence obtained by substituting the online
refresh batch into the DCA-PAGE bound of \citet{Nguyen2025}, without a refresh branch. The same rates hold for the DCA
fixed-point residual, stated both for the selected $R(\hat x;\hat w)$ and for the
choice-independent $R_{\min}(\hat x)$, which vanishes exactly at DC-critical points
(Lemma~\ref{lem:R-criticality}) and requires no smoothness of $r_2$; unlike the gap function
it replaces, its zero set is the critical set (Remark~\ref{rem:gap-issue}). The single-loop
variant (S-MDCA) reaches $\eps$-stationarity of the smoothed objective $F_\gamma$ in
$\mathcal O(\eps^{-4})$ iterations for every fixed $\gamma > 0$ for which $F_\gamma$ is bounded
below (Assumption~\ref{assu:Fgamma-bbel}); this condition restricts $\gamma$, depends on how
$F$ is split into DC components (Remark~\ref{rem:Fgamma-decomp}), and fixed-$\gamma$
stationarity of $F_\gamma$ does not by itself imply stationarity of $F$
(Remark~\ref{rem:gamma-tight}). Variance reduction on $\nabla H$ alone does not remove the
inner-step $\nabla G$ noise, which is why the MVR $\eps^{-3}$ dependence does not carry over
to the single-loop setting in our analysis (Remark~\ref{rem:inner-vr}).

\paragraph{The role of momentum.} Lemma~\ref{lem:separation} records a variance--drift
trade-off in steady-state \emph{upper bounds}: the momentum parameter $\alpha \in (0,1]$
trades the noise contribution $\alpha\sigma^2/b$ against the drift contribution, and the
optimized Polyak and MVR bounds scale as $\bar D^{1/3}$ and $\bar D^{1/2}$ respectively in the
drift. These are upper bounds and are not used in the proofs of the rates; they do not show
that momentum is necessary or that one estimator is better than another in a worst-case sense.
The existence of a non-vanishing floor without momentum at constant stepsize and batch is a
statement about a specific instance (Proposition~\ref{prop:noise-floor}), and that floor can
also be removed by decreasing stepsizes or growing batches.

\paragraph{Practical considerations.} Inexact inner solves are covered in
Appendix~\ref{app:inexact}: function-value accuracy $\bar\nu \lesssim \rho_1\eps^2$ preserves
the residual rates (Theorem~\ref{thm:inexact-residual}), while the critical-distance rates
require an approximate optimality certificate with $\bar\delta^2 \lesssim \eps^2/(1 +
(L+L_{r_2})^2/\rho_1^2)$ (Theorem~\ref{thm:inexact-crit}). Such accuracies are reached at
logarithmic inner cost only under additional structure, for instance a smooth strongly
convex $G$ with an available $\prox_{r_1}$; with a stochastic inner solver the cost per outer
step grows like $\eps^{-2}$, which motivates the single-loop variant. The constants in our
bounds are worst-case upper bounds and the experiments rely on grid tuning; developing
parameter-free or adaptive variants remains open.

\paragraph{Limitations and Future Directions.} Our variance recursions, and hence all rates,
rely on the $L$-smoothness of $H$ and on a uniform bound on the stochastic-gradient variance;
the descent template itself uses only differentiability of $H$. Further limitations:
\begin{itemize}[leftmargin=*]
    \item \textbf{No lower bounds.} We do not prove lower bounds for the constant-memory,
    constant-main-batch model considered here, so the rates should be read as upper bounds.
    \item \textbf{Preliminary experiments.} The experiments use three (main text) or five
    (Appendix~\ref{sec:experiments-regression}) seeds and approximate inner solves; they are
    descriptive, and the causes of the observed differences (in particular the DCA-PAGE
    plateau) have not been isolated.
    \item \textbf{Scope of the examples.} The truncated-quadratic robust loss, with the split
    used in Appendix~\ref{sec:experiments-regression}, has a nonsmooth $H$ with unbounded
    gradient variance, and is outside our assumptions (Example~2 in Section~\ref{sec:intro} gives
    alternatives inside them).
    \item \textbf{Decomposition dependence of the single-loop surrogate.} Whether $F_\gamma$ is
    bounded below depends on $\gamma$ and on the DC split, not only on $F$
    (Remark~\ref{rem:Fgamma-decomp}).
\end{itemize}
We identify several avenues for extending this framework:
\begin{itemize}[leftmargin=*]
    \item \textbf{Heavy-tailed noise regimes:} Standard bounded variance models are frequently violated in modern stochastic settings. A critical next step is to analyze these momentum-based DC algorithms under weak-$L_p$ tail assumptions, potentially incorporating robust scalar-clipping techniques to maintain convergence guarantees without strict variance bounds.
    \item \textbf{Nonsmooth concave components:} The case where $H$ is nonsmooth and accessed only via stochastic subgradients with bounded second moments remains open. There the drift term $L^2\|x^{t+1}-x^t\|^2$ of the variance recursions is unavailable, so a different bounding strategy is needed. Covering the truncated-quadratic split of Appendix~\ref{sec:experiments-regression} would in addition require handling gradient variance that grows with $\|\theta\|$.
    \item \textbf{Single-loop target:} the partially smoothed objective
    $\Phi_\gamma := (G+r_1)_\gamma - H - r_2$ of Appendix~\ref{app:smdca-subgrad} is closer
    to $F$ than $F_\gamma$ and requires no prox of $r_2$; building the single-loop
    algorithm around it from the outset is an attractive alternative
    (Remark~\ref{rem:phi-gamma-future}).
\end{itemize}
\section*{Acknowledgments}

\bibliography{main}
\bibliographystyle{plainnat}

\appendix
\newpage
\section{Robust Sparse Regression Experiment}
\label{sec:experiments-regression}

The experiments above are classification problems. As an \emph{out-of-assumption stress
test}, we consider a robust sparse regression instance: the truncated quadratic loss with a
capped-$\ell_1$ penalty,
\begin{equation}
\label{eq:exp-obj-reg}
F(\theta) = \underbrace{\tfrac{1}{n}\!\sum_i (y_i - a_i^\top\theta)^2
  + \lambda_1\|\theta\|_1 + \tfrac{\varrho}{2}\|\theta\|^2}_{G + r_1}
\; - \underbrace{\Big[\tfrac{1}{n}\!\sum_i \max\{(y_i - a_i^\top\theta)^2 - \tau_\ell^2,\, 0\}
  + \lambda_2\!\sum_j \max\{|\theta_j| - \tau,\, 0\}
  + \tfrac{\varrho}{2}\|\theta\|^2\Big]}_{H + r_2},
\end{equation}
so that the data-fitting term is $G_{\mathrm{data}} - H_{\mathrm{data}}
= \tfrac{1}{n}\sum_i \min\{r_i^2, \tau_\ell^2\}$ and, with
$\lambda_1 = \lambda_2$, the regularizer is the capped-$\ell_1$ penalty. The
matching ridge on both sides supplies Assumption~\ref{assu:sc}, and $F \ge 0$ since
$\lambda_2 \le \lambda_1$.

\paragraph{Relation to the assumptions.} This split violates the assumptions of every
theorem in the paper, in three ways.
(i) $H_{\mathrm{data}}(\theta) = \tfrac1n\sum_i\max\{r_i^2 - \tau_\ell^2, 0\}$, with
$r_i = y_i - a_i^\top\theta$, is convex but not differentiable: the derivative of
$r \mapsto \max\{r^2 - \tau_\ell^2, 0\}$ jumps from $0$ to $2\tau_\ell$ in absolute value at
$|r| = \tau_\ell$, so Assumption~\ref{assu:smooth-H} fails.
(ii) The per-sample gradient $-2r_i a_i\,\mathbf 1\{|r_i| > \tau_\ell\}$ has variance growing
like $\|\theta\|^2$, so no uniform bound $\sigma^2$ exists on $\R^d$.
(iii) For the single-loop method, $F_\gamma$ is unbounded below for every $\gamma > 0$,
whichever component carries the concave-side ridge. Let $Q_d := \tfrac2n A^\top A$ be the
Hessian of $G_{\mathrm{data}}$, which is positive definite almost surely since $n > d$ and the
$a_i$ are Gaussian, and let $q := G_{\mathrm{data}} + \tfrac\varrho2\|\cdot\|^2$, with Hessian
$Q = Q_d + \varrho I$ and minimizer $x^q$. \emph{Upper bound on the convex side:} taking
$u = (I + \gamma Q)^{-1}(x + \gamma Q x^q)$, the minimizer in the envelope of $q$, gives
$(G + r_1)_\gamma(x) \le q_\gamma(x) + \lambda_1\|u\|_1 = q_\gamma(x) + \mathcal O(\|x\|)$.
\emph{Lower bound on the concave side:} since
$\max\{r^2 - \tau_\ell^2, 0\} \ge r^2 - \tau_\ell^2$, $H_{\mathrm{data}} \ge G_{\mathrm{data}} -
\tau_\ell^2$; the envelope of the nonnegative capped-$\ell_1$ correction is nonnegative; and
the ridge contributes $\tfrac\varrho2\|x\|^2$ if it is in $H$ and
$(\tfrac\varrho2\|\cdot\|^2)_\gamma(x) = \tfrac{\varrho}{2(1+\gamma\varrho)}\|x\|^2$ if it is in
$r_2$. In both cases
$H(x) + (r_2)_\gamma(x) \ge G_{\mathrm{data}}(x) + \tfrac{\varrho}{2(1+\gamma\varrho)}\|x\|^2 -
\tau_\ell^2$. Hence $F_\gamma(x) \le \tfrac12 x^\top M x + \mathcal O(\|x\|)$ with
$M := Q(I+\gamma Q)^{-1} - Q_d - \tfrac{\varrho}{1+\gamma\varrho}I$. In an eigenbasis of $Q_d$
with eigenvalue $\lambda > 0$, $M$ has eigenvalue
$\phi(\lambda) = \tfrac{\lambda+\varrho}{1+\gamma(\lambda+\varrho)} - \lambda -
\tfrac{\varrho}{1+\gamma\varrho}$, and $\phi(0) = 0$,
$\phi'(\lambda) = (1+\gamma(\lambda+\varrho))^{-2} - 1 < 0$; so $M \prec 0$ and
$F_\gamma \to -\infty$ along every ray.
Consequently none of the theorems of the paper applies to this instance, and the results
below are reported as an empirical stress test only.

\paragraph{Instance.} We use a synthetic instance with known ground truth. We draw
$a_i \sim \mathcal N(0,\Sigma)$ with $\Sigma_{jk} = 0.7^{|j-k|}$,
$y_i = a_i^\top\theta^\star + \varepsilon_i$ with
$\varepsilon_i \sim \mathcal N(0,\sigma_\varepsilon^2)$,
$\sigma_\varepsilon = 0.1$, and $\theta^\star$ supported on $s = 10$ of
$d = 700$ coordinates with entries of magnitude in $[1,2]$; $n = 5000$ training
and $2000$ test points. A fraction $10\%$ of the training targets are corrupted
by additive noise of magnitude $8\sigma_\varepsilon$ or more. We report the empirical
quantities $L = 12.99$, $\rho_1 = 0.216$, $\kappa = 66.3$ and $\sigma^2 = 425.8$; because of
(i)--(ii) above, $L$ and $\sigma^2$ are \emph{not} global constants of
Assumption~\ref{assu:smooth-H} for this split, but local estimates. We set
$\tau_\ell = 3\sigma_\varepsilon$ from the known clean noise scale, which
activates the truncation on $10.2\%$ of the training points at $\theta^\star$,
matching the corruption rate; $\lambda_1 = 0.05$, $\lambda_2 = \lambda_1/4$,
$\tau = 0.1$, $\varrho = 0.02$. All methods are initialized at the ridge
solution, which lies inside the region where the truncation is inactive.

\paragraph{Methods and protocol.} The same six methods as in
Section~\ref{sec:experiments}, plus \textbf{SDCA-damped}: no momentum
($\alpha = 1$) but with $\eta$ tuned. Comparing MDCA against SDCA-damped
separates the momentum estimator from the damping (each method's $\eta$ is tuned
separately), and comparing SDCA-damped against SDCA isolates the damping. Constant per-iteration
batch $b = 32$, initial batch $b_0 = 256$, $T = 1000$ iterations, median and IQR
over five seeds. Hyperparameters are selected by grid search on a separate seed,
at a fixed budget of stochastic gradient evaluations rather than at a fixed
iteration count, so that methods spending more gradients per iteration are not
advantaged.

\paragraph{Results.} MDCA-Polyak attains the lowest training objective
($0.5868$, against $F(\theta^\star) = 0.5699$), the lowest stationarity at the
$F$-best iterate ($0.049$, versus $0.076$ for DCA-PAGE, $0.111$ for MDCA-MVR and
$0.201$ for SDCA-damped) and the lowest test RMSE ($0.143$ versus $0.293$ for
DCA-PAGE), while using $128{,}256$ stochastic gradient evaluations against
DCA-PAGE's $177{,}728$. With five seeds these are descriptive observations. In these runs,
momentum improves on the damped momentum-free baseline by a factor of about four in median
stationarity, and the constant-batch estimator reaches a lower median stationarity than
DCA-PAGE at a smaller budget. Because the instance is outside the assumptions, these
observations neither confirm nor contradict the theory; we do not offer a mechanistic
explanation of the DCA-PAGE gap, which would require measuring the estimator error directly.

Both single-loop variants plateau early and do not descend further, at
stationarity $\approx 1.3$; their test RMSE is large ($43.7$ for S-MDCA-Polyak,
$178.4$ for SMAG) although their
objective values remain finite. Theorem~\ref{thm:smdca} does not apply here: by (iii) above,
$F_\gamma$ is unbounded below for every $\gamma > 0$, so Assumption~\ref{assu:Fgamma-bbel}
fails and the single-loop iteration has no bounded-below surrogate to converge on. Even when
Assumption~\ref{assu:Fgamma-bbel} holds, fixed-$\gamma$ stationarity of $F_\gamma$ need not
imply proximity to a critical point of $F$ (Remark~\ref{rem:gamma-tight}).

\begin{figure}[t]
\centering
\includegraphics[width=\linewidth]{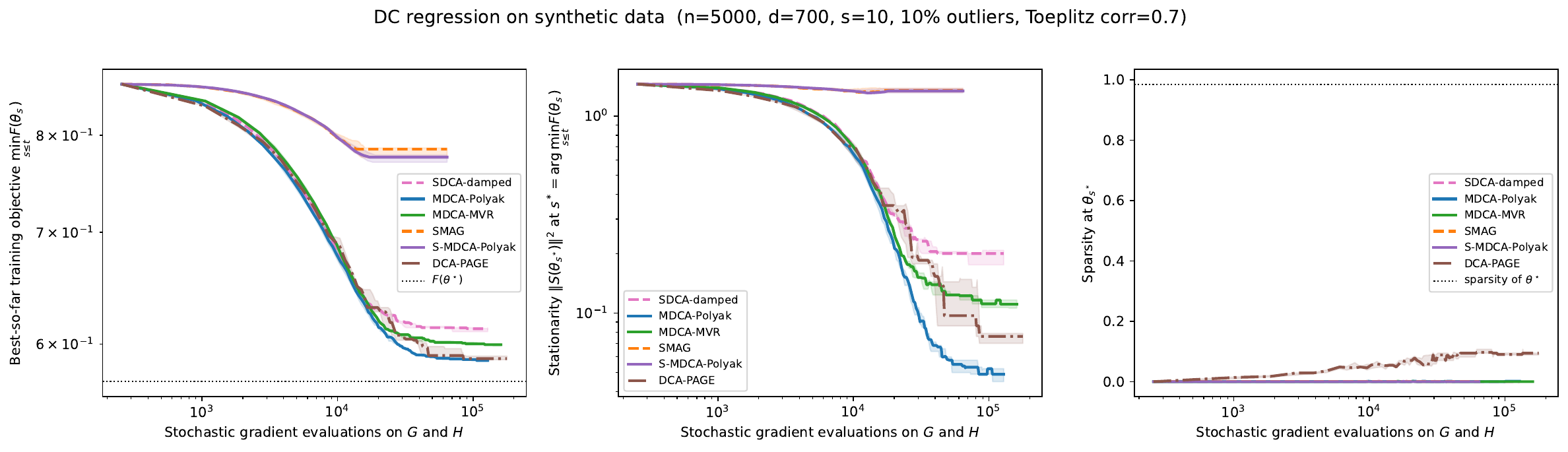}
\caption{Robust sparse regression on synthetic data ($n = 5000$, $d = 700$,
$s = 10$, $10\%$ corrupted targets, Toeplitz correlation $0.7$). Median over five
seeds; IQR shaded. Out-of-assumption stress test (see text). \textit{Left:} training objective, with $F(\theta^\star)$
marked. \textit{Center:} stationarity at the $F$-best iterate. \textit{Right:}
sparsity at the $F$-best iterate, with the sparsity of $\theta^\star$ marked; see
the discussion in the text. Horizontal axis: stochastic gradient evaluations on
$G$ and $H$ combined. Undamped SDCA diverged on all five seeds and is omitted.}
\label{fig:synthetic}
\end{figure}

Undamped SDCA, with $\eta_t \equiv 1$, diverged on all five seeds and is omitted
from Figure~\ref{fig:synthetic}, whereas SDCA-damped, which differs only in
$\eta$, diverged on none. At this noise level ($\sigma/\sqrt b \approx 3.6$
against $\|\nabla F(\theta^0)\| = \mathcal O(1)$) the undamped inner solve is
based on an estimate dominated by noise. This is consistent with, though not a test of, the
role of damping in the analysis (Section~\ref{sec:algorithm}), which requires the drift
$\|x^{t+1} - x^t\|^2$ to be small relative to the momentum parameter.

\section{Proof of Lemma~\ref{lem:polyak-var}}
\label{app:lem-polyak-proof}

We give the steps in full; this is the detailed expansion of the proof sketch in
Section~\ref{sec:variance}.

\begin{proof}
Throughout, all expectations not labelled $\E_t$ are taken over the full history.

\smallskip\noindent\textbf{Step 1 (centring).} Define
$\zeta^{t+1} := \hat\nabla H_{\mathcal I^{t+1}}(x^{t+1}) - \nabla H(x^{t+1})$.
Since the batch $\mathcal I^{t+1}$ is sampled fresh and is conditionally independent of $\mathcal F_t$ given $x^{t+1}$,
\[
\E_t[\zeta^{t+1}] = 0,
\qquad
\E_t\|\zeta^{t+1}\|^2 \le \sigma^2/b
\]
by Assumption~\ref{assu:smooth-H} and the iid mini-batch identity.

\smallskip\noindent\textbf{Step 2 (rewrite the Polyak update).}
The Polyak update is
$g^{t+1} = (1 - \alpha_t)\, g^t + \alpha_t \cdot \hat\nabla H_{\mathcal I^{t+1}}(x^{t+1})$.
Using $\hat\nabla H_{\mathcal I^{t+1}}(x^{t+1}) = \nabla H(x^{t+1}) + \zeta^{t+1}$, we obtain
\begin{align*}
g^{t+1} &= (1 - \alpha_t)\, g^t + \alpha_t\, \nabla H(x^{t+1}) + \alpha_t\, \zeta^{t+1}.
\end{align*}
Subtracting $\nabla H(x^{t+1})$ from both sides and grouping,
\begin{align*}
e^{t+1} &= g^{t+1} - \nabla H(x^{t+1}) \\
        &= (1 - \alpha_t)\, g^t + \alpha_t\, \nabla H(x^{t+1}) + \alpha_t\, \zeta^{t+1} - \nabla H(x^{t+1}) \\
        &= (1 - \alpha_t)\, g^t - (1 - \alpha_t)\, \nabla H(x^{t+1}) + \alpha_t\, \zeta^{t+1} \\
        &= (1 - \alpha_t)\,\big[g^t - \nabla H(x^{t+1})\big] + \alpha_t\, \zeta^{t+1} \\
        &= (1 - \alpha_t)\,\big[\underbrace{g^t - \nabla H(x^t)}_{= e^t} + \nabla H(x^t) - \nabla H(x^{t+1})\big] + \alpha_t\, \zeta^{t+1}.
\end{align*}

\smallskip\noindent\textbf{Step 3 (square and use $\E_t[\zeta^{t+1}] = 0$).}
The first term in the last line is $\mathcal F_t$-measurable, so by the bias-variance
decomposition,
\[
\E_t\|e^{t+1}\|^2 = \big\|(1 - \alpha_t)[e^t + \nabla H(x^t) - \nabla H(x^{t+1})]\big\|^2 + \alpha_t^2\, \E_t\|\zeta^{t+1}\|^2.
\]
The second term is bounded by Step~1: $\alpha_t^2\, \E_t\|\zeta^{t+1}\|^2 \le \alpha_t^2 \sigma^2/b$.

\smallskip\noindent\textbf{Step 4 (control the first term via Young's inequality).}
We use the inequality
$\|a + c\|^2 \le (1 + \theta)\|a\|^2 + (1 + 1/\theta)\|c\|^2$
for any $\theta > 0$, which follows from
$\|a + c\|^2 = \|a\|^2 + 2\langle a, c\rangle + \|c\|^2$ together with
$2\langle a, c\rangle \le \theta\|a\|^2 + \|c\|^2/\theta$. Apply this with
$a = e^t$, $c = \nabla H(x^t) - \nabla H(x^{t+1})$, and $\theta = \alpha_t/(1 - \alpha_t)$
(valid since $\alpha_t \in (0, 1)$; the case $\alpha_t = 1$ is handled separately at the end):
\[
1 + \theta = \frac{1}{1 - \alpha_t}, \qquad 1 + \tfrac{1}{\theta} = \frac{1}{\alpha_t}.
\]
Thus
\[
\big\|e^t + \nabla H(x^t) - \nabla H(x^{t+1})\big\|^2 \le \tfrac{1}{1 - \alpha_t}\, \|e^t\|^2 + \tfrac{1}{\alpha_t}\, \|\nabla H(x^t) - \nabla H(x^{t+1})\|^2.
\]
Multiplying by $(1 - \alpha_t)^2$:
\[
(1-\alpha_t)^2\, \|e^t + \nabla H(x^t) - \nabla H(x^{t+1})\|^2 \le (1 - \alpha_t)\, \|e^t\|^2 + \tfrac{(1-\alpha_t)^2}{\alpha_t}\, \|\nabla H(x^t) - \nabla H(x^{t+1})\|^2.
\]
By Assumption~\ref{assu:smooth-H},
$\|\nabla H(x^t) - \nabla H(x^{t+1})\|^2 \le L^2 \|x^{t+1} - x^t\|^2$.
Also $(1-\alpha_t)^2/\alpha_t \le 1/\alpha_t$, giving
\[
(1-\alpha_t)^2\, \|e^t + \nabla H(x^t) - \nabla H(x^{t+1})\|^2 \le (1 - \alpha_t)\, \|e^t\|^2 + \tfrac{L^2}{\alpha_t}\, \|x^{t+1} - x^t\|^2.
\]

\smallskip\noindent\textbf{Step 5 (assemble).}
Combining Steps~3--4,
\[
\E_t\|e^{t+1}\|^2 \le (1 - \alpha_t)\, \|e^t\|^2 + \tfrac{L^2}{\alpha_t}\, \|x^{t+1} - x^t\|^2 + \tfrac{\alpha_t^2 \sigma^2}{b}.
\]
Taking the unconditional expectation gives~\eqref{eq:polyak-var-rec}.
The case $\alpha_t = 1$ gives directly $e^{t+1} = \zeta^{t+1}$ so
$V^{t+1} \le \sigma^2/b$, consistent with~\eqref{eq:polyak-var-rec} at $\alpha_t = 1$.
\end{proof}

\section{Proof of Lemma~\ref{lem:mvr-var}}
\label{app:lem-mvr-proof}

\begin{proof}
The improvement over Lemma~\ref{lem:polyak-var} is the bias coefficient $L^2/b$ instead of
$L^2/\alpha_t$, which permits much smaller $\alpha_t$ at the same descent rate.

\smallskip\noindent\textbf{Step 1 (centred-batch and averaged smoothness).}
Define $\xi^{t+1}(x) := \hat\nabla H_{\mathcal I^{t+1}}(x) - \nabla H(x)$.
For any $\mathcal F_t$-measurable $x$, $\E_t[\xi^{t+1}(x)] = 0$ and
$\E_t\|\xi^{t+1}(x)\|^2 \le \sigma^2/b$. We claim that for any $\mathcal F_t$-measurable $x, y$,
\begin{equation}
\label{eq:avg-batch}
\E_t\|\xi^{t+1}(x) - \xi^{t+1}(y)\|^2 \le \tfrac{L^2}{b}\, \|x - y\|^2.
\end{equation}
Indeed, writing $\xi^{t+1}(x) - \xi^{t+1}(y) = \frac{1}{b}\sum_{i \in \mathcal I^{t+1}}\big[(\nabla h_i(x) - \nabla h_i(y)) - (\nabla H(x) - \nabla H(y))\big]$
and noting that the summands are iid and mean-zero, the cross terms vanish and
\begin{align*}
\E_t\|\xi^{t+1}(x) - \xi^{t+1}(y)\|^2
&= \tfrac{1}{b}\, \E_\xi\|(\nabla h_\xi(x) - \nabla h_\xi(y)) - (\nabla H(x) - \nabla H(y))\|^2 \\
&\le \tfrac{1}{b}\, \E_\xi\|\nabla h_\xi(x) - \nabla h_\xi(y)\|^2
\;\le\; \tfrac{L^2}{b}\, \|x - y\|^2,
\end{align*}
where the second-to-last inequality uses $\E\|Z - \E Z\|^2 \le \E\|Z\|^2$ and the last uses
Assumption~\ref{assu:avg-smooth}.

\smallskip\noindent\textbf{Step 2 (rewrite the MVR update).}
Substituting $\hat\nabla H_{\mathcal I^{t+1}}(x) = \nabla H(x) + \xi^{t+1}(x)$ into the MVR
update and grouping deterministic and noise terms,
\begin{align*}
g^{t+1}
&= (1 - \alpha_t)\big[g^t + \nabla H(x^{t+1}) - \nabla H(x^t) + \xi^{t+1}(x^{t+1}) - \xi^{t+1}(x^t)\big] \\
&\quad + \alpha_t\big[\nabla H(x^{t+1}) + \xi^{t+1}(x^{t+1})\big]
= (1 - \alpha_t)\big[g^t - \nabla H(x^t)\big] + \nabla H(x^{t+1}) + \Xi_t,
\end{align*}
where we used $(1-\alpha_t)\nabla H(x^{t+1}) + \alpha_t \nabla H(x^{t+1}) = \nabla H(x^{t+1})$ and set
\[
\Xi_t := (1 - \alpha_t)\big[\xi^{t+1}(x^{t+1}) - \xi^{t+1}(x^t)\big] + \alpha_t\, \xi^{t+1}(x^{t+1}).
\]
Therefore $e^{t+1} = (1 - \alpha_t)\, e^t + \Xi_t$.

\smallskip\noindent\textbf{Step 3 (variance bound on $\Xi_t$).}
$\E_t[\Xi_t] = 0$ by linearity. By $\|a + c\|^2 \le 2\|a\|^2 + 2\|c\|^2$,
\eqref{eq:avg-batch} with $(x, y) = (x^{t+1}, x^t)$, the single-point variance bound, and
$(1-\alpha_t)^2 \le 1$,
\[
\E_t\|\Xi_t\|^2 \le \tfrac{2L^2}{b}\, \|x^{t+1} - x^t\|^2 + \tfrac{2\alpha_t^2 \sigma^2}{b}.
\]

\smallskip\noindent\textbf{Step 4 (assemble).}
By bias-variance with the $\mathcal F_t$-measurable $(1 - \alpha_t) e^t$ and zero-mean $\Xi_t$,
$\E_t\|e^{t+1}\|^2 = (1 - \alpha_t)^2\|e^t\|^2 + \E_t\|\Xi_t\|^2$. Using
$(1 - \alpha_t)^2 \le (1 - \alpha_t)$ for $\alpha_t \in [0, 1]$ and substituting Step~3,
then taking the unconditional expectation, gives~\eqref{eq:mvr-var-rec}.
\end{proof}

\section{Proof of Lemma~\ref{lem:descent}}
\label{app:lem-descent-proof}

\begin{proof}
We proceed in six steps, fully expanding all algebraic manipulations. Throughout, $w^t$ is
an arbitrary element of $\partial r_2(x^t)$; \emph{no differentiability of $r_2$ is used}.

\smallskip\noindent\textbf{Step 1 (strong convexity of $H + r_2$ at $x^t$).}
By Assumption~\ref{assu:sc}, $H + r_2$ is $\rho_2$-strongly convex. Since $H$ is finite,
convex and differentiable and $w^t \in \partial r_2(x^t)$, the sum rule gives
$\nabla H(x^t) + w^t \in \partial(H + r_2)(x^t)$. The $\rho_2$-strong-convexity inequality
at base $x^t$ with this subgradient, evaluated at $\tilde x^{t+1}$, reads:
\begin{equation}
\label{eq:appA-1}
(H+r_2)(\tilde x^{t+1}) \ge (H+r_2)(x^t) + \langle\nabla H(x^t) + w^t,\; \tilde x^{t+1} - x^t\rangle + \tfrac{\rho_2}{2}\|\tilde x^{t+1} - x^t\|^2.
\end{equation}

\smallskip\noindent\textbf{Step 2 (optimality of $\tilde x^{t+1}$ and strong convexity of $G + r_1$).}
By line 4 of Algorithm~\ref{alg:mdca}, $\tilde x^{t+1}$ minimizes the strongly convex
function $\Psi^t(y) := G(y) + r_1(y) - \langle g^t + w^t,\, y\rangle$. The first-order
optimality condition is $g^t + w^t \in \partial(G + r_1)(\tilde x^{t+1})$. By
$\rho_1$-strong convexity of $G + r_1$, evaluated at $x^t$ with this subgradient at $\tilde x^{t+1}$:
\begin{equation}
\label{eq:appA-2}
(G+r_1)(x^t) \ge (G+r_1)(\tilde x^{t+1}) + \langle g^t + w^t,\, x^t - \tilde x^{t+1}\rangle + \tfrac{\rho_1}{2}\|x^t - \tilde x^{t+1}\|^2.
\end{equation}

\smallskip\noindent\textbf{Step 3 (combine into a bound at $\tilde x^{t+1}$).}
Writing $F = (G+r_1) - (H+r_2)$ and using \eqref{eq:appA-2} for the first bracket and
\eqref{eq:appA-1} for the second,
\begin{align}
F(\tilde x^{t+1}) - F(x^t)
&\le \langle g^t + w^t - \nabla H(x^t) - w^t,\; \tilde x^{t+1} - x^t\rangle - \tfrac{\rho}{2}\|\tilde x^{t+1} - x^t\|^2 \notag \\
&= \langle g^t - \nabla H(x^t),\; \tilde x^{t+1} - x^t\rangle - \tfrac{\rho}{2}\|\tilde x^{t+1} - x^t\|^2,
\label{eq:descent-tildex}
\end{align}
where the $w^t$ terms cancel exactly, the same subgradient appearing in both inequalities.
The displayed inequality~\eqref{eq:descent-tildex} is the standard DCA-style descent at
$\tilde x^{t+1}$.

\smallskip\noindent\textbf{Step 4 (extend to $x^{t+1}$ via convex combination).}
We apply the same two inequalities at $x^{t+1} = x^t + \eta_t(\tilde x^{t+1} - x^t)$:
\begin{align}
(H+r_2)(x^{t+1}) &\ge (H+r_2)(x^t) + \eta_t\,\langle\nabla H(x^t) + w^t,\, \tilde x^{t+1} - x^t\rangle + \tfrac{\rho_2 \eta_t^2}{2}\|\tilde x^{t+1} - x^t\|^2,
\label{eq:appA-1-eta}
\end{align}
using $x^{t+1} - x^t = \eta_t(\tilde x^{t+1} - x^t)$ for the inner product and squared norm.
The subgradient $g^t + w^t$ lies in $\partial(G + r_1)(\tilde x^{t+1})$, not necessarily in
$\partial(G + r_1)(x^{t+1})$, so instead of a subgradient inequality at $x^{t+1}$ we use the
standard ``three-point'' inequality for strongly convex functions along the segment:
\begin{equation}
\label{eq:appA-2-eta}
(G+r_1)(x^{t+1}) \le (1 - \eta_t)(G+r_1)(x^t) + \eta_t (G+r_1)(\tilde x^{t+1}) - \tfrac{\eta_t(1-\eta_t)\rho_1}{2}\|\tilde x^{t+1} - x^t\|^2.
\end{equation}

\smallskip\noindent\textbf{Step 5 (bound on $F(x^{t+1}) - F(x^t)$).}
Combining \eqref{eq:appA-1-eta}, \eqref{eq:appA-2-eta} and \eqref{eq:appA-2},
\begin{align*}
F(x^{t+1}) - F(x^t)
&\le \eta_t\,\langle g^t + w^t,\, \tilde x^{t+1} - x^t\rangle - \tfrac{\eta_t \rho_1}{2}\|\tilde x^{t+1} - x^t\|^2
- \tfrac{\eta_t(1-\eta_t)\rho_1}{2}\|\tilde x^{t+1} - x^t\|^2 \\
&\quad - \eta_t\,\langle \nabla H(x^t) + w^t,\, \tilde x^{t+1} - x^t\rangle - \tfrac{\rho_2 \eta_t^2}{2}\|\tilde x^{t+1} - x^t\|^2 \\
&= \eta_t\,\langle g^t - \nabla H(x^t),\, \tilde x^{t+1} - x^t\rangle - \tfrac{\eta_t}{2}\big[\rho_1(2 - \eta_t) + \eta_t \rho_2\big]\|\tilde x^{t+1} - x^t\|^2,
\end{align*}
the $w^t$ terms cancelling as in Step~3.

\smallskip\noindent\textbf{Step 6 (lower bound the curvature coefficient using $\rho_1 \ge \rho_2$).}
Since
\[
\rho_1(2 - \eta_t) + \eta_t \rho_2 - (\rho_1 + \rho_2)
= (1 - \eta_t)(\rho_1 - \rho_2) \ge 0
\]
for $\eta_t \le 1$ and $\rho_1 \ge \rho_2$, Step~5 gives
$F(x^{t+1}) - F(x^t) \le \eta_t\langle g^t - \nabla H(x^t), \tilde x^{t+1} - x^t\rangle - \tfrac{\eta_t \rho}{2}\|\tilde x^{t+1} - x^t\|^2$.
Applying Young's inequality
$\langle a, c\rangle \le \tfrac{1}{2\beta_t}\|a\|^2 + \tfrac{\beta_t}{2}\|c\|^2$
with $a = g^t - \nabla H(x^t)$, $c = \tilde x^{t+1} - x^t$ and grouping the
$\|\tilde x^{t+1} - x^t\|^2$ terms,
\[
F(x^{t+1}) - F(x^t) \le \tfrac{\eta_t}{2\beta_t}\,\|g^t - \nabla H(x^t)\|^2 - \tfrac{\eta_t(\rho - \beta_t)}{2}\,\|\tilde x^{t+1} - x^t\|^2 .
\]
Taking the unconditional expectation gives~\eqref{eq:descent-template}.
\end{proof}

\section{Proof of Theorem~\ref{thm:polyak-rate}}
\label{app:thm-polyak-proof}

\begin{proof}
We carry out every algebraic step.

\smallskip\noindent\textbf{Step 1 (telescope descent).}
Apply Lemma~\ref{lem:descent} with $\beta_t = \rho/2$:
\[
\E[F(x^{t+1}) - F(x^t)] \le \tfrac{\eta}{\rho}\, V^t - \tfrac{\eta\, \rho}{4}\, \tilde D^t.
\]
Sum from $t = 0$ to $T - 1$. The left-hand side telescopes to $\E[F(x^T) - F(x^0)] \ge -\Delta_0$,
so, solving for $\sum \tilde D^t$:
\begin{equation}
\label{eq:proof-thm1-D}
\sum_{t=0}^{T-1}\tilde D^t \;\le\; \frac{4\, \Delta_0}{\eta\, \rho} + \frac{4}{\rho^2}\sum_{t=0}^{T-1} V^t.
\end{equation}
Since $\eta = \min\{1,\alpha\rho/(4L)\}$ we have $1/\eta \le 1 + 4L/(\alpha\rho)$, hence
\begin{equation}
\label{eq:proof-thm1-D-capped}
\sum_{t=0}^{T-1}\tilde D^t \;\le\; \frac{4\Delta_0}{\rho} + \frac{16L\Delta_0}{\alpha\rho^2}
+ \frac{4}{\rho^2}\sum_{t=0}^{T-1}V^t .
\end{equation}
The extra $4\Delta_0/\rho$ contributes $\mathcal O(\Delta_0/(\rho T))$ to the average, a term
of order $1/T$ that does not affect the $\eps$-dependence. Steps~2--3 below
use $\eta$ only through upper bounds on $\eta^2$ and are unaffected by the cap.

\smallskip\noindent\textbf{Step 2 (telescope Polyak variance).}
By Lemma~\ref{lem:polyak-var} with $\alpha_t = \alpha$, $D^t = \eta^2 \tilde D^t$, summing
from $t = 0$ to $T-1$ and using $V^T \ge 0$,
\begin{equation}
\label{eq:proof-thm1-V}
\alpha\sum_{t=0}^{T-1} V^t \le V^0 + \tfrac{L^2 \eta^2}{\alpha}\sum_{t=0}^{T-1}\tilde D^t + \tfrac{T \alpha^2 \sigma^2}{b}.
\end{equation}

\smallskip\noindent\textbf{Step 3 (eliminate $\sum \tilde D^t$).}
Substitute~\eqref{eq:proof-thm1-D} into~\eqref{eq:proof-thm1-V} and move the $\sum V^t$ term
to the left:
\[
\Bigl(\alpha - \tfrac{4 L^2 \eta^2}{\alpha \rho^2}\Bigr)\sum V^t \le V^0 + \tfrac{4 L^2 \eta \Delta_0}{\alpha \rho} + \tfrac{T \alpha^2 \sigma^2}{b}.
\]
Since $\eta \le \alpha\rho/(4L)$ we have $\eta^2 \le \alpha^2\rho^2/(16L^2)$, so
$4 L^2 \eta^2/(\alpha \rho^2) \le \alpha/4$ and the left-hand-side coefficient is at least
$\alpha - \alpha/4 = 3\alpha/4 > 0$; likewise
$4 L^2 \eta \Delta_0/(\alpha\rho) \le L \Delta_0$. Hence
\begin{equation}
\label{eq:proof-thm1-sumV}
\sum_{t=0}^{T-1} V^t \le \tfrac{4(V^0 + L \Delta_0)}{3\alpha} + \tfrac{4 T\alpha\, \sigma^2}{3 b}.
\end{equation}

\smallskip\noindent\textbf{Step 4 (bound on $\sum \tilde D^t$).}
Substitute~\eqref{eq:proof-thm1-sumV} into~\eqref{eq:proof-thm1-D-capped}:
\begin{equation}
\label{eq:proof-thm1-sumD}
\sum \tilde D^t \le \frac{4\Delta_0}{\rho}
+ \frac{16 L \Delta_0 + \tfrac{16}{3}(V^0 + L\Delta_0)}{\alpha\rho^2}
+ \frac{16\, T\alpha\sigma^2}{3\rho^2 b} .
\end{equation}

\smallskip\noindent\textbf{Step 5 (convert to $\mathcal D^2$ via Lemma~\ref{lem:dist-bound}).}
By Lemma~\ref{lem:dist-bound},
$\sum_{t}\E[\mathcal D(\tilde x^{t+1})^2] \le 2(L + L_{r_2})^2 \sum \tilde D^t + 2\sum V^t$.
Write $x_L := (L+L_{r_2})^2$ for brevity. Substituting~\eqref{eq:proof-thm1-sumD}
and~\eqref{eq:proof-thm1-sumV},
\begin{align*}
\sum_t \E[\mathcal D(\tilde x^{t+1})^2]
&\le 2x_L\Bigl[\frac{4\Delta_0}{\rho} + \frac{16 L\Delta_0 + \tfrac{16}{3}(V^0 + L\Delta_0)}{\alpha\rho^2}
+ \frac{16\,T\alpha\sigma^2}{3\rho^2 b}\Bigr]
+ 2\Bigl[\frac{4(V^0 + L\Delta_0)}{3\alpha} + \frac{4T\alpha\sigma^2}{3b}\Bigr] \\
&\le \frac{8 x_L \Delta_0}{\rho}
+ \Bigl(\frac{128}{3}\frac{x_L}{\rho^2} + \frac{8}{3}\Bigr)\frac{V^0 + L\Delta_0}{\alpha}
+ \Bigl(\frac{32}{3}\frac{x_L}{\rho^2} + \frac{8}{3}\Bigr)\frac{T\alpha\sigma^2}{b},
\end{align*}
where in the second line we used $16L\Delta_0 + \tfrac{16}{3}(V^0 + L\Delta_0)
\le \tfrac{64}{3}(V^0 + L\Delta_0)$, valid since $L\Delta_0 \le V^0 + L\Delta_0$. Dividing
by $T$ gives exactly~\eqref{eq:polyak-davg-bound} with the constants $C_1, C_2$ stated in
Theorem~\ref{thm:polyak-rate}.

\smallskip\noindent\textbf{Step 6 (optimize $\alpha$).}
The bound is of the form $A_1/(\alpha T) + A_2 \alpha\sigma^2/b$ plus an $\mathcal O(1/T)$
term, where $A_1 = C_1(V^0 + L\Delta_0)$ and $A_2 = C_2$. Differentiating in $\alpha$
gives $\alpha^* = \sqrt{A_1 b/(A_2 \sigma^2 T)}$ and minimum value
$2\sqrt{A_1 A_2}\, \sigma\, b^{-1/2}\, T^{-1/2}$ when $\alpha^* \le 1$; when $\alpha^* > 1$,
$\alpha = 1$ gives $A_1/T + A_2\sigma^2/b < 2A_1/T$. This is~\eqref{eq:polyak-two-regime};
making each of its three terms at most $\eps^2/3$ gives the condition on $T$ stated in
Theorem~\ref{thm:polyak-rate}.

\smallskip\noindent\textbf{Step 7 (sample complexity).}
Since $V^0$ enters only through the sum $V^0 + L\Delta_0$, no $\eps$-dependent seeding is
required. With $b_0 = 1$ we have $V^0 \le \sigma^2$ and
$T = \mathcal O\bigl(\sigma^2(\sigma^2 + L\Delta_0)/(b\eps^4)\bigr)$, so
$T b + b_0 = \mathcal O\bigl(\sigma^2(\sigma^2+L\Delta_0)/\eps^4\bigr)$. Taking instead the
constant $b_0 = \lceil\sigma^2/(L\Delta_0)\rceil$ gives $V^0 \le L\Delta_0$ and
$T b + b_0 = \mathcal O(L\Delta_0\sigma^2/\eps^4)$.
\end{proof}

\section{Proof of Theorem~\ref{thm:mvr-rate}}
\label{app:thm-mvr-proof}

\begin{proof}
The proof structure is parallel to Theorem~\ref{thm:polyak-rate} with the variance
recursion replaced by Lemma~\ref{lem:mvr-var}.

\smallskip\noindent\textbf{Step 1 (telescope descent).}
Equations \eqref{eq:proof-thm1-D} and \eqref{eq:proof-thm1-D-capped} hold with
$1/\eta \le 1 + 4L/(\rho\sqrt{\alpha b})$ in place of $1/\eta \le 1 + 4L/(\alpha\rho)$.

\smallskip\noindent\textbf{Step 2 (telescope MVR variance).}
By Lemma~\ref{lem:mvr-var}, summing and using $V^T \ge 0$,
\[
\alpha \sum V^t \le V^0 + \tfrac{2 L^2 \eta^2}{b}\sum \tilde D^t + \tfrac{2 T\alpha^2 \sigma^2}{b}.
\]

\smallskip\noindent\textbf{Step 3 (eliminate $\sum \tilde D^t$).}
Substituting~\eqref{eq:proof-thm1-D} and moving the $\sum V^t$ term to the left,
\[
\Bigl(\alpha - \tfrac{8 L^2 \eta^2}{b\rho^2}\Bigr)\sum V^t \le V^0 + \tfrac{8 L^2\eta\Delta_0}{b\rho} + \tfrac{2 T\alpha^2 \sigma^2}{b}.
\]
Since $\eta \le \rho\sqrt{\alpha b}/(4L)$ we have $\eta^2 \le \rho^2\alpha b/(16 L^2)$, so
$8 L^2 \eta^2/(b\rho^2) \le \alpha/2$ and the left coefficient is at least $\alpha/2 > 0$;
likewise $8 L^2 \eta \Delta_0/(b\rho) \le 2 L \Delta_0\sqrt{\alpha}/\sqrt b$. Hence
\begin{equation}
\label{eq:proof-thm2-sumV}
\sum V^t \le \tfrac{2 V^0}{\alpha} + \tfrac{4 L\Delta_0}{\sqrt{\alpha b}} + \tfrac{4 T\alpha\sigma^2}{b}.
\end{equation}

\smallskip\noindent\textbf{Step 4 ($\sum \tilde D^t$ and conversion to $\mathcal D^2$).}
Substituting~\eqref{eq:proof-thm2-sumV} into~\eqref{eq:proof-thm1-D-capped} and applying
Lemma~\ref{lem:dist-bound}, then dividing by $T$ and collecting like terms,
\[
\tfrac{1}{T}\sum \E[\mathcal D(\tilde x^{t+1})^2] = \mathcal O\!\Bigl(\tfrac{V^0}{\alpha T} + \tfrac{L\Delta_0}{\sqrt{\alpha b}\, T} + \tfrac{\Delta_0}{\rho T} + \tfrac{\alpha\sigma^2}{b}\Bigr).
\]

\smallskip\noindent\textbf{Step 5 (optimize $\alpha$).}
The dominant trade-off is between $L\Delta_0/(\sqrt{\alpha b}\, T)$ and $\alpha\sigma^2/b$.
Differentiating in $\alpha$: $-L\Delta_0/(2 \alpha^{3/2}\sqrt b\, T) + \sigma^2/b = 0$, giving
$\alpha^{3/2} = L\Delta_0\sqrt b/(2\sigma^2 T)$, i.e.\
$\alpha^* = \bigl(L\Delta_0\sqrt b/(2\sigma^2 T)\bigr)^{2/3}$. Substituting,
\[
\tfrac{L\Delta_0}{\sqrt{\alpha^* b}\, T}
= \mathcal O\!\bigl((L\Delta_0)^{2/3}\sigma^{2/3}\, b^{-2/3}\, T^{-2/3}\bigr).
\]
Setting this to $\eps^2$ gives $T = \mathcal O(L\Delta_0\sigma/(b\eps^3))$ and total cost
$2 T b = \mathcal O(L\Delta_0\sigma/\eps^3)$.

\smallskip\noindent\textbf{Step 6 (initialization).}
At $T = \Theta(L\Delta_0\sigma/(b\eps^3))$ the optimized
$\alpha^* = \bigl(\sqrt b\,L\Delta_0/(2\sigma^2T)\bigr)^{2/3} = \Theta(b\eps^2/\sigma^2)$, so
$\alpha^* T = \Theta(L\Delta_0/(\sigma\eps))$ and the term $V^0/(\alpha^* T)$ is at most
$\eps^2$ if and only if $V^0 \lesssim L\Delta_0\eps/\sigma$, i.e.\ if and only if
$b_0 \ge \lceil \sigma^3/(L\Delta_0\eps)\rceil$, which is~\eqref{eq:mvr-b0}. The warm-up
cost $b_0 = \Theta(\eps^{-1})$ is dominated by $2Tb = \Theta(\eps^{-3})$ whenever
$\eps \le L\Delta_0/\sigma$, so the total remains $\mathcal O(L\Delta_0\sigma\eps^{-3})$. Note
also that $\alpha^* = b\eps^2/(2^{2/3}\sigma^2) \le 1$ holds iff $\eps^2 \le 2^{2/3}\sigma^2/b$,
which is part of the small-$\eps$ regime of Theorem~\ref{thm:mvr-rate}; the requirement
$\Delta_0/(\rho T) \lesssim \eps^2$ at this $T$ reads $\eps \lesssim \rho L\sigma/b$.
\end{proof}

\section{Inexact inner solves}
\label{app:inexact}

Throughout this appendix line~5 of Algorithm~\ref{alg:mdca} is replaced by
$x^{t+1} = x^t + \eta_t(\hat x^{t+1} - x^t)$, where $\hat x^{t+1}$ is an approximate
minimizer of
$\Psi^t(y) := G(y) + r_1(y) - \langle g^t + w^t, y\rangle$, whose exact minimizer is
$\tilde x^{t+1}$. We write $f := G + r_1$, $\varphi := H + r_2$ (so $F = f - \varphi$), and
\[
\hat D^t := \E\|\hat x^{t+1} - x^t\|^2, \qquad D^t = \E\|x^{t+1} - x^t\|^2 = \eta_t^2\hat D^t .
\]
The inner solver may be randomized; we only require that the batch $\mathcal I^{t+1}$ be
drawn independently of the history and of the inner solver's randomness at step $t$. Then
$x^{t+1}$ is measurable with respect to the $\sigma$-field on which the proofs of
Lemmas~\ref{lem:polyak-var} and~\ref{lem:mvr-var} condition, and both variance recursions
hold verbatim with $D^t = \eta_t^2\hat D^t$; they use the iterates only through
$\|x^{t+1} - x^t\|$. The accuracies $\nu_t$, $\delta_t$ below may be random.

\subsection{Descent with function-value accuracy}

\begin{lemma}[Inexact descent inequality]
\label{lem:inexact-descent}
Let Assumptions~\ref{assu:wellposed} and~\ref{assu:sc} hold and suppose
$\Psi^t(\hat x^{t+1}) - \min_y\Psi^t(y) \le \nu_t$. Then
$\|\hat x^{t+1} - \tilde x^{t+1}\|^2 \le 2\nu_t/\rho_1$, and for every $\eta_t \in (0,1]$ and
$\beta_t \in (0, \rho/2)$,
\begin{equation}
\label{eq:inexact-descent}
\E\bigl[F(x^{t+1}) - F(x^t)\bigr] \;\le\; \frac{\eta_t}{2\beta_t}\,V^t
\;-\; \frac{\eta_t(\rho/2 - \beta_t)}{2}\,\hat D^t \;+\; 2\eta_t\,\E\nu_t .
\end{equation}
With $\hat\rho := \rho/2$ this is the descent template~\eqref{eq:descent-template} with $\rho$
replaced by $\hat\rho$ and an additive $2\eta_t\E\nu_t$.
\end{lemma}

\begin{proof}
Fix $t$ and drop the index on $\nu$; write $\hat x = \hat x^{t+1}$, $\tilde x = \tilde x^{t+1}$,
$d := \|\hat x - x^t\|$. By $\rho_1$-strong convexity of $\Psi^t$ and optimality of $\tilde x$,
$\Psi^t(y) \ge \Psi^t(\tilde x) + \tfrac{\rho_1}{2}\|y - \tilde x\|^2$ for all $y$. With
$y = \hat x$ this gives $\|\hat x - \tilde x\|^2 \le 2\nu/\rho_1$. With $y = x^t$, together
with $\Psi^t(\tilde x) \ge \Psi^t(\hat x) - \nu$ and
$\|x^t - \tilde x\|^2 \ge \tfrac12 d^2 - \|\hat x - \tilde x\|^2 \ge \tfrac12 d^2 - 2\nu/\rho_1$,
\[
\Psi^t(x^t) \;\ge\; \Psi^t(\hat x) - 2\nu + \tfrac{\rho_1}{4}\,d^2,
\qquad\text{i.e.}\qquad
f(\hat x) \le f(x^t) + \langle g^t + w^t, \hat x - x^t\rangle + 2\nu - \tfrac{\rho_1}{4}d^2 .
\]
Since $x^{t+1} = (1-\eta_t)x^t + \eta_t\hat x$, $\rho_1$-strong convexity of $f$ along the
segment gives
$f(x^{t+1}) \le (1-\eta_t)f(x^t) + \eta_t f(\hat x) - \tfrac{\rho_1}{2}\eta_t(1-\eta_t)d^2$,
hence
\[
f(x^{t+1}) - f(x^t) \le \eta_t\langle g^t + w^t, \hat x - x^t\rangle + 2\eta_t\nu
- \eta_t\bigl[\tfrac{\rho_1}{4} + \tfrac{\rho_1}{2}(1-\eta_t)\bigr]d^2 .
\]
By $\rho_2$-strong convexity of $\varphi$ with the subgradient
$\nabla H(x^t) + w^t \in \partial\varphi(x^t)$,
$\varphi(x^{t+1}) \ge \varphi(x^t) + \eta_t\langle\nabla H(x^t) + w^t, \hat x - x^t\rangle
+ \tfrac{\rho_2}{2}\eta_t^2 d^2$. Subtracting, the $w^t$ terms cancel and
\[
F(x^{t+1}) - F(x^t) \le \eta_t\langle e^t, \hat x - x^t\rangle + 2\eta_t\nu
- \tfrac{\eta_t}{2}\bigl[(\tfrac32 - \eta_t)\rho_1 + \eta_t\rho_2\bigr]d^2 .
\]
For $\eta_t \le 1$ and $\rho_1 \ge \rho_2$,
$(\tfrac32 - \eta_t)\rho_1 + \eta_t\rho_2 = \tfrac{\rho_1}{2} + (1-\eta_t)\rho_1 + \eta_t\rho_2
\ge \tfrac{\rho_1}{2} + \rho_2 \ge \tfrac{\rho}{2}$, so the curvature term is at most
$-\tfrac{\eta_t\rho}{4}d^2$. Young's inequality
$\langle e^t, \hat x - x^t\rangle \le \|e^t\|^2/(2\beta_t) + \beta_t d^2/2$ and expectation
give~\eqref{eq:inexact-descent}.
\end{proof}

\subsection{Residual criterion under function-value accuracy}

\begin{theorem}[Inexact Polyak MDCA, DCA residual]
\label{thm:inexact-residual}
Let Assumptions~\ref{assu:wellposed}, \ref{assu:sc}, \ref{assu:smooth-H} and~\ref{assu:bbel}
hold, and run Algorithm~\ref{alg:mdca} with the Polyak update, inexact solves with
$\Psi^t(\hat x^{t+1}) - \min\Psi^t \le \nu_t$, and
\[
\eta_t \equiv \eta := \min\Bigl\{1,\ \frac{\alpha\rho}{8L}\Bigr\},\qquad \beta_t \equiv \rho/4,
\qquad \alpha_t \equiv \alpha \in (0,1].
\]
Let $\tau$ be uniform on $\{0,\dots,T-1\}$, $A := V^0 + L\Delta_0$ and
$\bar\nu := \tfrac1T\sum_{t<T}\E\nu_t$. Then
\[
\E\bigl[R_{\min}(x^\tau)\bigr] \le \E\bigl[R(x^\tau; w^\tau)\bigr]
\;\le\; \frac{\hat K_1 A}{\alpha T} + \frac{24\Delta_0}{\rho T} + \frac{\hat K_2\,\alpha\sigma^2}{b}
+ \frac{72\,\bar\nu}{\rho_1},
\]
with $\hat\kappa := 48/\rho^2 + 3/\rho_1^2$, $\hat K_1 := 192/\rho^2 + \tfrac43\hat\kappa$ and
$\hat K_2 := \tfrac43\hat\kappa$. With $\alpha$ chosen as in
Theorem~\ref{thm:polyak-rate-gap} (with $\hat K_i$ in place of $K_i$), the right-hand side
is at most $2\sqrt{\hat K_1\hat K_2A\sigma^2/(bT)} + 2\hat K_1A/T + 24\Delta_0/(\rho T) +
72\bar\nu/\rho_1$; hence $\E[R_{\min}(x^\tau)] \le \eps^2$ whenever
$\bar\nu \le \rho_1\eps^2/288$ and
$T \ge \max\{64\hat K_1\hat K_2A\sigma^2/(b\eps^4),\ 8\hat K_1A/\eps^2,\ 96\Delta_0/(\rho\eps^2)\}$,
the same $\eps$-dependence as Theorem~\ref{thm:polyak-rate-gap}.
\end{theorem}

\begin{proof}
Write $\hat\rho := \rho/2$, so that $\beta_t = \hat\rho/2$ and $\eta \le \alpha\hat\rho/(4L)$.
\emph{Descent.} Lemma~\ref{lem:inexact-descent} gives
$\E[F(x^{t+1}) - F(x^t)] \le \tfrac{\eta}{\hat\rho}V^t - \tfrac{\eta\hat\rho}{4}\hat D^t + 2\eta\E\nu_t$;
telescoping with $\E F(x^T) \ge F^*$,
\begin{equation}
\label{eq:inexact-sumD}
\sum_{t<T}\hat D^t \le \frac{4\Delta_0}{\eta\hat\rho} + \frac{4}{\hat\rho^2}\sum_{t<T}V^t
+ \frac{8}{\hat\rho}\sum_{t<T}\E\nu_t .
\end{equation}
\emph{Variance.} As in Step~2 of Appendix~\ref{app:thm-polyak-proof},
$\alpha\sum V^t \le V^0 + \tfrac{L^2\eta^2}{\alpha}\sum\hat D^t + T\alpha^2\sigma^2/b$.
Insert~\eqref{eq:inexact-sumD} and use $\eta^2 \le \alpha^2\hat\rho^2/(16L^2)$: the three
contributions are at most $\tfrac{\alpha}{4}\sum V^t$, $L\Delta_0$ and
$\tfrac{\alpha\hat\rho}{2}\sum\E\nu_t$, so
\begin{equation}
\label{eq:inexact-sumV}
\sum_{t<T}V^t \le \frac{4A}{3\alpha} + \frac{4T\alpha\sigma^2}{3b} + \frac{2\hat\rho}{3}\sum_{t<T}\E\nu_t .
\end{equation}
\emph{Residual.} With $\bar x^{t+1} = T(x^t, w^t)$, \eqref{eq:argmin-contraction} gives
$\|\bar x^{t+1} - \tilde x^{t+1}\| \le \|e^t\|/\rho_1$, and
$\|\tilde x^{t+1} - \hat x^{t+1}\| \le \sqrt{2\nu_t/\rho_1}$ by
Lemma~\ref{lem:inexact-descent}. The triangle inequality and $(a+c+d)^2 \le 3(a^2+c^2+d^2)$
give $R(x^t; w^t) \le 3\|\hat x^{t+1} - x^t\|^2 + 3\|e^t\|^2/\rho_1^2 + 6\nu_t/\rho_1$.
Summing and inserting~\eqref{eq:inexact-sumD} and then~\eqref{eq:inexact-sumV},
\[
\sum_{t<T}\E R(x^t;w^t) \le \frac{12\Delta_0}{\eta\hat\rho} + \hat\kappa\sum V^t
+ \Bigl(\frac{24}{\hat\rho} + \frac{6}{\rho_1}\Bigr)\sum\E\nu_t
\le \frac{24\Delta_0}{\eta\rho} + \hat\kappa\Bigl(\frac{4A}{3\alpha} + \frac{4T\alpha\sigma^2}{3b}\Bigr)
+ c_\nu\sum\E\nu_t,
\]
where $\hat\kappa = 12/\hat\rho^2 + 3/\rho_1^2$ and
$c_\nu = 32/\hat\rho + 2\hat\rho/\rho_1^2 + 6/\rho_1$. Since $\rho_2 \le \rho_1$ we have
$\rho_1/2 \le \hat\rho \le \rho_1$, so $c_\nu \le 64/\rho_1 + 2/\rho_1 + 6/\rho_1 = 72/\rho_1$.
Finally $1/\eta \le 1 + 8L/(\alpha\rho)$ gives
$24\Delta_0/(\eta\rho) \le 24\Delta_0/\rho + 192L\Delta_0/(\alpha\rho^2)$ and
$L\Delta_0 \le A$. Dividing by $T$ and using $R_{\min}(x^t) \le R(x^t;w^t)$ proves the bound;
the complexity statement follows by making each of the four terms at most $\eps^2/4$.
\end{proof}

\begin{remark}[MVR]
\label{rem:inexact-mvr}
Under additionally Assumption~\ref{assu:avg-smooth}, the same argument with
$\eta := \min\{1, \rho\sqrt{\alpha b}/(8L)\}$ and Lemma~\ref{lem:mvr-var} gives the bound of
Theorem~\ref{thm:mvr-rate-gap} (with modified absolute constants) plus an additional
$c'_\nu\bar\nu$, where
$c'_\nu = 48/\hat\rho + 6\hat\rho/\rho_1^2 + 6/\rho_1 \le 108/\rho_1$. The
$\mathcal O(L\Delta_0\sigma\eps^{-3})$ dependence is preserved if $\bar\nu \lesssim \rho_1\eps^2$.
\end{remark}

\subsection{Critical distance under an approximate optimality certificate}

Function-value accuracy does not control $\mathcal D$; see the counterexample in
Section~\ref{sec:algorithm}. We therefore require the inner solver to return a pair
$(\hat x^{t+1}, s^{t+1})$ with
\begin{equation}
\label{eq:kkt-cert}
s^{t+1} \in \partial(G + r_1)(\hat x^{t+1}), \qquad \|s^{t+1} - (g^t + w^t)\| \le \delta_t .
\end{equation}
Since $s^{t+1} - (g^t + w^t) \in \partial\Psi^t(\hat x^{t+1})$ and $\Psi^t$ is
$\rho_1$-strongly convex, $\Psi^t(\hat x^{t+1}) - \min\Psi^t \le \delta_t^2/(2\rho_1)$, so
Lemma~\ref{lem:inexact-descent} applies with $\nu_t = \delta_t^2/(2\rho_1)$.

\begin{theorem}[Inexact Polyak MDCA, critical distance]
\label{thm:inexact-crit}
Let Assumptions~\ref{assu:wellposed}, \ref{assu:sc}, \ref{assu:smooth-H},
\ref{assu:smooth-r2} and~\ref{assu:bbel} hold, let $w^t = \nabla r_2(x^t)$, and let the inner
solver satisfy~\eqref{eq:kkt-cert}. With $\eta$, $\beta_t$, $\alpha$ as in
Theorem~\ref{thm:inexact-residual}, $x_L := (L + L_{r_2})^2$,
$\bar\delta^2 := \tfrac1T\sum_{t<T}\E\delta_t^2$ and output $\hat x^{\tau+1}$,
\[
\E\bigl[\mathcal D(\hat x^{\tau+1})^2\bigr] \le \frac{\hat C_1 A}{\alpha T}
+ \frac{24x_L\Delta_0}{\rho T} + \frac{\hat C_2\,\alpha\sigma^2}{b}
+ \Bigl(\frac{32x_L}{\rho_1^2} + 4\Bigr)\bar\delta^2,
\qquad \hat C_1 := \frac{256x_L}{\rho^2} + 4,\quad \hat C_2 := \frac{64x_L}{\rho^2} + 4 .
\]
Thus the $\mathcal O(\eps^{-4})$ dependence of Theorem~\ref{thm:polyak-rate} is preserved when
$\bar\delta^2 \lesssim \eps^2/(1 + x_L/\rho_1^2)$.
\end{theorem}

\begin{proof}
By~\eqref{eq:kkt-cert},
$\mathcal D(\hat x^{t+1}) \le \|\nabla H(\hat x^{t+1}) + \nabla r_2(\hat x^{t+1}) - s^{t+1}\|
\le (L + L_{r_2})\|\hat x^{t+1} - x^t\| + \|e^t\| + \delta_t$, so
$\mathcal D(\hat x^{t+1})^2 \le 3x_L\|\hat x^{t+1} - x^t\|^2 + 3\|e^t\|^2 + 3\delta_t^2$. By
\eqref{eq:inexact-sumD}--\eqref{eq:inexact-sumV} with $\hat\rho = \rho/2$ and
$\nu_t = \delta_t^2/(2\rho_1)$,
\[
\sum_t\E\mathcal D(\hat x^{t+1})^2 \le \frac{24x_L\Delta_0}{\eta\rho}
+ \Bigl(\frac{48x_L}{\rho^2} + 3\Bigr)\sum V^t + \frac{48x_L}{\rho}\sum\E\nu_t + 3\sum\E\delta_t^2 ,
\]
and inserting~\eqref{eq:inexact-sumV} the $\nu$-coefficient becomes
$64x_L/\rho + \rho$. With $\nu_t = \delta_t^2/(2\rho_1)$, $\rho \ge \rho_1$ and $\rho \le 2\rho_1$,
it contributes at most $(32x_L/\rho_1^2 + 1)\sum\E\delta_t^2$. The $\Delta_0$ and $\sigma^2$
terms are handled as in Theorem~\ref{thm:inexact-residual}
($1/\eta \le 1 + 8L/(\alpha\rho)$, $L\Delta_0 \le A$), giving $\hat C_1$ and $\hat C_2$.
Divide by $T$.
\end{proof}

\section{Proof of Theorem~\ref{thm:smdca}}
\label{app:thm-smdca-proof}

We prove the Polyak-momentum case in detail; the MVR case is handled at the end of
this section by replacing the Polyak variance recursion with Lemma~\ref{lem:mvr-var}.
The proof is self-contained and uses no external results.

\paragraph{Setup.}
Recall $F_\gamma(x) := (G+r_1)_\gamma(x) - H(x) - (r_2)_\gamma(x)$, $L_\gamma = L + 2/\gamma$,
$u^*(x) := \prox_{\gamma(G+r_1)}(x)$, $z^*(x) := \prox_{\gamma r_2}(x)$, and the gradient
identity $\nabla F_\gamma(x) = \tfrac{1}{\gamma}(z^*(x) - u^*(x)) - \nabla H(x)$ from
\eqref{eq:grad-Fgamma}. The algorithm computes $z^t = z^*(x^t)$ exactly, advances $u^t$
by one stochastic prox-gradient step, and forms the search direction
$\mathcal G^t = (z^t - u^{t+1})/\gamma - g^t$. Define
\[
\phi^t(u) := G(u) + \tfrac{1}{2\gamma}\|u - x^t\|^2, \qquad
L_\phi := L_G + \tfrac{1}{\gamma}, \qquad
\mu_\phi := \tfrac{1}{\gamma},
\]
so that $\phi^t$ is $L_\phi$-smooth and $\mu_\phi$-strongly convex (more generally
$\mu_\phi = \mu_G + 1/\gamma$ if $G$ is $\mu_G$-strongly convex; only
$\mu_\phi \ge 1/\gamma$ is used below), and $u^*(x^t) = \arg\min_u\{\phi^t(u) + r_1(u)\}$. Note that
the schedule $\eta^1 \le \gamma/(1+\gamma L_G)$ of~\eqref{eq:smdca-params} is exactly
$\eta^1 \le 1/L_\phi$, and implies $\eta^1 \le \gamma$. Define the random sequences
\begin{align*}
e^t &:= g^t - \nabla H(x^t), \qquad
\xi^t := \mathcal G^t - \nabla F_\gamma(x^t), \qquad
P^t := \E\|\nabla F_\gamma(x^t)\|^2, \\
V^t &:= \E\|e^t\|^2, \qquad
\mathcal E^t := \E\|u^t - u^*(x^t)\|^2, \qquad
\hat E^t := \E\|u^{t+1} - u^*(x^t)\|^2, \qquad
D^t := \E\|x^{t+1} - x^t\|^2 .
\end{align*}
A direct computation using \eqref{eq:grad-Fgamma} and $z^t = z^*(x^t)$ gives
\begin{equation}
\label{eq:xi-decomp}
\xi^t = \tfrac{1}{\gamma}\bigl(u^*(x^t) - u^{t+1}\bigr) - \bigl(g^t - \nabla H(x^t)\bigr)
      = -\tfrac{1}{\gamma}\bigl(u^{t+1} - u^*(x^t)\bigr) - e^t .
\end{equation}

\paragraph{Step 1: Smoothness descent.}
By $L_\gamma$-smoothness of $F_\gamma$ (Assumption~\ref{assu:Fgamma-bbel} guarantees
$F_\gamma^* > -\infty$) and the update $x^{t+1} = x^t - \eta^0 \mathcal G^t$,
\[
F_\gamma(x^{t+1}) - F_\gamma(x^t)
\le -\eta^0 \langle \nabla F_\gamma(x^t), \mathcal G^t \rangle + \tfrac{L_\gamma (\eta^0)^2}{2}\|\mathcal G^t\|^2.
\]
Writing $\langle \nabla F_\gamma, \mathcal G^t\rangle = \|\nabla F_\gamma\|^2 + \langle \nabla F_\gamma, \xi^t\rangle$,
$\|\mathcal G^t\|^2 \le 2\|\nabla F_\gamma\|^2 + 2\|\xi^t\|^2$, and
$-\langle \nabla F_\gamma, \xi^t\rangle \le \tfrac{1}{2}\|\nabla F_\gamma\|^2 + \tfrac{1}{2}\|\xi^t\|^2$,
\[
F_\gamma(x^{t+1}) - F_\gamma(x^t)
\le -\eta^0\big(\tfrac{1}{2} - L_\gamma \eta^0\big)\|\nabla F_\gamma(x^t)\|^2
   + \eta^0\big(\tfrac{1}{2} + L_\gamma\eta^0\big)\|\xi^t\|^2 .
\]
Whenever $\eta^0 \le 1/(4 L_\gamma)$ we have $L_\gamma\eta^0 \le 1/4$, so
\begin{equation}
\label{eq:descent-step}
\E\bigl[F_\gamma(x^{t+1}) - F_\gamma(x^t)\bigr]
\le -\tfrac{\eta^0}{4}P^t + \tfrac{3\eta^0}{4}\E\|\xi^t\|^2 .
\end{equation}

\paragraph{Step 2: Polyak variance recursion.}
By Lemma~\ref{lem:polyak-var} (whose proof in Appendix~\ref{app:lem-polyak-proof} does not
depend on the algorithm's choice of $\eta^0$),
\begin{equation}
\label{eq:V-rec-app}
V^{t+1} \le (1-\alpha) V^t + \tfrac{L^2}{\alpha}\, D^t + \alpha^2 \tfrac{\sigma^2}{b}.
\end{equation}

(Here the mini-batch noise $\hat\nabla H_{\mathcal I^{t+1}}(x^{t+1}) - \nabla H(x^{t+1})$
of Appendix~\ref{app:lem-polyak-proof}, Step~1, is centred conditionally on
$\mathcal F_t \vee \sigma(\xi^G_t)$, with respect to which $x^{t+1}$ is measurable; this is
legitimate because $\mathcal I^{t+1}$ is drawn after $\xi^G_t$ and is independent of it
given the past, by Assumption~\ref{assu:G-oracle}.)

\paragraph{Step 3: Proximal-tracking recursion.}
\begin{lemma}[One inner step]
\label{lem:smdca-tracking-step}
Let $\eta^1 \le 1/L_\phi$ and set $q := \eta^1\mu_\phi \in (0,1]$. Then under
Assumption~\ref{assu:G-oracle},
\[
\E_t\|u^{t+1} - u^*(x^t)\|^2 \;\le\; (1 - q)\,\|u^t - u^*(x^t)\|^2 + (\eta^1)^2 M^2 .
\]
\end{lemma}

\begin{proof}
Write $u^* := u^*(x^t) = \arg\min_u\{\phi^t(u) + r_1(u)\}$, so
$0 \in \nabla\phi^t(u^*) + \partial r_1(u^*)$ and hence
$u^* = \prox_{\eta^1 r_1}\bigl(u^* - \eta^1\nabla\phi^t(u^*)\bigr)$. Let
$\varepsilon_t := \nabla G(u^t;\xi^G_t) - \nabla G(u^t)$, so that $\E_t[\varepsilon_t] = 0$,
$\E_t\|\varepsilon_t\|^2 \le M^2$ (Assumption~\ref{assu:G-oracle}) and
$\nabla\phi^t(u^t) + \varepsilon_t = \nabla G(u^t;\xi^G_t) + (u^t - x^t)/\gamma$. By
non-expansiveness of $\prox_{\eta^1 r_1}$,
\[
\|u^{t+1} - u^*\| \le \bigl\|A_t - \eta^1\varepsilon_t\bigr\|,
\qquad
A_t := \bigl(u^t - \eta^1\nabla\phi^t(u^t)\bigr) - \bigl(u^* - \eta^1\nabla\phi^t(u^*)\bigr),
\]
where $A_t$ is $\mathcal F_t$-measurable. Since $\E_t[\varepsilon_t] = 0$ the cross term
vanishes and $\E_t\|u^{t+1}-u^*\|^2 \le \|A_t\|^2 + (\eta^1)^2 M^2$. For $\|A_t\|^2$, put
$d := u^t - u^*$ and $\Delta := \nabla\phi^t(u^t) - \nabla\phi^t(u^*)$. Using
$\|\Delta\|^2 \le L_\phi\langle d, \Delta\rangle$ ($L_\phi$-smoothness and convexity of
$\phi^t$) and $\eta^1 L_\phi \le 1$,
\[
\|A_t\|^2 = \|d\|^2 - 2\eta^1\langle d,\Delta\rangle + (\eta^1)^2\|\Delta\|^2
\le \|d\|^2 - \eta^1\bigl(2 - \eta^1 L_\phi\bigr)\langle d,\Delta\rangle
\le \|d\|^2 - \eta^1\langle d,\Delta\rangle .
\]
Finally $\langle d,\Delta\rangle \ge \mu_\phi\|d\|^2$ gives
$\|A_t\|^2 \le (1-\eta^1\mu_\phi)\|d\|^2$, and
$\eta^1\mu_\phi \le \mu_\phi/L_\phi \le 1$ keeps the factor nonnegative.
\end{proof}

\noindent
Only the strong-convexity modulus of the \emph{smooth part} $\phi^t$ may appear in
Lemma~\ref{lem:smdca-tracking-step}. Using the modulus of the full inner objective $\phi^t + r_1$ would be incorrect: when $r_1$ carries the strong convexity, that
modulus can exceed $L_\phi$ and the contraction factor becomes negative -- already for
deterministic quadratic instances.

Taking total expectations in Lemma~\ref{lem:smdca-tracking-step},
\begin{equation}
\label{eq:hatE}
\hat E^t \;\le\; (1-q)\,\mathcal E^t + (\eta^1)^2 M^2 .
\end{equation}
To advance the base point, use $1$-Lipschitzness of $u^*$ (the prox is non-expansive), so
$\|u^*(x^{t+1}) - u^*(x^t)\| \le \|x^{t+1}-x^t\|$, and Young's inequality with parameter
$\nu = q/2$:
\[
\|u^{t+1} - u^*(x^{t+1})\|^2 \le (1+\tfrac{q}{2})\|u^{t+1} - u^*(x^t)\|^2
+ (1+\tfrac{2}{q})\|x^{t+1}-x^t\|^2 .
\]
Since $(1+\tfrac q2)(1-q) = 1 - \tfrac q2 - \tfrac{q^2}{2} \le 1-\tfrac q2$,
$(1+\tfrac2q) \le \tfrac3q$ and $1+\tfrac q2 \le \tfrac32$ for $q \in (0,1]$, combining with
\eqref{eq:hatE} gives
\begin{equation}
\label{eq:E-rec}
\mathcal E^{t+1} \;\le\; (1-c)\,\mathcal E^t + \frac{3}{2c}\,D^{t} + \tfrac{3}{2}(\eta^1)^2 M^2,
\qquad c := \tfrac{q}{2} = \tfrac{1}{2}\eta^1 \mu_\phi .
\end{equation}
The drift term is the \emph{current} $D^t$: no lagged $D^{t-1}$ appears and no shifted-index
argument is needed. The initial value is
$\mathcal E^0 = \|x^0 - u^*(x^0)\|^2 = \gamma^2\|p_1^0\|^2$ with
$p_1^0 := (x^0 - u^*(x^0))/\gamma \in \partial(G+r_1)(u^*(x^0))$, a finite constant.
Combining \eqref{eq:xi-decomp} with \eqref{eq:hatE} and Young's inequality,
\begin{equation}
\label{eq:xi-norm-new}
\E\|\xi^t\|^2 \;\le\; \frac{2}{\gamma^2}\,\hat E^t + 2V^t
\;\le\; \frac{2}{\gamma^2}\mathcal E^t + \frac{2(\eta^1)^2M^2}{\gamma^2} + 2V^t .
\end{equation}

\paragraph{Step 4: Drift bound.}
$D^t = (\eta^0)^2\,\E\|\mathcal G^t\|^2 \le 2(\eta^0)^2(P^t + \E\|\xi^t\|^2)$, so
by~\eqref{eq:xi-norm-new},
\begin{equation}
\label{eq:D-bound}
D^t \;\le\; 2(\eta^0)^2 P^t + \frac{4(\eta^0)^2}{\gamma^2}\mathcal E^t + 4(\eta^0)^2 V^t
+ \frac{4(\eta^0)^2(\eta^1)^2M^2}{\gamma^2} .
\end{equation}

\paragraph{Step 5: Lyapunov function.}
Define
\[
\mathcal V^t := \E[F_\gamma(x^t) - F_\gamma^*] + \frac{a_E}{\gamma^2}\,\mathcal E^t + a_V\, V^t,
\qquad a_E := \frac{3\eta^0}{c}, \quad a_V := \frac{3\eta^0}{\alpha}.
\]
Substituting \eqref{eq:xi-norm-new} into \eqref{eq:descent-step} and adding
\eqref{eq:E-rec} and \eqref{eq:V-rec-app}, the coefficients of $\mathcal E^t/\gamma^2$ and
$V^t$ become $\tfrac{3\eta^0}{2} - a_E c = -\tfrac{3\eta^0}{2}$ and
$\tfrac{3\eta^0}{2} - a_V\alpha = -\tfrac{3\eta^0}{2}$ respectively, so
\begin{equation}
\label{eq:Ls-onestep-raw}
\mathcal V^{t+1} - \mathcal V^t \le -\frac{\eta^0}{4}P^t - \frac{3\eta^0}{2\gamma^2}\mathcal E^t
- \frac{3\eta^0}{2}V^t + \kappa_D\, D^t + \mathcal N,
\end{equation}
\[
\kappa_D := \frac{9\eta^0}{2c^2\gamma^2} + \frac{3\eta^0 L^2}{\alpha^2},
\qquad
\mathcal N := \frac{3\eta^0}{2}\Bigl(1+\frac{3}{c}\Bigr)\frac{(\eta^1)^2M^2}{\gamma^2}
+ \frac{3\eta^0\alpha\sigma^2}{b} .
\]
We now absorb $\kappa_D D^t$ using \eqref{eq:D-bound}. The three requirements are
$2\kappa_D(\eta^0)^2 \le \tfrac{\eta^0}{8}$ (for $P^t$) and
$4\kappa_D(\eta^0)^2 \le \tfrac{3\eta^0}{2}$ (for $\mathcal E^t/\gamma^2$ and for $V^t$);
the first is the binding one, i.e.\ $\kappa_D\eta^0 \le 1/16$. Allocating $1/32$ to each
summand of $\kappa_D$ and using $c\gamma = \eta^1/2$ (valid since $\mu_\phi = 1/\gamma$), it
suffices that
\begin{equation}
\label{eq:eta0-constraints}
\eta^0 \le \frac{1}{4L_\gamma}, \qquad
\eta^0 \le \frac{c\gamma}{12} = \frac{\eta^1}{24}, \qquad
\eta^0 \le \frac{\alpha}{L\sqrt{96}} ,
\end{equation}
which is the schedule~\eqref{eq:smdca-params}. Under \eqref{eq:eta0-constraints} the
residual noise generated by the absorption is
$\kappa_D \cdot 4(\eta^0)^2(\eta^1)^2M^2/\gamma^2 \le \tfrac{\eta^0}{4}(\eta^1)^2M^2/\gamma^2$,
whence
\begin{equation}
\label{eq:Ls-onestep}
\mathcal V^{t+1} - \mathcal V^t \;\le\; -\frac{\eta^0}{8}P^t + \mathcal N
+ \frac{\eta^0}{4}\frac{(\eta^1)^2M^2}{\gamma^2} .
\end{equation}

\paragraph{Step 6: Telescope.}
Summing \eqref{eq:Ls-onestep} over $t = 0,\dots,T-1$ and using $\mathcal V^T \ge 0$ together with
$\mathcal V^0 = \Delta_\gamma + \tfrac{a_E}{\gamma^2}\mathcal E^0 + a_V V^0$, where
\[
\frac{a_E}{\gamma^2}\mathcal E^0 = \frac{3\eta^0}{c}\|p_1^0\|^2
= \frac{6\eta^0\gamma}{\eta^1}\|p_1^0\|^2 \le \frac{\gamma}{4}\|p_1^0\|^2
\quad\text{(by $\eta^0 \le \eta^1/24$)},
\]
and dividing by $\eta^0 T/8$,
\begin{equation}
\label{eq:VF-bound}
\frac{1}{T}\sum_{t=0}^{T-1} P^t \;\le\;
\frac{8\,\bar\Ls}{\eta^0 T} + \frac{24\, V^0}{\alpha T}
+ \frac{86\,\eta^1 M^2}{\gamma} + \frac{24\,\alpha\sigma^2}{b},
\qquad \bar\Ls := \Delta_\gamma + \tfrac{\gamma}{4}\|p_1^0\|^2 ,
\end{equation}
where the constant $\bar\Ls$ bounds the non-variance part of the initial Lyapunov value,
$\mathcal V^0 - a_V V^0 = \Delta_\gamma + \tfrac{a_E}{\gamma^2}\mathcal E^0 \le \bar\Ls$; the
remaining contribution $a_V V^0$ is reported separately as the term $24V^0/(\alpha T)$.
The prox-noise constant follows from $1/c = 2\gamma/\eta^1$ and $\eta^1 \le \gamma$:
\[
\frac{8\mathcal N}{\eta^0} + \frac{2(\eta^1)^2M^2}{\gamma^2}
= \Bigl(12 + \frac{36}{c}\Bigr)\frac{(\eta^1)^2M^2}{\gamma^2} + \frac{24\alpha\sigma^2}{b}
+ \frac{2(\eta^1)^2M^2}{\gamma^2}
\le \frac{(12 + 72 + 2)\,\eta^1M^2}{\gamma} + \frac{24\alpha\sigma^2}{b} .
\]
Both $\bar\Ls$ and all constants are independent of $\eps$, and no term involving $\Delta_0$
appears.

\paragraph{Step 7: Optimal choice of $\eta^1, \alpha, \eta^0, T$ (Polyak).}
With $\hat x$ uniform over $\{x^0,\dots,x^{T-1}\}$, $\E\|\nabla F_\gamma(\hat x)\|^2$ equals
the left-hand side of \eqref{eq:VF-bound}. We make each of the four terms $\le \eps^2/4$.

\textbf{(a) $\eta^1$ from $G$-side noise.}
$86\eta^1M^2/\gamma \le \eps^2/4$ gives
\begin{equation}
\label{eq:smdca-eta1-opt}
\eta^1 \le \frac{\gamma\,\eps^2}{344\, M^2} \;=\; \mathcal O(\gamma\eps^2/M^2).
\end{equation}

\textbf{(b) $\alpha$ from $H$-side noise.}
$24\alpha\sigma^2/b \le \eps^2/4$ gives $\alpha \le b\eps^2/(96\sigma^2) = \mathcal O(b\eps^2/\sigma^2)$.

\textbf{(c) $\eta^0$ from \eqref{eq:eta0-constraints}.}
With $\gamma = \mathcal O(1)$, $1/L_\gamma = \mathcal O(1)$, while
$\eta^1/24 = \mathcal O(\gamma\eps^2/M^2)$ and
$\alpha/(L\sqrt{96}) = \mathcal O(b\eps^2/(L\sigma^2))$. For sufficiently small $\eps$ both
are below $1/(4L_\gamma)$, so
\begin{equation}
\label{eq:smdca-eta0-polyak}
\eta^0 = \min\bigl\{\eta^1/24,\ \alpha/(L\sqrt{96})\bigr\}
= \Theta\!\bigl(\min(\gamma\eps^2/M^2,\ b\eps^2/(L\sigma^2))\bigr) = \Theta(\eps^2).
\end{equation}

\textbf{(d) $T$ and $b_0$.}
$8\bar\Ls/(\eta^0T) \le \eps^2/4$ requires
$T \ge \mathcal O\bigl(\bar\Ls/\eps^4 \cdot \max(M^2/\gamma,\ L\sigma^2/b)\bigr)$. Since
$\eta^0 \le \alpha/(L\sqrt{96})$, the requirement $24V^0/(\alpha T) \le \eps^2/4$ is implied
by the descent requirement as soon as $V^0 \lesssim L\bar\Ls$, i.e.\
$b_0 = \Theta(\sigma^2/(L\bar\Ls))$ -- a \emph{constant}, independent of $\eps$. Hence
\begin{equation}
\label{eq:smdca-T-polyak}
T = \mathcal O\!\Bigl(\frac{\bar\Ls}{\eps^4}\max\Bigl(\frac{M^2}{\gamma},\ \frac{L\sigma^2}{b}\Bigr)\Bigr),
\qquad
T b = \mathcal O\!\bigl((L\sigma^2 + bM^2/\gamma)/\eps^4\bigr).
\end{equation}

\paragraph{Step 7$'$: MVR case.}
Lemma~\ref{lem:mvr-var} replaces \eqref{eq:V-rec-app} by
$V^{t+1} \le (1-\alpha)V^t + (2L^2/b)D^t + 2\alpha^2\sigma^2/b$. In
\eqref{eq:Ls-onestep-raw} this changes only the second summand of $\kappa_D$, from
$a_V L^2/\alpha = 3\eta^0L^2/\alpha^2$ to $a_V\cdot 2L^2/b = 6\eta^0L^2/(\alpha b)$, and
doubles the $\sigma^2$ term of $\mathcal N$. The binding condition
$\kappa_D\eta^0 \le 1/16$ therefore replaces the third constraint of
\eqref{eq:eta0-constraints} by
\begin{equation}
\label{eq:eta0-constraints-mvr}
\eta^0 \le \frac{\sqrt{\alpha b}}{L\sqrt{192}},
\end{equation}
and \eqref{eq:VF-bound} holds with $24\alpha\sigma^2/b$ replaced by $48\alpha\sigma^2/b$.
Nothing else changes. In particular, no term of the form
$\mathcal O(L\Delta_0/(\sqrt{\alpha b}\,T))$ arises: the sum $\sum_t D^t$ is absorbed
self-consistently through \eqref{eq:D-bound}, which follows from the update
$x^{t+1} = x^t - \eta^0\mathcal G^t$ of Algorithm~\ref{alg:smdca}, and
Lemma~\ref{lem:descent} -- which describes the damped DCA step of
Algorithm~\ref{alg:mdca} and does \emph{not} apply to Algorithm~\ref{alg:smdca} -- is never
invoked. Optimizing as in Step~7 gives $\eta^1 = \mathcal O(\gamma\eps^2/M^2)$ and
$\alpha = \mathcal O(b\eps^2/\sigma^2)$; since
$\sqrt{\alpha b}/(L\sqrt{192}) = \Theta(b\eps/(L\sigma)) = \Theta(\eps)$ while
$\eta^1/24 = \Theta(\eps^2)$, the constraint $\eta^0 \le \eta^1/24$ binds for small $\eps$,
so $\eta^0 = \Theta(\gamma\eps^2/M^2)$ exactly as in the Polyak case and
$T = \Theta(\bar\Ls M^2/(\gamma\eps^4))$. See Remark~\ref{rem:inner-vr}.

\section{Proof of Lemma~\ref{lem:gamma-equivalence} (smoothing equivalence)}
\label{app:smoothing-eq}

We restate the lemma. Recall $u^*(x) := \prox_{\gamma(G+r_1)}(x)$ and
$z^*(x) := \prox_{\gamma r_2}(x)$.

\begin{lemma}[Smoothing equivalence, restated]
\label{lem:gamma-equivalence}
Let Assumption~\ref{assu:sc} hold. Then for any $x \in \R^d$:
\begin{enumerate}[leftmargin=*,topsep=2pt,itemsep=2pt]
\item[(i)] $\frac{1}{\gamma}(x - u^*(x)) \in \partial(G+r_1)(u^*(x))$ and
$\frac{1}{\gamma}(x - z^*(x)) \in \partial r_2(z^*(x))$.

\item[(ii)] $\nabla F_\gamma(x) = p_1 - \nabla H(x) - p_2$ with
$p_1 := (x - u^*(x))/\gamma \in \partial(G+r_1)(u^*(x))$ and
$p_2 := (x - z^*(x))/\gamma \in \partial r_2(z^*(x))$.

\item[(iii)] If $\|\nabla F_\gamma(x)\| \le \eps$ and $r_2$ is $L_{r_2}$-smooth on the
relevant domain, then $\hat x := u^*(x)$ is an $\hat\eps$-approximate critical point of $F$:
\(
\dist\!\big(0,\, \partial(G+r_1)(\hat x) - \nabla H(\hat x) - \partial r_2(\hat x)\big) \le \hat\eps,
\)
with
\[
\hat\eps \;=\; \eps \;+\; L\,\|x - u^*(x)\| \;+\; L_{r_2}\,\|u^*(x) - z^*(x)\|
       \;\le\; \eps(1 + \gamma L_{r_2}) + \gamma\big(L \|p_1\| + L_{r_2}\|\nabla H(x)\|\big).
\]
\end{enumerate}
\end{lemma}

\begin{proof}
(i) is the optimality condition of the prox map: $u^*(x) = \arg\min_y \{(G+r_1)(y) + \tfrac{1}{2\gamma}\|y - x\|^2\}$
yields $0 \in \partial(G+r_1)(u^*) + \tfrac{1}{\gamma}(u^* - x)$, equivalent to the stated
inclusion. The same argument applies to $z^*$.

(ii) With $p_1, p_2$ as defined, $p_1 - p_2 = (z^*(x) - u^*(x))/\gamma$, so by
\eqref{eq:grad-Fgamma}, $\nabla F_\gamma(x) = p_1 - p_2 - \nabla H(x)$.

(iii) Pick any $q \in \partial r_2(\hat x) = \partial r_2(u^*(x))$. Then
\[
\|p_1 - \nabla H(\hat x) - q\|
= \|(p_1 - \nabla H(x) - p_2) + (\nabla H(x) - \nabla H(\hat x)) + (p_2 - q)\|
\le \|\nabla F_\gamma(x)\| + L\|x - \hat x\| + \|p_2 - q\|,
\]
using $L$-smoothness of $H$ on the second term. For the third term, $L_{r_2}$-smoothness
of $r_2$ on the domain (which makes $\partial r_2$ single-valued and Lipschitz) gives
$\|p_2 - q\| = \|\nabla r_2(z^*(x)) - \nabla r_2(\hat x)\| \le L_{r_2}\|z^*(x) - u^*(x)\|$.
Moreover $\|x - u^*(x)\| = \gamma\|p_1\|$, and
the identity $(z^* - u^*)/\gamma - \nabla H = \nabla F_\gamma$ gives
$z^*(x) - u^*(x) = \gamma(\nabla F_\gamma(x) + \nabla H(x))$, so
$\|z^*(x) - u^*(x)\| \le \gamma(\eps + \|\nabla H(x)\|)$. Combining,
\[
\|p_1 - \nabla H(\hat x) - q\|
\le \eps + L\gamma\|p_1\| + L_{r_2}\gamma(\eps + \|\nabla H(x)\|)
= \eps(1 + \gamma L_{r_2}) + \gamma\big(L\|p_1\| + L_{r_2}\|\nabla H(x)\|\big). \qedhere
\]
\end{proof}

\noindent
For $\gamma = \mathcal O(\eps)$ the bound reduces to $\hat\eps = \mathcal O(\eps)$ provided
$\|p_1\|, \|\nabla H(x)\|$ remain bounded along the iterates. The $\gamma$-dependent terms
cannot be removed: see Remark~\ref{rem:gamma-tight}, where they are attained with equality.

\section{Inexact prox of $r_1$ and $r_2$}
\label{app:smdca-inexact-prox}

The main-text Theorem~\ref{thm:smdca} assumes $\prox_{\eta^1 r_1}$ and $\prox_{\gamma r_2}$
are computable in closed form. Since $r_1$ and $r_2$ are \emph{convex}, the relevant
examples are convex pieces: for $r_1$, $\lambda\|\cdot\|_1$, the group-$\ell_1$ norm and
indicators of convex sets with a known projection; for $r_2$, the convex correction terms in
the DC decompositions of nonconvex penalties (e.g.\ $\lambda\sum_i[|x_i| - \theta]_+$ for
capped-$\ell_1$, and the smooth convex $h_{\mathrm{MCP}}$, $h_{\mathrm{SCAD}}$ of
Section~\ref{sec:intro}), not the nonconvex penalties themselves. We now record what happens when the prox is computed only to some
accuracy.

\subsection{Inexact prox of $r_2$.}
\label{app:smdca-inexact-prox-r2}

Let $z^t$ be a deterministic approximation of $z^*(x^t) = \prox_{\gamma r_2}(x^t)$ with
controllable error
\begin{equation}
\label{eq:inexact-r2}
\|z^t - z^*(x^t)\|^2 \le \delta_{r_2}^t.
\end{equation}

The prox objective $r_2(y) + \tfrac{1}{2\gamma}\|y - x^t\|^2$ is $1/\gamma$-strongly convex,
but strong convexity alone does not give a linearly convergent method (with only a
subgradient oracle for $r_2$ it does not). Linear convergence, and hence
$\mathcal O(\log(1/\delta_{r_2}))$ iterations, holds under additional structure, e.g.\
$r_2 = r_2^{s} + r_2^{p}$ with $r_2^{s}$ convex and $L_{2}$-smooth and $\prox_{r_2^{p}}$
available in closed form; see the cost paragraph at the end of this appendix.

Since $z^t$ enters the search direction $\mathcal G^t = (z^t - u^{t+1})/\gamma - g^t$ instead
of $z^*(x^t)$, the error $\xi^t = \mathcal G^t - \nabla F_\gamma(x^t)$
(cf.~\eqref{eq:xi-decomp}) gains an additive term $(z^t - z^*(x^t))/\gamma$:
\[
\xi^t \;=\; -\tfrac{1}{\gamma}\bigl(u^{t+1}-u^*(x^t)\bigr) - e^t + \tfrac{1}{\gamma}\bigl(z^t - z^*(x^t)\bigr),
\]
hence, by Young's inequality with three terms, \eqref{eq:xi-norm-new} becomes
\begin{equation}
\label{eq:xi-norm-inexact}
\E\|\xi^t\|^2 \le \frac{3}{\gamma^2}\mathcal E^t + \frac{3(\eta^1)^2M^2}{\gamma^2} + 3V^t
+ \frac{3\,\delta_{r_2}^t}{\gamma^2}.
\end{equation}

\noindent
The subgradient substitution of Remark~\ref{rem:smdca-subgrad} is \emph{not} of this form:
it converges to a point other than $z^*(x^t)$, and is treated separately in
Appendix~\ref{app:smdca-subgrad}.

\subsection{Inexact prox of $r_1$.}
\label{app:smdca-inexact-prox-r1}

If $\prox_{\eta^1 r_1}$ is itself approximated by a routine returning $\hat u^{t+1}$ with
\begin{equation}
\label{eq:inexact-r1}
\|\hat u^{t+1} - u^{t+1}\|^2 \le \delta_{r_1}^t,
\end{equation}
then Lemma~\ref{lem:smdca-tracking-step} acquires an additive term. The naive bound
$\|a+b\|^2 \le 2\|a\|^2 + 2\|b\|^2$ must \emph{not} be used here: it would produce a leading
factor $2(1-q)$, which exceeds $1$ whenever $q < 1/2$, destroying the contraction and hence
the absorption step of Appendix~\ref{app:thm-smdca-proof}, Step~5. Instead apply Young's
inequality with parameter $q/2$:
\begin{align*}
\E_t\|\hat u^{t+1} - u^*(x^t)\|^2
&\le (1+\tfrac q2)\,\E_t\|u^{t+1} - u^*(x^t)\|^2 + (1+\tfrac2q)\,\delta_{r_1}^t \\
&\le (1+\tfrac q2)(1-q)\,\|u^t - u^*(x^t)\|^2 + (1+\tfrac q2)(\eta^1)^2M^2 + \tfrac3q\,\delta_{r_1}^t \\
&\le (1-\tfrac q2)\,\|u^t - u^*(x^t)\|^2 + \tfrac32(\eta^1)^2M^2 + \tfrac{3}{q}\,\delta_{r_1}^t .
\end{align*}
Advancing the base point as in \eqref{eq:E-rec}, now with Young parameter $\nu = q/4$ (so
that $(1+\tfrac q4)(1-\tfrac q2) \le 1 - \tfrac q4$, $(1+\tfrac4q) \le \tfrac5q$ and
$1+\tfrac q4 \le \tfrac54$ for $q \in (0,1]$), gives
\begin{equation}
\label{eq:E-rec-inexact}
\mathcal E^{t+1} \le (1-c_\delta)\,\mathcal E^t + \frac{5}{4c_\delta}\,D^t
+ 2(\eta^1)^2M^2 + \frac{1}{c_\delta}\,\delta_{r_1}^t,
\qquad c_\delta := \tfrac{q}{4} = \tfrac14\eta^1\mu_\phi ,
\end{equation}
which contracts, carries the \emph{current} drift $D^t$, and reduces to \eqref{eq:E-rec} up
to absolute constants when $\delta_{r_1}^t = 0$.

\subsection{Combined statement.}

\begin{theorem}[Single-loop with inexact prox]
\label{thm:smdca-inexact}
Under the assumptions of Theorem~\ref{thm:smdca}, suppose $z^t$ approximates
$\prox_{\gamma r_2}(x^t)$ and $\hat u^{t+1}$ approximates $\prox_{\eta^1 r_1}(\cdot)$ with
errors \eqref{eq:inexact-r2} and \eqref{eq:inexact-r1}, and write
$\bar\delta_{r_j} := \tfrac1T\sum_t \delta_{r_j}^t$. Then, with $c_\delta$ as in
\eqref{eq:E-rec-inexact},
\[
\E\|\nabla F_\gamma(\hat x)\|^2 \;\le\;
\text{\emph{(bound of Theorem~\ref{thm:smdca}, up to absolute constants)}}
\;+\; \mathcal O\!\Bigl(\frac{\bar\delta_{r_2}}{\gamma^2}\Bigr)
\;+\; \mathcal O\!\Bigl(\frac{\bar\delta_{r_1}}{c_\delta^2\,\gamma^2}\Bigr).
\]
The two error paths therefore require \emph{different} accuracies. Taking
\[
\bar\delta_{r_2} \;\lesssim\; \gamma^2\eps^2
\qquad\text{and}\qquad
\bar\delta_{r_1} \;\lesssim\; c_\delta^2\,\gamma^2\eps^2 \;=\; \Theta\!\bigl(\gamma^2\eps^6/M^4\bigr)
\]
preserves the $\eps^{-4}$ rate of Theorem~\ref{thm:smdca}. The asymmetry is structural: the
$r_2$ error enters the search direction directly through \eqref{eq:xi-norm-inexact},
whereas the $r_1$ error enters through the contracting tracking recursion
\eqref{eq:E-rec-inexact}, picking up one factor $1/c_\delta$ there and a second from the
Lyapunov weight $a_E \propto 1/c_\delta$.
\end{theorem}

\begin{proof}
Substitute \eqref{eq:xi-norm-inexact} for \eqref{eq:xi-norm-new} and
\eqref{eq:E-rec-inexact} for \eqref{eq:E-rec} in Steps~1--6 of
Appendix~\ref{app:thm-smdca-proof}, with $c$ replaced by $c_\delta$.

\emph{Effect of \eqref{eq:xi-norm-inexact}.} The coefficients of $\mathcal E^t/\gamma^2$ and
$V^t$ produced by Step~1 grow from $\tfrac{3\eta^0}{2}$ to $\tfrac{9\eta^0}{4}$, so the
Lyapunov weights become $a_E := \tfrac{9\eta^0}{2c_\delta}$ and
$a_V := \tfrac{9\eta^0}{2\alpha}$ in order to collapse those coefficients to
$-\tfrac{9\eta^0}{4}$; both remain $\Theta(\eta^0/c_\delta)$ and $\Theta(\eta^0/\alpha)$
respectively. In addition $\mathcal N$ gains the term
$\tfrac{3\eta^0}{4}\cdot\tfrac{3\delta_{r_2}^t}{\gamma^2} = \tfrac{9\eta^0}{4}\tfrac{\delta_{r_2}^t}{\gamma^2}$.

\emph{Effect of \eqref{eq:E-rec-inexact}.} The drift coefficient $\tfrac{3}{2c}$ becomes
$\tfrac{5}{4c_\delta}$, which only changes absolute constants in $\kappa_D$, and
$\mathcal N$ gains $\tfrac{a_E}{\gamma^2}\cdot\tfrac{\delta_{r_1}^t}{c_\delta}
= \Theta\bigl(\eta^0\delta_{r_1}^t/(c_\delta^2\gamma^2)\bigr)$.

\emph{Absorption and telescoping.} The condition $\kappa_D\eta^0 \le 1/16$, and hence the
form of \eqref{eq:eta0-constraints}, is unchanged; only the absolute constants in
$\kappa_D$ grow by a bounded factor, which is absorbed by tightening $\eta^0$ by the same
bounded factor. Consequently \eqref{eq:Ls-onestep} holds with the descent term
$-\tfrac{\eta^0}{8}P^t$ intact. Telescoping as in Step~6 and dividing by $\eta^0T/8$ turns
the two extra contributions to $\mathcal N$ into
$\tfrac{18\,\bar\delta_{r_2}}{\gamma^2}$ and
$\Theta\bigl(\bar\delta_{r_1}/(c_\delta^2\gamma^2)\bigr)$ respectively, which is the stated
bound. Setting each $\le \eps^2/6$ gives the stated accuracy requirements; with
$\eta^1 = \Theta(\gamma\eps^2/M^2)$ and $\mu_\phi = 1/\gamma$ one has
$c_\delta = \Theta(\eps^2/M^2)$, whence $c_\delta^2\gamma^2\eps^2 = \Theta(\gamma^2\eps^6/M^4)$.
\end{proof}

\noindent
\emph{Cost of the inner solvers.} The problems
$\min_y\{r_j(y) + \tfrac{1}{2\eta}\|y - z\|^2\}$ ($\eta \in \{\eta^1, \gamma\}$) are
$1/\eta$-strongly convex, but this alone does not yield a logarithmic cost. Suppose in
addition that $r_j = r_j^{s} + r_j^{p}$, with $r_j^{s}$ convex and $L_j$-smooth and
$\prox_{r_j^{p}}$ computable in closed form. Then proximal gradient on
$y \mapsto r_j^{s}(y) + \tfrac{1}{2\eta}\|y - z\|^2$ (smooth part, $(L_j + 1/\eta)$-smooth and
$1/\eta$-strongly convex) plus $r_j^{p}$, with step $1/(L_j + 1/\eta)$, contracts the squared
distance to the solution by $1 - 1/(1 + \eta L_j)$ per iteration. Warm-started at the
previous solution, it reaches accuracy $\Theta(\gamma^2\eps^6/M^4)$, respectively
$\Theta(\gamma^2\eps^2)$, in $\mathcal O\bigl((1 + \eta L_j)\log(1/\eps)\bigr)$ iterations
(the initial distance is bounded by the data of the problem). Without such structure, or an
alternative assumption (error bound, finite-sum structure with a linearly convergent
variance-reduced solver, etc.), we make no claim about the inner cost.

\section{The subgradient variant and the objective $\Phi_\gamma$}
\label{app:smdca-subgrad}

Throughout this section Algorithm~\ref{alg:smdca} is run with line~3 replaced by
$z^t := x^t - \gamma w^t$, $w^t \in \partial r_2(x^t)$, so that the search direction is
$\mathcal G^t = (x^t - u^{t+1})/\gamma - w^t - g^t$. Define
\[
\psi_\gamma := (G+r_1)_\gamma - H, \qquad
\Phi_\gamma := \psi_\gamma - r_2 = (G+r_1)_\gamma - H - r_2,
\qquad
v^t := \nabla\psi_\gamma(x^t) - w^t .
\]
Note $\psi_\gamma$ is $L_\psi$-smooth with $L_\psi = L + 1/\gamma \le L_\gamma$, and
$v^t \in \nabla\psi_\gamma(x^t) - \partial r_2(x^t)$, so $\|v^t\|$ upper-bounds the distance
from $0$ to that set: $v^t \to 0$ certifies criticality of $\Phi_\gamma$.

\begin{theorem}[Subgradient variant: guarantee for $\Phi_\gamma$]
\label{thm:smdca-subgrad}
Let Assumptions~\ref{assu:wellposed}, \ref{assu:sc}, \ref{assu:smooth-H}, \ref{assu:G-oracle}
hold (so that every $w^t \in \partial r_2(x^t)$ exists) and fix $\gamma > 0$ with
$\Phi_\gamma^* := \inf_x \Phi_\gamma(x) > -\infty$ (sufficient conditions in
Lemma~\ref{lem:Fgamma-lb}). Run the subgradient variant under the schedule
\eqref{eq:smdca-params}, with $\kappa_V = \alpha/\sqrt{96}$ for Polyak and
$\kappa_V = \sqrt{\alpha b}/\sqrt{192}$ for MVR, and put
$\tilde{\bar\Ls} := \Phi_\gamma(x^0) - \Phi_\gamma^* + \tfrac{\gamma}{4}\|p_1^0\|^2$. Then
\[
\frac{1}{T}\sum_{t=0}^{T-1}\E\|v^t\|^2 \;\le\;
\frac{8\tilde{\bar\Ls}}{\eta^0 T} + \frac{24 V^0}{\alpha T}
+ \frac{86\,\eta^1 M^2}{\gamma} + \frac{C_4\,\alpha\sigma^2}{b},
\]
with $C_4 = 24$ (Polyak) and $C_4 = 48$ (MVR). Consequently the complexity statements of
Theorem~\ref{thm:smdca} hold verbatim with $F_\gamma$ replaced by $\Phi_\gamma$, for every
$\gamma > 0$ for which $\Phi_\gamma^* > -\infty$; no prox of $r_2$, no smoothness of $r_2$ and
no bound on $\partial r_2$ is required. The certificate is \emph{selection-dependent}: it
controls $v^t = \nabla\psi_\gamma(x^t) - w^t$ for the subgradients $w^t$ chosen by the
algorithm, which upper-bounds $\dist(0, \nabla\psi_\gamma(x^t) - \partial r_2(x^t))$ but is
not a unique stationarity mapping (Remark~\ref{rem:subgrad-not-Fgamma}).
\end{theorem}

\begin{proof}
\emph{Error decomposition.} Since $\nabla\psi_\gamma(x^t) = p_1(x^t) - \nabla H(x^t)$ with
$p_1(x^t) = (x^t - u^*(x^t))/\gamma$,
\[
\xi^t := \mathcal G^t - v^t
= \frac{x^t - u^{t+1}}{\gamma} - w^t - g^t - \frac{x^t - u^*(x^t)}{\gamma} + \nabla H(x^t) + w^t
= -\frac{1}{\gamma}\bigl(u^{t+1} - u^*(x^t)\bigr) - e^t ,
\]
which is exactly \eqref{eq:xi-decomp}: the $w^t$ terms cancel identically, because $w^t$ is
both the algorithm's own subgradient and the first-order object of $\Phi_\gamma$ at $x^t$.
Hence \eqref{eq:xi-norm-new} holds unchanged.

\emph{Descent.} Let $\Delta := x^{t+1} - x^t = -\eta^0 \mathcal G^t$. By $L_\psi$-smoothness
of $\psi_\gamma$ and convexity of $r_2$ at $x^t$ with the \emph{same} $w^t$, i.e.\
$r_2(x^{t+1}) \ge r_2(x^t) + \langle w^t, \Delta\rangle$ (valid for nonsmooth $r_2$),
\begin{align*}
\Phi_\gamma(x^{t+1}) - \Phi_\gamma(x^t)
&= \bigl[\psi_\gamma(x^{t+1}) - \psi_\gamma(x^t)\bigr] - \bigl[r_2(x^{t+1}) - r_2(x^t)\bigr] \\
&\le \langle \nabla\psi_\gamma(x^t), \Delta\rangle + \tfrac{L_\psi}{2}\|\Delta\|^2
   - \langle w^t, \Delta\rangle
= -\eta^0\langle v^t, \mathcal G^t\rangle + \tfrac{L_\psi(\eta^0)^2}{2}\|\mathcal G^t\|^2 .
\end{align*}
This is Step~1 of Appendix~\ref{app:thm-smdca-proof} with $\nabla F_\gamma(x^t)$ replaced by
$v^t$ and $L_\gamma$ by $L_\psi$. Since $L_\psi = L + 1/\gamma \le L + 2/\gamma = L_\gamma$,
the condition $\eta^0 \le 1/(4L_\gamma)$ of \eqref{eq:eta0-constraints} implies
$\eta^0 \le 1/(4L_\psi)$, which is what Step~1 requires; hence
\eqref{eq:descent-step} holds with $v^t$ in place of $\nabla F_\gamma(x^t)$ under the
schedule \eqref{eq:smdca-params} unchanged.

\emph{Remainder.} Steps~2--6 of Appendix~\ref{app:thm-smdca-proof} involve only $\xi^t$,
$\mathcal E^t$, $V^t$ and the drift $D^t$, none of which references $r_2$ or $F_\gamma$.
Running them with $P^t := \E\|v^t\|^2$ yields \eqref{eq:VF-bound} with $\bar\Ls$ replaced by
$\tilde{\bar\Ls}$. The MVR case is Step~7$'$ verbatim.
\end{proof}

\begin{remark}[The subgradient variant does not target $F_\gamma$]
\label{rem:subgrad-not-Fgamma}
Criticality of $\Phi_\gamma$ reads $\nabla\psi_\gamma(x) \in \partial r_2(x)$, whereas
$\nabla F_\gamma(x) = 0$ reads $\nabla\psi_\gamma(x) = p_2(x)$, where $p_2(x)$ is the unique
vector with $p_2 \in \partial r_2(x - \gamma p_2)$: a subgradient of $r_2$ at $x$ need not
equal the Moreau gradient at $x$, which is a subgradient at the proximal point $z^*(x)$.
The discrepancy does not close as $\gamma \to 0$, because a nonsmooth $\partial r_2$ can
jump by $\Theta(\sup\|\partial r_2\|)$ across the distance
$\|x - z^*(x)\| \le \gamma\sup\|\partial r_2\|$, however small.

Concretely, take $n = 1$, $G + r_1 = x^2$, $H(x) = \tfrac12 x^2 - \tfrac{\lambda}{2}x$,
$r_2 = \lambda|x|$ and any $\gamma < \tfrac12$. Then $\rho_1 = 2 \ge \rho_2 = 1$, $L = 1$,
$F(x) = \tfrac12 x^2 + \tfrac{\lambda}{2}x - \lambda|x|$ is bounded below, and $\Phi_\gamma$
is bounded below. At $x = 0$ we have $p_1(0) = 0$, $\nabla H(0) = -\lambda/2$, and the choice
$w^0 = \lambda/2 \in \partial r_2(0) = [-\lambda,\lambda]$ is admissible, giving
$v^0 = 0 - \tfrac{\lambda}{2} - (-\tfrac{\lambda}{2}) = 0$: the certificate of
Theorem~\ref{thm:smdca-subgrad} is satisfied exactly at $x = 0$ (and with exact inner solve
and exact gradients the iterate is a genuine fixed point). Yet $p_2(0) = 0$, so
$\|\nabla F_\gamma(0)\| = \lambda/2$. Hence the $F_\gamma$-stationarity gap of the
subgradient variant is $\Theta(\sup\|\partial r_2\|)$ for every $\gamma$, and it is
choice-dependent: $w^0 = \lambda$ would move the iterate. This is why
Theorem~\ref{thm:smdca} assumes $\prox_{\gamma r_2}$ in closed form when $F_\gamma$ is the
target.
\end{remark}

\begin{remark}[$\Phi_\gamma$ as an alternative primary target]
\label{rem:phi-gamma-future}
Since $\Phi_\gamma$ smooths only one of the two convex components, it is closer to $F$ than
$F_\gamma$: repeating the argument of Lemma~\ref{lem:gamma-equivalence}(iii) for
$\Phi_\gamma$ compares $x$ with $u^*(x)$ at distance $\gamma\|p_1(x)\|$ and yields
$\hat\eps \le \eps + \gamma(L + L_{r_2})\|p_1(x)\|$, i.e.\ a single smoothing error rather
than two. It also requires neither $\prox_{\gamma r_2}$ nor smoothness of $r_2$, and gives
$L_\psi = L + 1/\gamma$ in place of $L_\gamma = L + 2/\gamma$. Building the single-loop
algorithm around $\Phi_\gamma$ from the outset is therefore an attractive alternative; we
keep $F_\gamma$ as the primary target here because it makes the surrogate differentiable and
permits the standard smooth-descent template.
\end{remark}

\section{Constant-stepsize behaviour of momentum-free methods}
\label{app:noise-floor}

For context, we record an elementary observation about constant-stepsize behaviour. The
instance is fixed in advance and bounded below; the result records what cannot be done at
constant stepsize, and is \emph{not} a claim that momentum is necessary in any stronger
sense.

\begin{proposition}[Constant-stepsize floor, no momentum]
\label{prop:noise-floor}
Fix $L > \mu > 0$. Let $G(x) = (L/2)\|x\|^2$, $H(x) = ((L-\mu)/2)\|x\|^2$, $r_1 = r_2 = 0$,
so $F(x) = (\mu/2)\|x\|^2$ is $\mu$-strongly convex with $F^* = 0$. Take stochastic
gradients $\nabla h_\xi(x) = (L-\mu) x + \xi$, $\xi \sim \mathcal{N}(0, \sigma^2 I_d)$. For
Algorithm~\ref{alg:mdca} with $\alpha_t \equiv 1$ (no momentum), $\eta_t \equiv 1$, and
constant batch $b$, the iterates satisfy
\[
\lim_{t \to \infty}\E\|\nabla F(x^t)\|^2 \;=\; \frac{\beta_*}{2 - \beta_*}\, \frac{\sigma^2 d}{b},
\qquad \beta_* := \mu/L,
\]
which is bounded away from zero. Here the oracle noise is state-independent and its variance
is exactly $\sigma_A^2 := \E\|\nabla h_\xi(x) - \nabla H(x)\|^2 = \sigma^2 d$ at every $x$,
so Assumption~\ref{assu:smooth-H} holds with constant $\sigma_A^2$ and \emph{with equality};
the limit equals $\tfrac{\beta_*}{2-\beta_*}\,\sigma_A^2/b$. The floor is removed, on the
same instance, by damped momentum (Theorems~\ref{thm:polyak-rate}--\ref{thm:mvr-rate}), by
decreasing stepsizes, or by an increasing batch: since the floor scales as $\sigma_A^2/b$,
any $b \ge \sigma_A^2/\eps^2$ brings it below $\eps^2$.
\end{proposition}

\begin{proof}
With $G(x) = (L/2)\|x\|^2$, $H(x) = ((L-\mu)/2)\|x\|^2$, $r_1 = r_2 = 0$, the inner
subproblem in Algorithm~\ref{alg:mdca} has closed form: minimizing
$(L/2)\|x\|^2 - \langle g^t, x\rangle$ gives $\tilde x^{t+1} = g^t/L$. With $\alpha_t = 1$ and $\eta_t = 1$ (the no-momentum, undamped setting of SDCA, not the
schedule of Theorem~\ref{thm:polyak-rate}), $g^t = \hat\nabla H_{\mathcal I^t}(x^t) = (L-\mu) x^t + \bar\xi^t$ where
$\bar\xi^t = (1/b)\sum_{i \in \mathcal I^t}\xi_i \sim \mathcal{N}(0, (\sigma^2/b) I_d)$, so
$x^{t+1} = (1 - \mu/L)\, x^t + \bar\xi^t/L$.

With $\rho^* := 1 - \mu/L$ and $a^* := 1/L$, stationarity in distribution gives
$\E\|x^{t+1}\|^2 = (\rho^*)^2 \E\|x^t\|^2 + (a^*)^2 \sigma^2 d/b$ (the cross term vanishes
by independence of $\bar\xi^t$ and $x^t$), so
$\E\|x^t\|^2 \to (a^*)^2\sigma^2 d/(b(1 - (\rho^*)^2))$. Now
$1 - (\rho^*)^2 = (\mu/L)(2 - \mu/L) = \beta_*(2 - \beta_*)$, so
$\lim\E\|x^t\|^2 = \sigma^2 d/(b L^2 \beta_*(2 - \beta_*))$. Since $\nabla F(x) = \mu x$,
$\lim\E\|\nabla F(x^t)\|^2 = \beta_*/(2 - \beta_*) \cdot \sigma^2 d/b$.
\end{proof}

The proposition shows that, on this specific instance and \emph{at constant stepsize and
constant batch}, the momentum-free method has a noise floor that does not vanish. This is
the genuine existence-of-a-floor statement referred to in Section~\ref{sec:separation};
Lemma~\ref{lem:separation} only provides upper bounds and cannot establish such a floor. The
proposition is one of several ways to escape the floor (momentum, decreasing stepsizes,
growing batches) and does not show that momentum is necessary, nor that it is preferable to
the alternatives in any oracle-complexity sense.

\end{document}